\documentclass[letterpaper, 10 pt, conference]{ieeeconf}  % Comment this line out if you need a4paper

\IEEEoverridecommandlockouts                              % This command is only needed if 
\usepackage{amsmath,amssymb,amsfonts}
\usepackage{mathtools}
\usepackage{algorithmic}
\usepackage{graphicx}
\usepackage{textcomp}
\usepackage{mdframed}
\let\labelindent\relax
\usepackage{enumitem}
\usepackage{xcolor}

\usepackage{subcaption}
\usepackage{booktabs}
\usepackage{multirow}
\usepackage{tikz}
\usepackage{xspace}
\usepackage{wrapfig}
\usepackage{pifont}

\usepackage[backend=biber,
            style=ieee,
            natbib=true, 
            url=false,
            eprint=false,
            doi=false]{biblatex} 
\usepackage{hyperref}
\hypersetup{
    breaklinks=true,letterpaper=true,colorlinks,citecolor=blue
}

\newcommand{\cmark}{\ding{51}}%
\newcommand{\xmark}{\ding{55}}%
\usepackage{listings}

\lstdefinestyle{mystyle}{
    backgroundcolor=\color{backcolour},
    basicstyle=\ttfamily\footnotesize,
    backgroundcolor=\color{backcolour},
    commentstyle=\color{codegreen},
    keywordstyle=\color{blue},
    numberstyle=\tiny\color{codegray},
    stringstyle=\color{codepurple},
    breaklines=true,
    breakatwhitespace=true,
    columns=fullflexible,
    keepspaces=true,
    showstringspaces=false,
    frame=single,
    tabsize=2,
    language=Python
}

\lstdefinestyle{prompt}{
    basicstyle=\ttfamily\scriptsize,
    breaklines=true,
    breakatwhitespace=true,
    columns=fullflexible,
    keepspaces=true,
    showstringspaces=false,
    frame=single,
    backgroundcolor=\color{backcolour},
    postbreak=\mbox{\textcolor{codegray}{$\hookrightarrow$}\space},
    language={},
}

\definecolor{codegreen}{rgb}{0,0.5,0}
\definecolor{codegray}{rgb}{0.5,0.5,0.5}
\definecolor{codepurple}{rgb}{0.58,0,0.82}
\definecolor{backcolour}{rgb}{0.97,0.97,0.97}

\usepackage{cleveref}
\crefname{equation}{Eq.}{Eqs.}
\Crefname{equation}{Eq.}{Eqs.}
\crefname{table}{Tab.}{Tabs.}
\Crefname{table}{Tab.}{Tabs.}
\crefname{figure}{Fig.}{Figs.}
\Crefname{figure}{Fig.}{Figs.}
\crefname{section}{Sec.}{Secs.}
\Crefname{section}{Sec.}{Secs.}

\definecolor{method}{HTML}{D81B60}
\definecolor{gvl}{HTML}{1E88E5}
\definecolor{topreward}{HTML}{9467BD}
\definecolor{robometer}{HTML}{FF7F0E}
\definecolor{roboreward}{HTML}{FCBF07}
\definecolor{baseline}{HTML}{007D68}
\definecolor{soler1}{HTML}{17BECF}

\newcommand{\Method}{\textcolor{method}{\texttt{\method}}\xspace}
\newcommand{\Methodalpha}{\textcolor{method}{\texttt{D\&A$^*$}}\xspace}
\newcommand{\naive}{\textcolor{baseline}{\texttt{Naive VLM}}\xspace}
\newcommand{\gvl}{\textcolor{gvl}{\texttt{GVL}}\xspace}
\newcommand{\topreward}{\textcolor{topreward}{\texttt{TOPReward}}\xspace}
\newcommand{\robometer}{\textcolor{robometer}{\texttt{Robometer}}\xspace}
\newcommand{\soler}{\textcolor{soler1}{\texttt{SOLE-R1}}\xspace}
\newcommand{\roboreward}{\textcolor{roboreward}{\texttt{RoboReward}}\xspace}
\newcommand{\roborewardfourb}{\textcolor{roboreward}{\texttt{RoboReward-4B}}\xspace}

\title{\LARGE \bf \raisebox{-0.6ex}{\includegraphics[height=8mm]{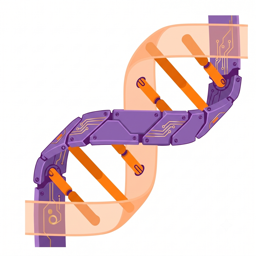}} D\&A: Decoupling Description and Assessment \\Enables VLMs as Zero-Shot Robotic Reward Models}

\author{
Marius Memmel$^{1,2}$, Shawn Yin$^{1,*}$, Urvi Rutia$^{1,*}$, Angad Wadhwa$^{1,*}$, Eldon Li$^{1}$, Gavin Angeloff$^{1}$\\
Ankit Goyal$^{2}$, Jesse Zhang$^{1}$, Dieter Fox$^{1}$, Fabio Ramos$^{2}$, Anqi Li$^{2,\dag}$, Abhishek Gupta$^{1,\dag}$%
\\$^{1}$University of Washington, $^{2}$NVIDIA, $^*$Equal Contribution, $^\dag$Equal Advising
}

\begin{document}

\newcommand{\todo}[1]{\textcolor{red}{TODO: #1}}
\newcommand{\method}{D\&A\xspace}

\maketitle

\thispagestyle{empty}
\pagestyle{empty}

%%%%%%%%%%%%%%%%%%%%%%%%%%%%%%%%%%%%%%%%%%%%%%%%%%%%%%%%%%%%%%%%%%%%%%%%%%%%%%%%
\begin{abstract}

Vision-Language Models (VLMs) hold promise for zero-shot video task-progress estimation in robot learning, yet existing frameworks frequently suffer from confirmation bias, hallucinating progress in failed executions due to implicit success assumptions or relative temporal ranking designs. Na\"ive prompting of frontier models collapses into a binary success detector that fails to capture nuanced, non-linear progress and failures. While post-training approaches can mitigate this through explicit training, they demand expensive real-world robotics data and require computationally intensive post-training for every individual base model.
To resolve these limitations, we introduce D\&A (Describe \& Assess): our approach decouples progress estimation into an objective, task-grounded textual description of scene dynamics followed by an assessment phase using explicit rubrics. By compelling the VLM to synthesize an explicit, text-based definition of absolute success before scoring, the framework anchors evaluations to an external, global criterion rather than the video's internal timeline. Empirical results demonstrate that our method calibrates zero-shot estimates and mitigates confirmation bias, providing fine-grained, partial progress metrics that accurately penalize failed executions while correctly interpolating successful runs. Being purely prompt-based, D\&A generalizes across tasks, embodiments, and benchmarks---matching or surpassing post-trained reward models---and its reward estimates enable efficient downstream reinforcement learning.

Website at \url{https://weirdlabuw.github.io/dna}

\end{abstract}

\section{Introduction}

First impressions are notoriously difficult to overcome~\citep{gregg2006easier}. When evaluating complex situations, humans often exhibit confirmation bias~\citep{nickerson1998confirmation}, seeking and interpreting evidence that aligns with their initial hypotheses and introducing systematic errors into judgment. Similar biases appear in language models, which, trained to appear competent to human evaluators, distort answers to agree with users or take ``shortcuts'' to appear to solve given tasks~\citep{kim2025challenging, zhong2026impossiblebench, feng2026good, petrov2025brokenmath, wan2025unveiling, cheng2026elephant}. Crucially, the psychology literature shows such biases can be mitigated through structured procedures that mandate a comprehensive \emph{description} of the available data before any task \emph{assessment}~\cite{croskerry2003cognitive, larrick2004debiasing}.

\begin{figure}[t]
    \centering
    \includegraphics[width=1.0\linewidth]{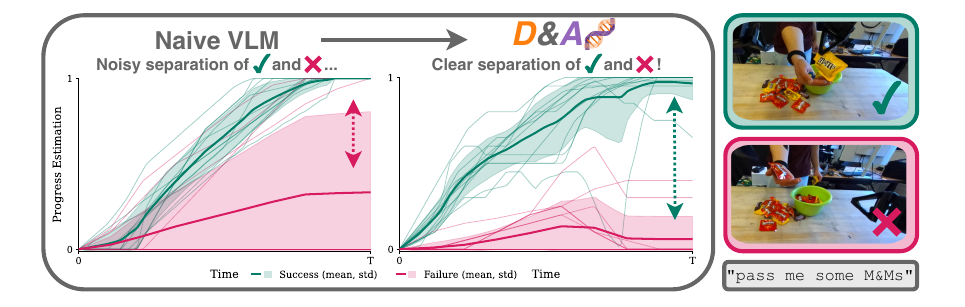}
    \caption{\textbf{Confirmation Bias in Na\"ive VLM Reward Estimation:} Zero-shot querying causes models to hallucinate progress even on failed tasks, resulting in high prediction variance and no clear difference in reward estimates for \textcolor[HTML]{027d68}{successful} and \textcolor[HTML]{e25689}{failed} trajectories.}
    \label{fig:teaser}
\end{figure}

We build on this intuition to design better robotics \emph{reward models} for downstream applications, e.g., real-world reinforcement learning and failure explanation. When na\"ively querying Vision-Language Models (VLMs) for zero-shot video reward estimation, we observe these same confirmation-bias shortcuts: the models frequently hallucinate progress even when little to no meaningful advancement has occurred (see \cref{fig:teaser}). Current zero-shot robotics reward models fall victim to the same issue---frameworks like GVL~\citep{ma2024vision} and TOPReward~\citep{chen2026topreward} estimate progress by ranking shuffled video frames or tracking token logits---our experiments show that these approaches bias towards task success. For downstream reinforcement or imitation learning algorithms, these erroneous evaluations become highly misleading reward signals that severely impede policy learning.

% --- Problems with Post-Trained Methods ---
To bypass the limitations of zero-shot foundation models, recent approaches post-train models on task-specific robotic datasets~\cite{zhai2025vision, tan2025robo, lee2026roboreward, liang2026robometer}. Yet diverse real-world robotics data, with labeled failures, is hard to obtain~\citep{tian2026position}, limiting the generalization capabilities of the resulting models. Post-training is further limited to smaller, less capable open-source models and must be repeated for each individual base VLM, creating a cost, compute, and capability barrier to truly generalist robotics reward models.

% --- Solution: Phase 2 (Success Grounding) ---
Instead, we propose \method (\textbf{D}escribe \textbf{\&} \textbf{A}ssess), a zero-shot reward model that avoids the bias issues of prior zero-shot methods while achieving better visual and semantic generalization than post-trained reward models.
Being zero-shot is central to its effectiveness: requiring no robotics-specific training, the same prompt-based recipe drops directly onto any promptable VLM and \emph{improves for free} as the underlying models advance.
\method decouples evaluation into two distinct stages, \emph{description} and \emph{assessment}.
It first forces the VLM to generate an explicit, textual \emph{description} of what the robot actually did, inducing an objective foundation for progress measurement.
It then \emph{assesses} progress by reasoning about what task success means semantically and anchoring its reward estimate to the success criterion.
Conditioned on the trajectory description and this success reasoning, the VLM predicts progress over key frames in $[0, 1]$ as dense rewards for policy learning.
We further show that this paradigm can be easily adapted to generate rich natural language feedback that goes beyond binary reward signals and provides targeted corrective context for policy improvement.

In summary, \method is a reward model paradigm that decouples video description from progress assessment, enabling zero-shot robotics reward modeling with pre-trained VLMs.
Because it is purely prompt-based, \method is model-agnostic and requires no robotics data or fine-tuning; it generalizes across tasks, embodiments, and language, matching or outperforming post-trained models on reward benchmarks. Our scaling experiments show \method \emph{improves with VLM capability} across four model families and 28 open- and closed-source models.
On real-world online and offline reinforcement learning with a DROID Franka and two bimanual YAM arm tasks, \method improves success rate over both zero-shot and post-trained reward models by $+95\%$, $+15\%$, and $+45\%$, respectively. Finally, scaling up \method to GPT-6 Astra achieves SOTA performance on \texttt{D\&A-Bench}, \texttt{RBM-EVAL-OOD}, and \texttt{RoboRewardBenchmark}, surpassing all seven pre-trained and zero-shot baselines.

\section{Related Work}
\label{sec:related}
\noindent\textbf{Foundation models as reward models.}
LLMs and VLMs have been widely used to specify rewards, e.g., by generating executable reward code or shaping terms from natural language~\citep{ma2023eureka, yu2023language, xie2024textreward, kwon2023reward}, but such methods typically assume access to privileged low-dimensional state that is unavailable in real-world deployment.
A more deployable alternative uses pre-trained VLMs as zero-shot progress or success estimators directly from pixels~\citep{roboclip, ma2023vipuniversalvisualreward, rocamonde2024visionlanguage, ma2024vision, chen2026topreward, yang2026rarm}.
\method belongs to this zero-shot family, but whereas prior methods extract a scalar signal directly---e.g., ranking shuffled frames in GVL~\citep{ma2024vision} or reading the logit of a ``True'' token in TOPReward~\citep{chen2026topreward}---we route reward estimation through explicit description and assess phases, which we find essential for countering the success bias of direct prompting.

\noindent\textbf{Post-trained reward models.}
To improve reliability, a growing line of work fine-tunes VLMs and VLAs on robotics data with progress-prediction objectives~\citep{zhai2025vision, lee2026roboreward, liang2026robometer, tan2025robo, zhang2025rewind, intelligence2025pi, huang2026rynnvalue} or language reasoning~\cite{zhang2026progresslm, schroeder2026sole}.
RoboReward~\citep{lee2026roboreward} predicts discretized progress labels obtained via counterfactual instruction relabeling, while Robometer~\citep{liang2026robometer} jointly predicts progress, success, and trajectory preferences from a large curated dataset.
These models inherit the cost of collecting diverse, high-quality robot data~\citep{tian2026position}, are restricted to open-source backbones, and must be retrained as stronger base models appear.
\method instead remains fully zero-shot, making it model-agnostic and improving with frontier VLM improvements, while matching or exceeding post-trained models on out-of-distribution benchmarks.

%\textbf{Bias and structured reasoning in foundation models.}
%LLMs and VLMs trained on human feedback exhibit systematic biases---sycophancy, shortcut-taking, and a tendency to confirm user assumptions~\citep{kim2025challenging, zhong2026impossiblebench, feng2026good, petrov2025brokenmath, wan2025unveiling, cheng2026elephant}---which surface as hallucinated progress when these models are queried for rewards.
%A parallel line of work improves model judgments through structured reasoning, such as chain-of-thought prompting~\citep{wei2022chain} and explicit rubric- or criteria-based evaluation~\citep{gunjal2025rubrics}.
%Motivated by cognitive-science debiasing procedures that separate observation from judgment~\citep{croskerry2003cognitive, larrick2004debiasing}, \method operationalizes this idea for robotics rewards by decoupling an instruction-blind \emph{description} from a success criterion-grounded \emph{assessment} stage, mitigating confirmation bias. % without any task-specific training.

\section{Demonstrating Bias in Zero-Shot VLM Prompting}
\label{sec:method_motivation}

\begin{figure}
\vspace{1em}
\centering
\includegraphics[width=\linewidth]{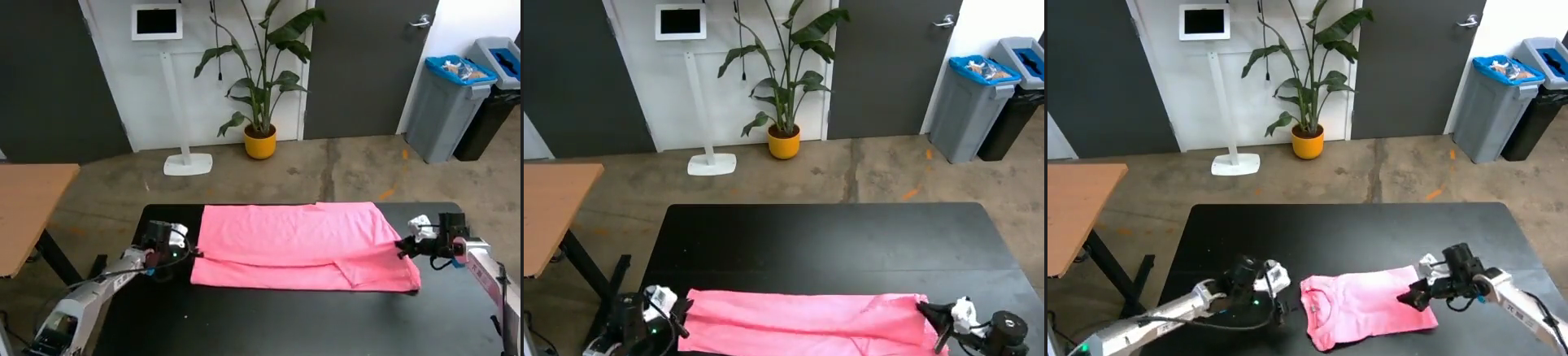}
\begin{lstlisting}[basicstyle=\scriptsize\ttfamily]
rewards = []
prev_frame = None
for frame in sample_frames(video):
    if prev_frame is None:
        rewards.append(0.0)
    else:
        reward = mean_diff(frame, prev_frame)
        rewards.append(reward)
    prev_frame = frame
rewards = normalize_0_to_1(rewards)
\end{lstlisting}
\caption{Example reasoning trace generated by GPT-5.3 (queried April 11, 2026) for the \emph{opposite instruction} ``unfold the shirt.''}
\label{fig:zeroshot_vlm_bias}
\end{figure}

We begin with a qualitative example showing that directly prompting a zero-shot VLM~\citep{singh2025openai} for per-frame progress---conditioned on a robot video and its task instruction---fails to produce accurate rewards.
Given an \emph{expert demonstration} paired with the correct instruction, VLMs generally produce reasonable per-frame $[0, 1]$ rewards that end at $1.0$.
However, when the robot performs the task incorrectly or suboptimally, they often ``shortcut'' by assuming success---precisely the case a reward model must detect to correct a deployed policy.

For example, in \Cref{fig:zeroshot_vlm_bias} we prompt GPT-5.3 with ``provide per-timestep rewards for this video of a robot `unfolding the shirt'.'' However, the video (from the $\pi_0$-FAST blog~\citep{pertsch2025fast}) in fact shows a $\pi_0$ policy \emph{folding} a shirt.
Rather than reasoning about task completion, the model writes a reward function that scores normalized \emph{mean pixel differences} between consecutive frames, rewarding motion rather than progress.
This reflects a confirmation bias arising from fine-tuning on human feedback, which makes models unlikely to challenge user assumptions~\citep{kim2025challenging, zhong2026impossiblebench, feng2026good, petrov2025brokenmath, wan2025unveiling, cheng2026elephant}.

While prompt engineering can mitigate specific cases, our experiments (\Cref{sec:experiments}) show that existing zero-shot reward methods---built on both closed- and open-source VLMs with custom prompting and reward-extraction schemes---still perform poorly across benchmarks, real-world tasks, and embodiments.
Next, we detail our solution: decoupled \emph{description} and \emph{assessment} phases.

\section{\method: Decoupling Description and Assessment}
\begin{figure*}[tb]
    \centering
    \includegraphics[width=1.00\linewidth]{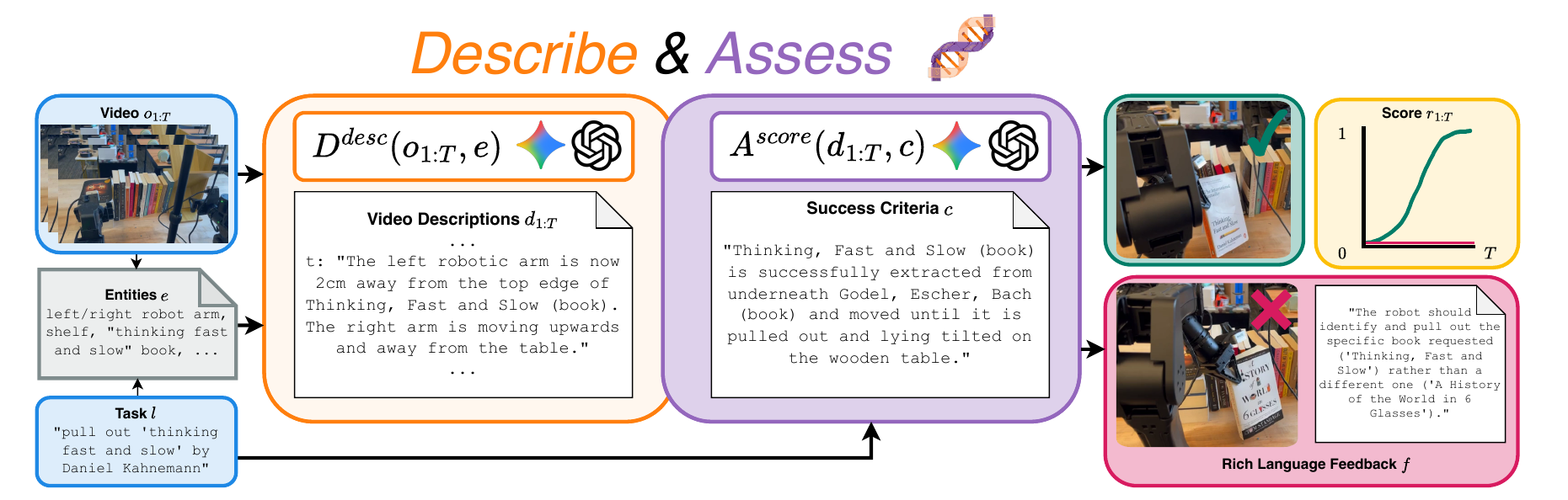}
    \caption{\textbf{Overview of \method{}.} Given a video $o_{1:T}$ and instruction $l$, \method{} decomposes reward prediction $R(o_{1:T}, l) = A\big(l,\, D(o_{1:T})\big)$ into a \emph{Describer} $D$ and an \emph{Assessor} $A$. $D$ narrates the trajectory \emph{blind to $l$} to avoid confirmation bias, producing a temporally consistent description $d_{1:T}$; $A$ then scores each timestep of $d_{1:T}$ against an explicit success criterion for $l$, yielding dense per-frame rewards $r_{1:T} \in [0,1]$ as well as rich language feedback.}
    \label{fig:method}
\end{figure*}

\noindent\textbf{Problem Formulation.}
Given an image sequence (video) $o_{1:T}\coloneqq \{o_1, o_2,\ldots,o_T\}$ of length $T$ of a robot performing a task specified by natural language instruction $l$, our goal is to predict a corresponding sequence of reward values $r_{1:T}\coloneqq \{r_1, r_2, \dots, r_T\}$. 
Each scalar value $r_t \in [0, 1]$ represents the estimated reward at timestep $t$, where $r_t = 0$ indicates no progress and $r_t = 1$ corresponds to task completion. 
We construct a mapping function $R : (o_{1:T},\, l) \mapsto r_{1:T}$ that estimates rewards from the multi-modal inputs $o_{1:T}$ and $l$.
Prior works either post-train a VLM to learn $R$~\citep{zhai2025vision,tan2025robo,lee2026roboreward,liang2026robometer}, directly query a VLM with a custom prompt and shuffled image inputs~\citep{ma2024vision, budzianowski2025opengvl}, or produce $r_{1:T}$ by inspecting the logits of a specific token after processing $o_{1:T}, l$~\citep{chen2026topreward}.

\noindent\textbf{Method Overview.}
We would like to directly use pre-trained VLMs, zero-shot, for reward prediction; doing so allows us to leverage the generalization capabilities and learned reasoning chains of ever-improving frontier VLMs trained on internet-scale data, all without the tedium of robotics-specific post-training. 
But in practice, as demonstrated in \Cref{sec:method_motivation}, we find that na\"ively instantiating $R(o_{1:T}, l)$ as a VLM suffers from confirmation bias, where the model predicts increasing rewards regardless of whether the observation-language pairing represents a success or failure.
To address these limitations, we propose Describe and Assess (\method{}), which decomposes $R$ into two components with an explicit natural-language interface:
\begin{equation}
    \label{eq:dna}
    \resizebox{1\linewidth}{!}{$
    \begin{array}{lll}
        \text{Describer} & d_{1:T} = D(o_{1:T}), & \color{gray}{\vartriangleright\text{Describes observed behavior in language}}\\
        \text{Assessor} & r_{1:T} = A(l, d_{1:T}). & \color{gray}{\vartriangleright\text{Scores behavior against description}}\\
    \end{array}$}
\end{equation}
Both functions can use the same VLM (or another function with the same specification) but are specified by different inputs and prompts and have no shared context. Each function has a specific task to help prevent inherent VLM biases that can contaminate reward generation.
The full reward function is the composition $R(o_{1:T}, l) = A\big(l,\,  D(o_{1:T})\big)$.
While we use $o_{1:T}$ notation here, in practice, we linearly downsample the $T$ frames to $H$ frames with $H\ll T$, e.g., $H=8$, and then linearly interpolate the rewards to length $T$ to reduce inference cost and time.
We detail the describer next in \Cref{sec:method:describer}, followed by the assessor in \Cref{sec:method:assessor}. We provide a visual overview of \method in \Cref{fig:method}.

\subsection{Describing the Robot Trajectory Video}
\label{sec:method:describer}

The role of the describer $D(o_{1:T})$ is to describe what the robot has accomplished in $o_{1:T}$ \emph{unconditioned} on the task instruction $l$.
Withholding $l$ avoids confirmation bias~\citep{nickerson1998confirmation, kim2025challenging}: a describer that knows the intended task tends to narrate behavior as if that task were being accomplished, regardless of what occurred. See \Cref{fig:method} for an example description.

\begin{figure*}[tb]
    \centering
    \includegraphics[width=1.\textwidth]{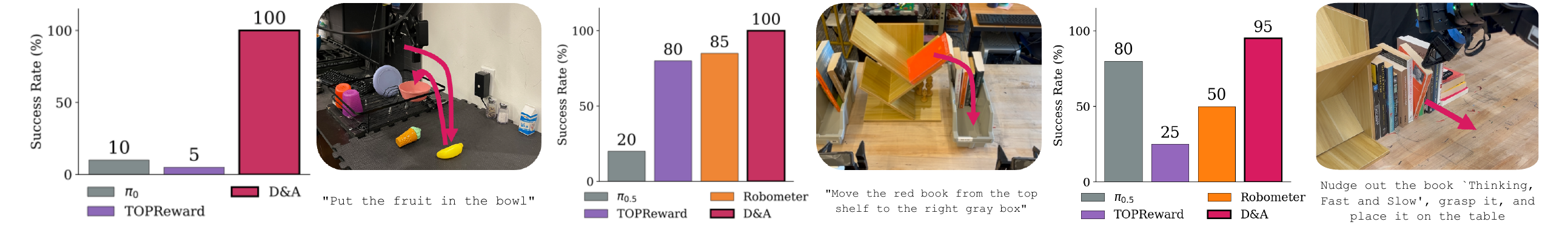}
    \caption{\textbf{Real-world RL Results.} \Method improves real-world online (\textit{left}) and offline reinforcement learning (\textit{middle, right}) while other zero-shot and post-trained models struggle. \textbf{Left:} in the online setting, \topreward fails to distinguish between success and failure, predicting high reward for both picking up the ice cream cone and the banana.
    \textbf{Middle, right:}Behavior cloning on suboptimal data hurts performance, while offline RL with \method leads to a higher success rate and faster completion time. \topreward and \robometer lack generalization and assign similar reward to optimal and suboptimal action chunks.}
    \label{fig:setup_dsrl}
\end{figure*}

\noindent\textbf{Describing the full trajectory at once.}
In contrast to prior work~\citep{zhai2025vision,tan2025robo,lee2026roboreward,liang2026robometer, zhang2025rewind} that predicts a single reward $r_t = R(o_{1:t}, l)$ causally and then iterates over $t$ to label dense rewards, \method{} consumes the entire trajectory $o_{1:T}$ in a single pass.
By describing the trajectory holistically, $D$ can place each timestep in the context of what happens before \emph{and after} it, yielding descriptions that are temporally consistent.

\noindent\textbf{Grounding the description in scene entities.}
Withholding $l$ avoids confirmation bias but introduces a new failure mode: without any knowledge of the task, the describer may omit details that turn out to be critical for downstream assessment.
Take the example in \Cref{fig:method}, where an entity extractor may describe the target book as ``Thinking, Fast and Slow by Daniel Kahnemann, ``the book with an orange spine'', ``the book with a pencil on the white cover'', etc. The entity should reflect the granularity in the instruction to avoid ambiguity in the description.
To resolve this without leaking $l$ into the describer, we decompose $D$ into two stages: an \emph{entity extractor} $E$ that lists \emph{all} scene entities $e$, both task-relevant and task-irrelevant, % (including every fork and spoon in \Cref{fig:underspecified_instruction}), 
from the instruction and video, and a \emph{grounded describer} $D^{\text{desc}}$ that narrates the trajectory anchored to those entities: $d_{1:T} = D^{\text{desc}}\big(o_{1:T},\, E(o_{1:T}, l)\big)$.
Since $D^{\text{desc}}$ does not know which entities are task-relevant, the full describer $D$ is effectively blind to $l$.
Next, we describe how the description is used by the assessor.

\subsection{Success Criteria Generation and Reward Assessment}
\label{sec:method:assessor}

Given the grounded description $d$ and instruction $l$, the assessor's role is to produce the per-timestep reward sequence $r_{1:T}$. 
At this stage, one might again be tempted to na\"ively query an LLM/VLM directly by giving it $(l, d)$ and asking it to score progress. 
In practice, however, we find this na\"ive version fails in a characteristic way: the assessor frequently collapses to outputting all zeros (when it cannot tell whether meaningful progress has been made) or becomes a binary success detector and simply interpolates from $0$ to $1$ as shown in \Cref{fig:teaser}.
However, to generate rewards useful for downstream applications, we would want them to encode much more granular progress information, e.g., partial progress towards task or subtask completions.
Such granular rewards would provide more useful, dense signals for policy learning, e.g., by improving the sample efficiency of reinforcement learning policies or by helping to filter out poor-quality actions in imitation learning~\citep{gupta2022unpacking}.

\noindent\textbf{Grounding the assessor in an explicit success criterion.}
This failure mode suggests the assessor lacks a clear \emph{target} for progress: $l$ specifies what to do, but not necessarily what describes task \emph{success}.
We address this by having the assessor first produce task completion explicitly. Concretely, given $(l, d_{1:T})$, the assessor internally generates a natural-language success criterion $c$---a description of the observable state of the world when $l$ has been fully accomplished---and then scores each timestep of the description against $c$ rather than against $l$ alone:
$r_{1:T} = A(l,\, d_{1:T}) = A^{\text{score}}\big(d_{1:T},\, C(l,\, d_{1:T}) \big),$
where $C$ generates the success criterion $c$ and $A^{\text{score}}$ scores the description against it.
Anchoring assessment to $c$ reframes ``estimate progress toward $l$'' as the better-posed question ``how close is $d$ to satisfying success criteria $c$?'', helping (1) admit a meaningful notion of partial progress and (2) resist both reward collapse modes.
In practice, criterion generation and scoring occur within the single assessor call $A$ via system prompting: the model is instructed to first articulate the success condition for $l$, then score each timestep of $d$ relative to it---forcing an implicit rubric~\citep{gunjal2025rubrics} design through CoT reasoning~\citep{wei2022chain}. See \Cref{fig:method} for an example.

\subsection{From Progress to Rich Language Feedback}
Scalar reward signals execution failure without offering actionable context for policy correction. By evaluating trajectory progress relative to defined success criteria, \method constructs an implicit semantic rubric over time. Tracking criteria fulfillment across timesteps enables \method to localize execution deviations, demonstrating an implicit understanding of \textit{how} and \textit{why} a trajectory failed.
To convert these diagnostic capabilities into downstream guidance, we augment the assessor $A$ with a feedback generator $F(l, d_{1:T})$ that outputs a text string $f$ containing a failure analysis and corrective instructions. This updates the assessor formulation to: $A(l, d_{1:T}) = A^{\text{score}}\big(d_{1:T}, C(l, d_{1:T}), F(l, d_{1:T}) \big)$.
See \Cref{fig:method} for feedback examples.

\subsection{Using \method{} for Downstream Policy Improvement}
\label{sec:method:policy}

The dense progress rewards and rich language feedback produced by \method{} can be readily leveraged across various downstream policy improvement paradigms. Here we outline how \method{} accelerates the paradigms used in our experiments in \Cref{sec:experiments}, online and offline reinforcement learning (RL) and Code-as-Policies (CaP)~\cite{liang2022code}. Though we specifically detail those, \method{} is a general-purpose reward and rich feedback framework and can be paired with other applications.

\noindent\textbf{Online RL.}
In online RL, the agent improves its behavior through active interaction with the environment. Here, the absolute progress score $r_{1:T} = R(o_{1:T}, l)$ from \method{} serves directly as the dense environment reward and can be optimized by any online RL algorithm. In our experiments, we fine-tune a frozen base policy with latent-space RL~\cite{wagenmaker2025steering}.

\noindent\textbf{Offline RL.} % via advantage conditioning
In offline RL, we are given a dataset $\mathcal{D} = \{\tau_n\}_{n=1}^N$ of trajectories \emph{without} ground-truth reward labels, and use \method{} to supply the missing reward signal. Since \method's progress scores are bounded within $[0, 1]$ and reflect global task completion, we treat them as a \emph{potential function}~\citep{Ng1999PolicyIU}, from which we derive per-timestep potential-based dense rewards $r_t$ and, in turn, offline advantages: $A_t = r_{min(t+C, T)} - r_t$ with action chunk size $C$.
We use these advantages for filtering the training data or conditioning a policy on them (cf. RECAP and CFGRL~\citep{intelligence2025pi, frans2025diffusion}); at test time, conditioning on a positive advantage biases generation towards optimal trajectories, maximizing downstream performance using \method's zero-shot rewards.

\noindent\textbf{Code-as-Policies.}
Recent observations in robot failure analysis~\citep{liu2023reflect} and feedback-driven optimization~\citep{lee2025feedback} suggest that rich language feedback can improve policy performance.
We evaluate \method{} as a between-attempt feedback source for an open-loop CaP planner. At every attempt, the planner observes the initial scene state and generates a code plan which is then executed open-loop in the environment. Across attempts, the planner keeps track of previous plans, and feedback generated by \method{} which identifies the failure modes and proposes corrections.

\section{Experimental Evaluation}
\label{sec:experiments}

\begin{figure*}[tb]
    \centering

    \begin{subfigure}[b]{0.7\linewidth}
        \centering
        \raisebox{0.0em}{\includegraphics[width=\linewidth]{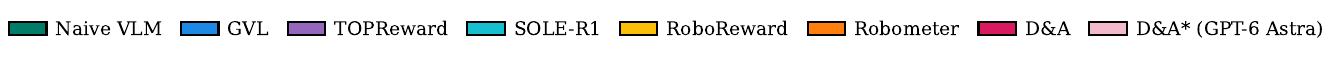}}
        % \raisebox{0.0em}{\includegraphics[width=\linewidth]{figures/legend_only.png}}
        \vspace{-1em}
    \end{subfigure}
    
    \begin{subfigure}[b]{0.3\linewidth}
        \centering
        \raisebox{0.1em}{\includegraphics[width=\linewidth]{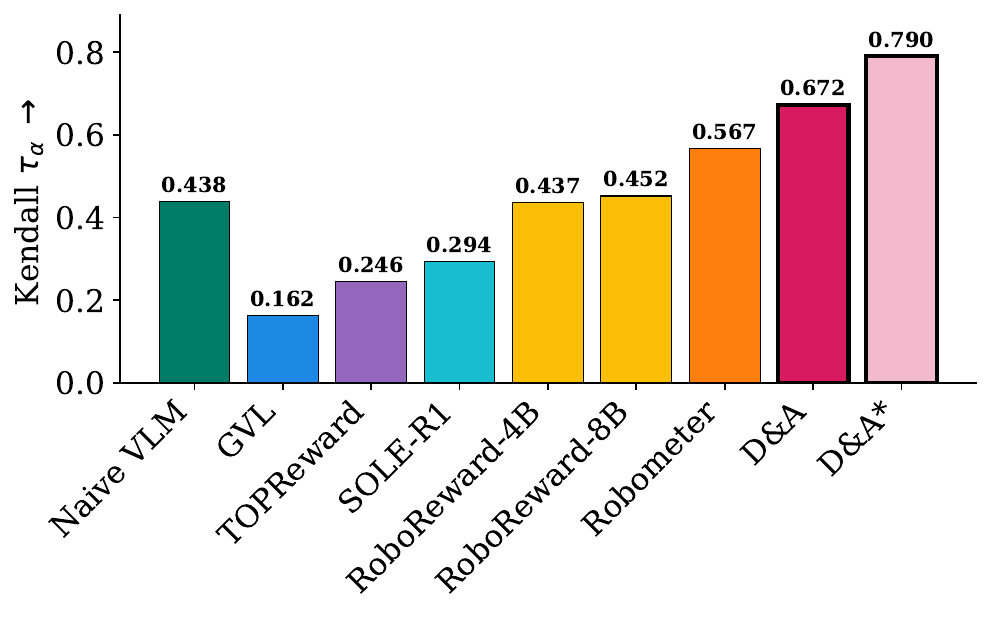}}
        % \raisebox{0.1em}{\includegraphics[width=\linewidth]{figures/kendall_policy_ranking_dna.png}}
        \caption{\texttt{D\&A-Bench} ($\uparrow$)}
        \label{fig:bimanualgen}
    \end{subfigure}
    \hfill
    \begin{subfigure}[b]{0.3\linewidth}
        \centering
        \raisebox{0.47em}{\includegraphics[width=\linewidth]{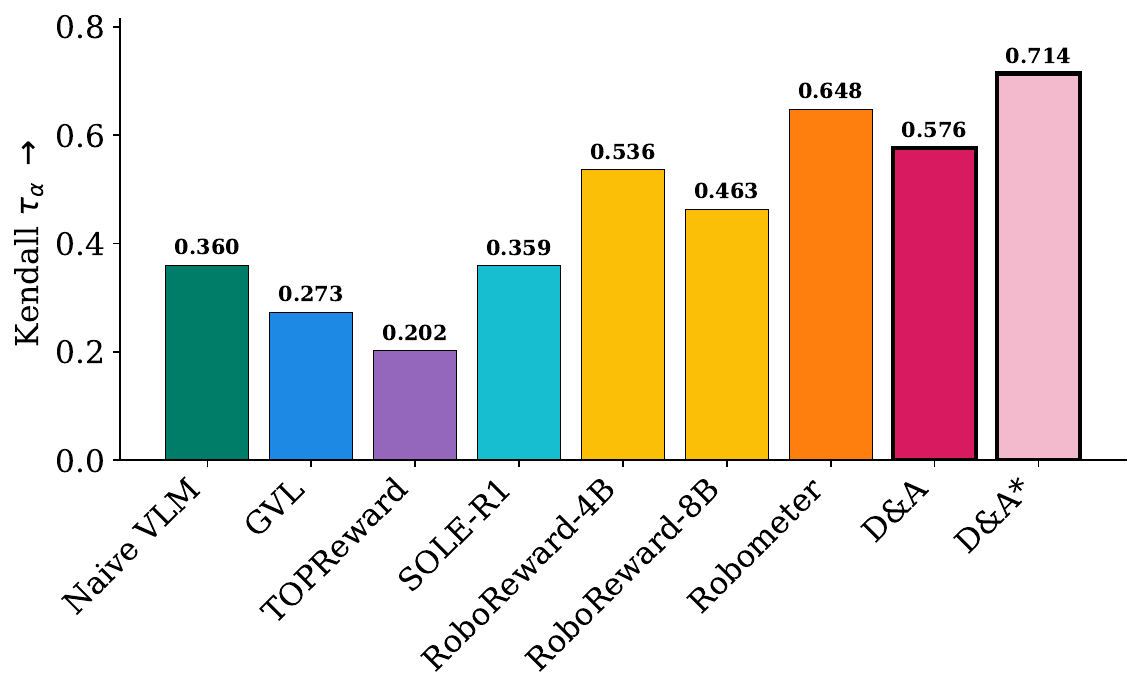}}
        % \raisebox{0.47em}{\includegraphics[width=\linewidth]{figures/kendall_policy_ranking.png}}
        \caption{\texttt{RBM-EVAL-OOD} ($\uparrow$)
        % gpt sol 5.6 dna 0.662 naive 0.690 
        }
        \label{fig:rbm_eval_ood}
    \end{subfigure}
    \hfill
    \begin{subfigure}[b]{0.27\linewidth}
        \centering
        \raisebox{-2em}{\includegraphics[width=\linewidth]{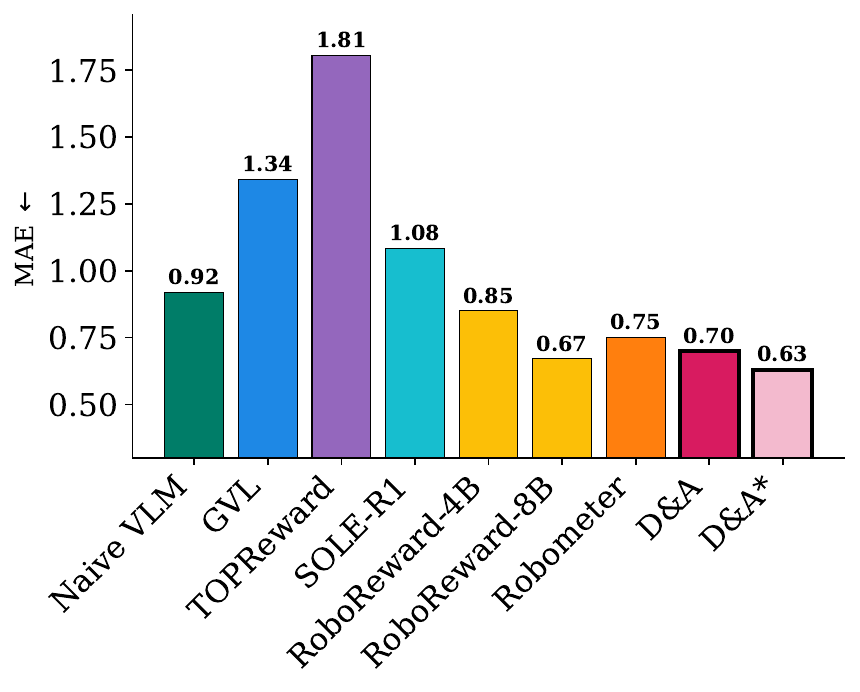}}
        % \raisebox{-2em}{\includegraphics[width=\linewidth]{figures/roboreward_mae_results.png}}
        \caption{\texttt{RoboRewardBench} ($\downarrow$)
        }
        \label{fig:roboreward}
    \end{subfigure}
    \caption{\textbf{Results on \texttt{D\&A-Bench}, \texttt{RBM-EVAL-OOD}~\cite{liang2026robometer}, and \texttt{RoboRewardBenchmark}~\cite{lee2026roboreward}.} \Method (Gemini 3 Flash) emerges as the strongest zero-shot method, outperforming (\soler) and matching the performance (\robometer/ \roboreward) of post-trained models. Scaling up the base model, \Methodalpha (GPT-6 Astra), tops every benchmark.}
    \label{fig:offline_benchmarks}
\end{figure*}

\begin{figure*}[tb]
    \centering
    \begin{subfigure}[b]{0.32\linewidth}
        \centering
        \includegraphics[width=\linewidth]{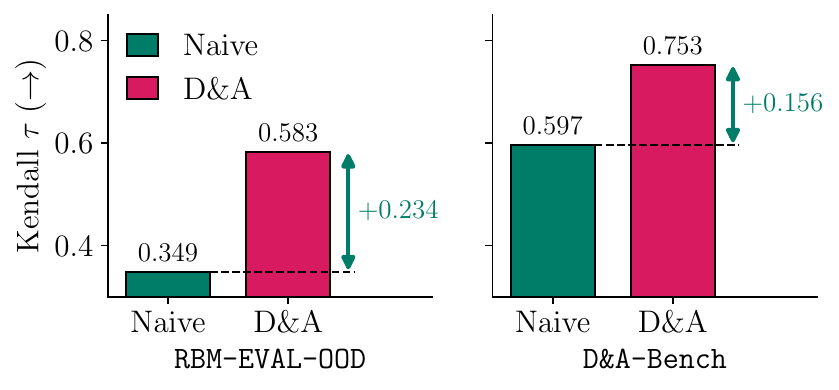}
        \caption{Qwen3.6-Plus}
        \label{fig:dna_gains_qwen_main}
    \end{subfigure}
            \hfill
    \begin{subfigure}[b]{0.32\linewidth}
        \centering
        \includegraphics[width=\linewidth]{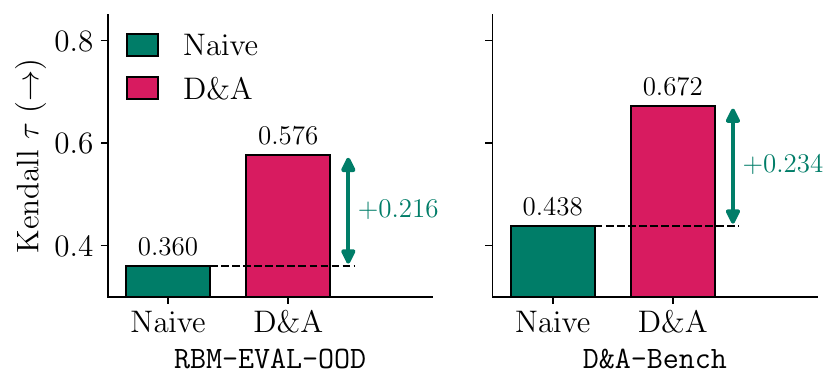}
        \caption{Gemini 3 Flash}
        \label{fig:dna_gains_3_flash_main}
    \end{subfigure}
    \hfill
    \begin{subfigure}[b]{0.32\linewidth}
        \centering
        \includegraphics[width=\linewidth]{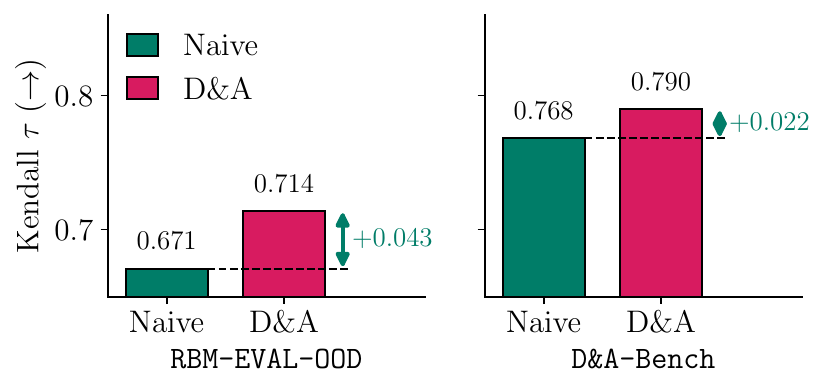}
        \caption{GPT-6 Astra}
        \label{fig:dna_gains_astra_main}
    \end{subfigure}
    \label{fig:dna_gains_three}
    \caption{\textbf{Relative gains of \method over Naive VLM across various base models on \texttt{RBM-EVAL-OOD} and \texttt{D\&A-Bench}.}}
\end{figure*}

Our evaluation aims to address the following questions:

\begin{enumerate}[label=\textbf{(Q\arabic*)} ,noitemsep ]
    \item \label{q1} \textit{Effectiveness}: Is \method an effective reward model for real-world offline and online RL?
    \item \label{q2} \textit{Generalization}: How well does \method generalize to different tasks and embodiments? % benchmarks
    \item \label{q3} \textit{Scaling}: How does \method scale with the underlying VLM?
    \item \label{q4} \textit{Ablation and analysis}: What components contribute to \method's performance?
    \item \label{q5} \textit{Rich feedback}: Can rich feedback generated by \method improve over scalar rewards?
\end{enumerate}

We evaluate \Method against the six strongest-performing approaches, spanning both explicitly post-trained and zero-shot reward models to provide a comprehensive analysis of the current landscape. If not denoted otherwise, we choose Gemini 3 Flash as a base model for \Method to balance cost and performance, but also report benchmark numbers for GPT-6 Astra (\Methodalpha) to demonstrate capability scaling.

\noindent\textbf{Post-trained Baselines} are explicitly trained on robotics datasets to predict progress. The following approaches all use Qwen-3-VL~\cite{bai2025qwen3} as the base model.
\roboreward~\cite{lee2026roboreward} fine-tunes to predict discrete end-of-episode scores from 1 to 5, trained on large-scale robotic data~\cite{collaboration2023open,atreya2025roboarena} augmented with counterfactual language instructions that simulate failed or mismatched trajectories.
\robometer~\cite{liang2026robometer} fine-tunes on RBM-1M (optimal and suboptimal data) with a multi-objective framework that jointly predicts continuous progress in $[0,1]$, binary success, and pairwise trajectory preferences.
\soler~\cite{schroeder2026sole} fine-tunes on CoT~\cite{wei2022chain} and continuous progress estimates and then uses reinforcement learning from verifiable rewards (RLVR) to further tune the model.

\begin{figure*}[tb]
    \centering
    \includegraphics[width=\linewidth]{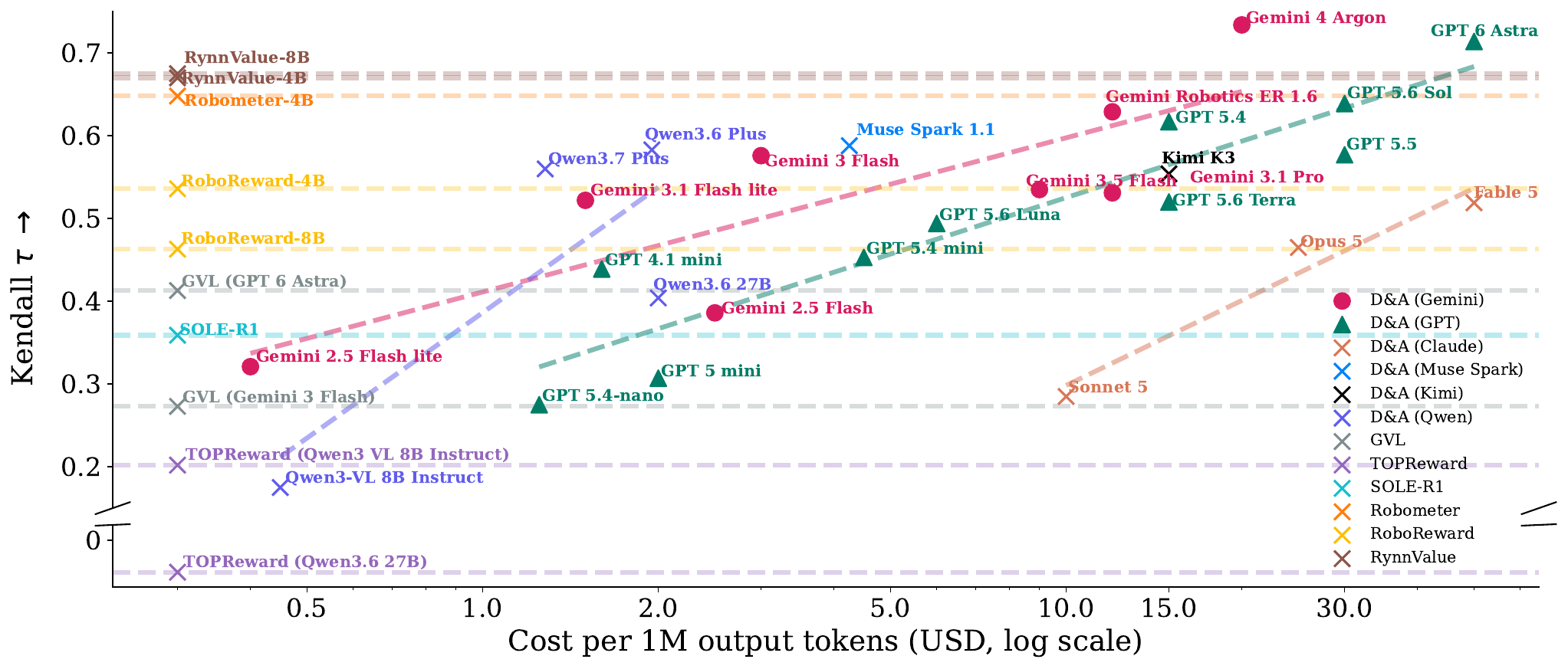}
    \caption{\textbf{\method scales with model capability across four open- and closed-source model families.} \gvl scales much worse than \Method, while \topreward and \roboreward show an inverse correlation between performance and model capability.
    % Baselines are considered ``free''.
    % * denotes average across 3 seeds.
    }
    \label{fig:scaling}
\end{figure*}

\noindent\textbf{Zero-shot Baselines} leverage the world knowledge of off-the-shelf VLMs without any task- or robotics-specific fine-tuning.
\naive models the mapping $R : (o_{1:T},\, l) \mapsto r_{1:T}$ with a single VLM call, predicting progress without decoupling or success-criteria reasoning.
\gvl~\cite{ma2024vision} prompts a VLM with temporally shuffled frames, forcing it to judge each frame's intrinsic progress independently of chronological order to counter the VLM's bias toward time continuity.
\topreward~\cite{chen2026topreward} prompts a VLM with the full video and instruction and, rather than emitting progress as text, extracts it from the token logit of the ``True'' response to whether the instruction matches the execution.
For fair comparison, we select comparable models for \topreward (Qwen-VL 8B) and \gvl (Gemini 3 Flash), and report their scaling performance on better models in~\Cref{fig:scaling} for \ref{q3}.

\begin{figure*}[tb]
    \centering
    \includegraphics[width=0.62\linewidth]{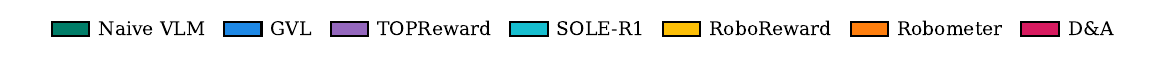}\\[2pt]
    \vspace{-0.5em}
    \begin{subfigure}[b]{0.49\linewidth}
        \centering
        \includegraphics[width=\linewidth]{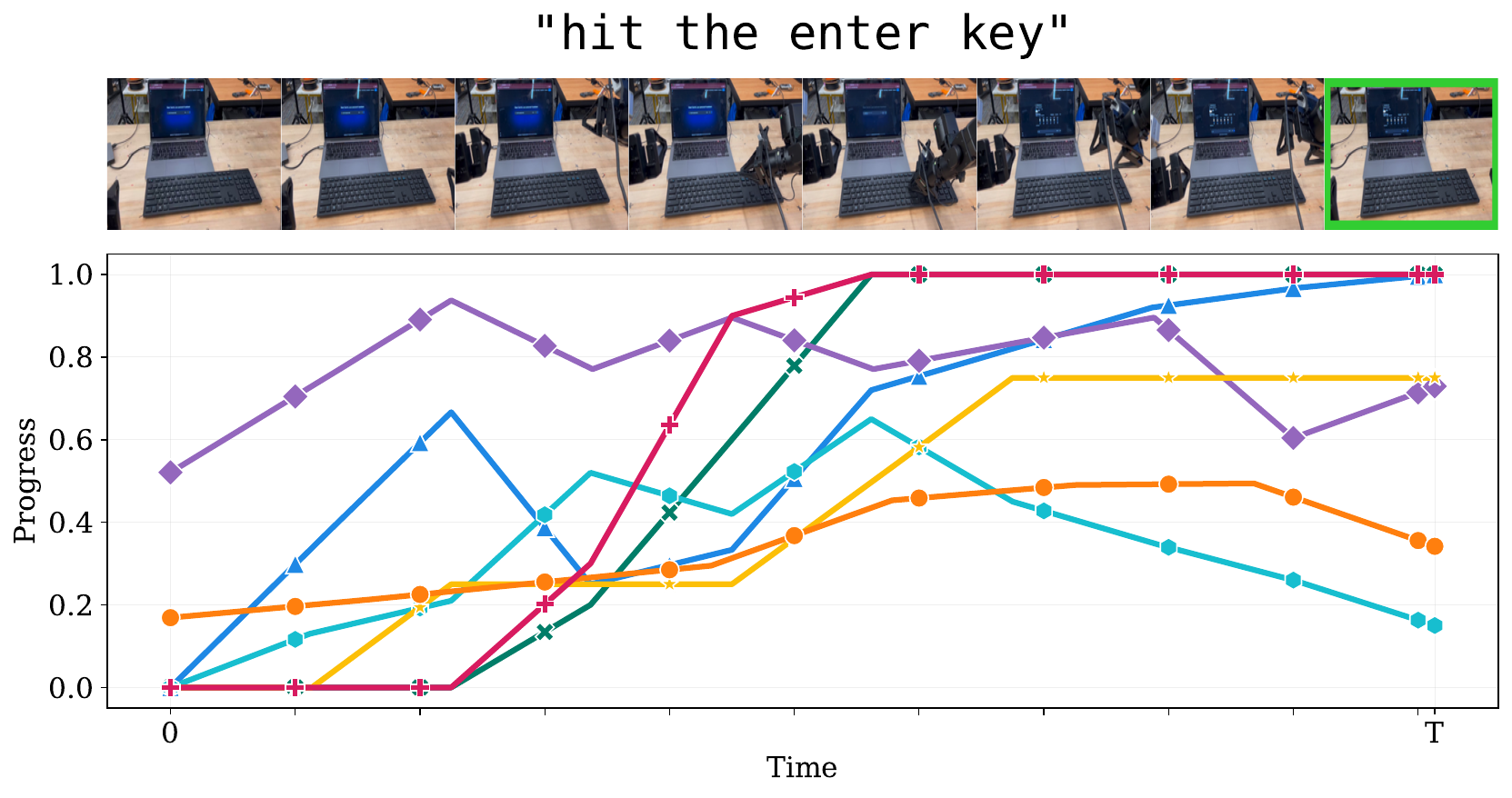}
        \caption{Success: the robot presses enter.}
        \label{fig:qual_enter_success}
    \end{subfigure}
    \hfill
    \begin{subfigure}[b]{0.49\linewidth}
        \centering
        \includegraphics[width=\linewidth]{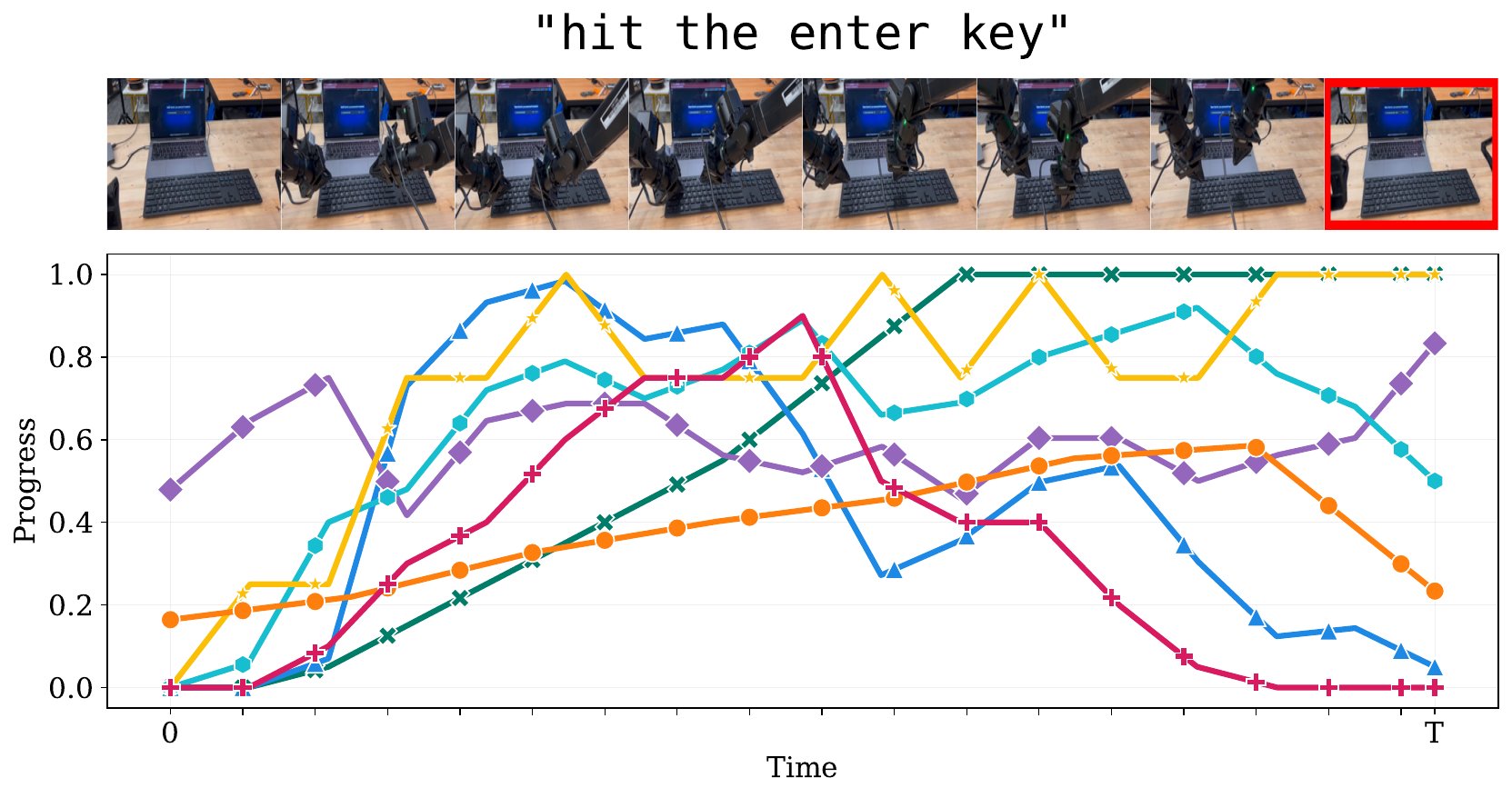}
        \caption{Failure: the robot pressed a different key.}
        \label{fig:qual_enter_failure}
    \end{subfigure}
    \label{fig:qualitative_main}
    \caption{\textbf{Qualitative \texttt{D\&A-Bench} predictions for ``hit the enter key'':} Completion (an on-screen change) is not visible in the final frame (\textcolor{green!60!black}{green}: success, \textcolor{red}{red}: failure). \Method separates success ($r_T\!\to\!1$) from the visually similar failure ($r_T\!\to\!0$), while \robometer scores both alike, \roboreward/\soler score the failure as a success, and \gvl/\topreward score both high. In the failure case, \Method captures visual nuance of the gripper close to the correct key without pressing it.}% More examples in \Cref{fig:qualitative_appendix}.\vspace{-1em}}
\end{figure*}

\subsection*{\ref{q1} Is \method an effective reward model for real-world offline and online RL?}

\noindent\textbf{Online RL.} We apply DSRL~\citep{wagenmaker2025steering} to the $\pi_{0}$~\citep{black2024pi_0} policy fine-tuned on DROID~\citep{khazatsky2024droid}, using the \method{} progress score directly as the dense reward. We evaluate on ``put the fruit in the bowl'' in a DROID kitchen setup (\cref{fig:setup_dsrl}, left), where the $\pi_{0}$ base model keeps grasping the ice cream cone and initially solves the task only $10\%$ of the time; learning the correct behavior via RL requires rewards that discriminate good from bad state transitions. \Method provides this signal, assigning zero reward to ice-cream-cone trajectories and high, dense reward for manipulating the banana, raising the success rate to $\mathbf{100\%}$ in ${\sim}90$ minutes of wallclock time. In contrast, \topreward assigns high reward to both successful and failed trajectories, leaving success at only $5\%$.

\noindent\textbf{Offline RL.} We condition $\pi_{0.5}$~\citep{intelligence2025pi_} on advantages computed from \Method rewards via CFGRL/RECAP~\citep{frans2025diffusion, intelligence2025pi}, centering returns against a global value baseline estimated over the full dataset so that advantages rank state-action pairs against the dataset-wide distribution of outcomes. We collect 50 optimal, 50 suboptimal, and 50 failure demonstrations on a bimanual YAM (\cref{fig:setup_dsrl}) for the task ``move the red book from the top shelf to the gray box'', where failures include using the wrong box or arm, or dropping the book mid-transfer. We LoRA-fine-tune $\pi_{0.5}$ for 50k steps. The baseline $\pi_{0.5}$ fine-tuned on the successful and suboptimal data only achieves $20\%$ success, reproducing many suboptimal behaviors. \topreward and \robometer leverage the suboptimal data better, but their noisy rewards fail to fully filter out suboptimality; \Method instead filters these behaviors, improving BC by $\mathbf{+80\%}$ and outperforming all other models.

\textbf{Advantage-filtered Behavior Cloning.}
We collect 50 optimal and 100 suboptimal demonstrations for ``nudge out the book `Thinking, Fast and Slow', grasp it, and place it on the table'' on a bimanual YAM. Suboptimal demonstrations include nudging or pulling out different books but eventually complete the task.
We rank all chunks in the dataset by their advantage $A_t$ with a single dataset-wide threshold $\tau$, so that chunks are compared against the full distribution of outcomes rather than within their own episode.
We LoRA-fine-tune $\pi_{0.5}$ for 25k steps on the top 35th percentile of chunks ($A_t\ge\tau$).
\robometer and \topreward both perform poorly, struggling with the long horizon of the task by assigning similar scores to optimal and suboptimal action chunks.
Trained on the full dataset, $\pi_{0.5}$ is competitive but frequently performs suboptimal actions, leading to competitive performance but slow execution.
On success, \method completes the task faster than the baseline fine-tuned $\pi_{0.5}$, with $\mathbf{18.8\pm1.8}$ and $28.5\pm9.2$ seconds on average, and with markedly lower variance across rollouts. This shows that \method filters out suboptimal and retains optimal action chunks.
% $565\pm56$ versus $856\pm277$ timesteps

\subsection*{\ref{q2} How well does \method generalize to different tasks and embodiments?}

% benchmarks
To evaluate \Method across tasks, embodiments, and language, we evaluate on three offline benchmarks: \texttt{D\&A-Bench}, a \emph{generalization} benchmark we collected manually that is guaranteed to be out-of-distribution to all models to stress-test fine-grained scene and language understanding, and two external benchmarks \texttt{RBM-EVAL-OOD}~\cite{liang2026robometer} and \texttt{RoboRewardBenchmark}~\cite{lee2026roboreward}.
We propose \texttt{D\&A-Bench} to evaluate nuanced scene understanding (e.g., ``dose exactly 16\,g of coffee beans''), counting (``build a 5 block tower''), and semantic reasoning (``assemble the car''). The benchmark comprises 54 (27 success, 27 failure) trajectories over 14 YAM tasks each annotated with 10 VLM-generated paraphrases of the instruction to test language robustness. We report the Kendall $\tau_\alpha$ coefficient~\cite{muslimani2025towards} for \texttt{D\&A-Bench} and \texttt{RBM-EVAL-OOD}, measuring how well a reward model's final-step prediction align with the ground-truth ordering of optimal, sub-optimal, and failure trajectories ($\in[-1,+1]$; higher is better). \texttt{RoboRewardBenchmark} uses discrete end-of-episode progress labels reports Mean Absolute Error (MAE; lower is better).

Across all three benchmarks (\cref{fig:offline_benchmarks}), \Method closes the gap to post-trained models, outperforming \roboreward on \texttt{RBM-EVAL-OOD}, \robometer on \texttt{RoboRewardBenchmark}, and \soler on both. On \texttt{D\&A-Bench}, \Method leverages the generalization and reasoning capabilities of its base VLM, scoring highest ($0.67$) and beating the best post-trained models \robometer ($0.56$) and \roboreward ($0.43$). Zero-shot \topreward ($0.24$) and \gvl ($0.27$) fail to generalize to those challenging tasks. \method improves $+0.23$ over \naive, which here is a surprisingly strong baseline: because the benchmark rewards scene and language generalization, even unstructured prompting of a strong base VLM generalizes far better than \gvl and \topreward, which cannot reliably compare trajectories. \Cref{fig:qualitative_main} illustrates this on ``hit the enter key,'' whose completion is only seen by a subtle on-screen change: \Method tracks the full trajectory and cleanly separates the success from a visually similar failure, whereas \robometer assigns the two nearly the same reward, \roboreward/\soler score them oppositely, and zero-shot \naive/\gvl/\topreward score both high. More qualitative examples in \Cref{sec:appdx:dnabench}.
Finally, we demonstrate the scalability of \Method by using GPT-6 Astra as a base model (\Methodalpha), achieving SOTA results on all three benchmarks.

\begin{figure}[tb]
% \begin{wrapfigure}{r}{0.5\linewidth}
% \vspace{-1.5em}
    \centering

    \includegraphics[width=\linewidth]{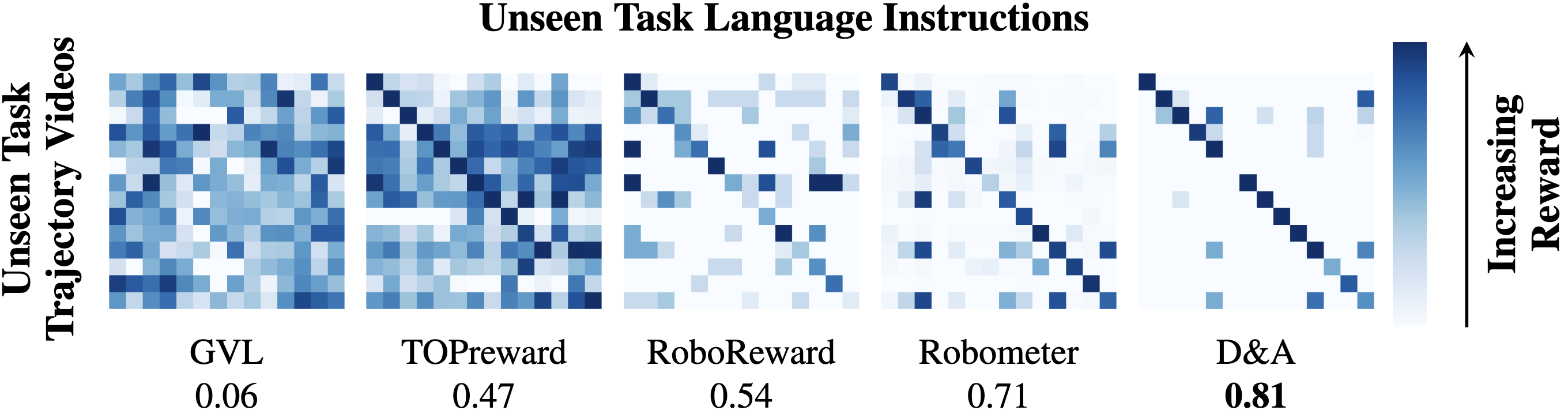}
    
    \vspace{1em}
    
    \includegraphics[width=\linewidth]{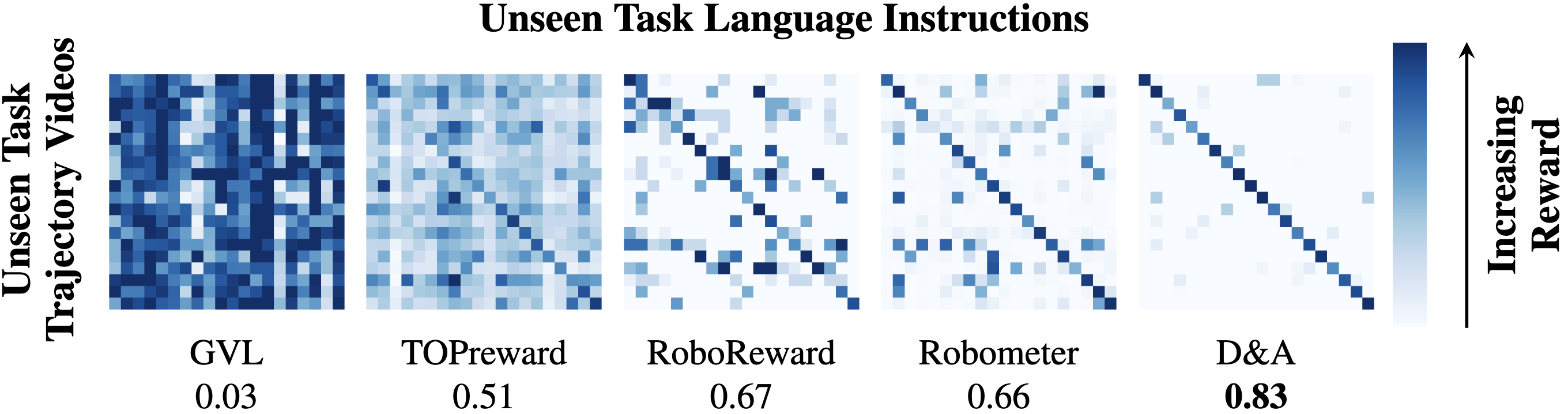}
    
    % \vspace{-1em}
    %\caption{\textbf{Video--language reward confusion matrices} over all video--instruction pairings from \texttt{RBM-EVAL-OOD}. \method is distinctly diagonal-heavy, aligning demonstrations with their instructions; the reported value is the column-normalized diagonal mean (fraction of reward on correct matches).\vspace{-1em}}
    \caption{\textbf{Video--language reward confusion matrices} over all video--instruction pairings from \texttt{D\&A-Bench} (top) \texttt{RBM-EVAL-OOD} (bottom). The reported value is the column-normalized diagonal mean (fraction of reward on correct matches).\vspace{-1em}}
    \label{fig:confusion_matrices}
    % \vspace{-1em}
% aliangdw_usc_franka_policy_ranking_usc_franka_policy_ranking, jesbu1_utd_so101_clean_policy_ranking_top_utd_so101_clean_policy_ranking_top, ykorkmaz_usc_trossen_rfm_usc_trossen
% \end{wrapfigure}
\end{figure}

% confusion matrix
To evaluate semantic understanding, we plot the confusion matrices between sets of unseen task trajectory videos (rows) and unseen task language instructions (columns) in \cref{fig:confusion_matrices}, where {\color[HTML]{083877}{\rule{0.7em}{0.7em}}} means the model assigned high and {\color[HTML]{CADDF0}{\rule{0.7em}{0.7em}}} low reward to the combination.
% NOTE: confusion matrix datasets adapted from robometer to ensure full diagonal is possible
A reward model should discriminate mismatched video–instruction pairs while predicting high reward on the diagonal (correct matches). \Method consistently assigns zero progress to mismatched pairs while accurately scoring correct matches; post-trained models also separate the pairs, whereas zero-shot models cannot, reflecting their bias toward non-zero progress.

\begin{figure*}[tb]
    \centering
            \centering
        \includegraphics[width=\linewidth]{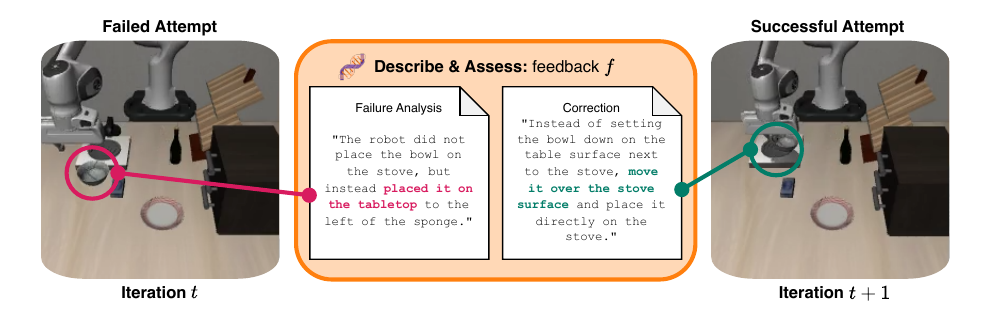}
        \label{fig:cap}
    \caption{\textbf{CaP with rich feedback.} Example of iterative improvement with \Method feedback in LIBERO.}
\end{figure*}

\subsection*{\ref{q3} How does \method scale with the underlying VLM?}

We investigate how \Method's ability to predict progress from video scales with base model capability. In \cref{fig:scaling}, we plot trajectory ranking performance (Kendall $\tau_\alpha$ on \texttt{RBM-EVAL-OOD}) against the cost per million tokens, using cost as a proxy for model capability. Baselines are shown as dashed horizontal lines.
Across all base model families, we observe a clear upward trend: as \Method's base model capability increases, so does the ability to discriminate trajectory optimality. 
\gvl also scales with the underlying model but at a much lower rate.
\topreward faces a major barrier on closed-source models through its reliance on token logprobs---which are unavailable for Gemini 3---and degrading performance under mandatory chat templates enforced by proprietary platforms. Our results further follow the negative scaling trend of \topreward observed in the original publication \cite{chen2026topreward}.
\Method with Gemini 4 Argon and GPT-6 Astra as a base model achieves SOTA performance on the benchmark with $0.734$ and $0.714$, validating the observed trend and efficacy of our approach.

To disentangle the base model's capability from \Method's gains, we compare \naive and \Method on different base models in \Cref{fig:dna_gains_three}.
On smaller models (Gemini 3 Flash, Qwen3.6-Plus), we observe the biggest improvements through \Method, outperforming \roboreward on both benchmarks and \robometer on \texttt{D\&A-Bench}.
Consequently, \Method enables competitive reward modeling in small open-source and closed-source models.
\Method applied to GPT-6 Astra leads to smaller but still substantial gains in reward modeling capabilities. Qualitative examples in \Cref{sec:dna_gains_frontier}.

% \begin{wraptable}{R}{0.3\textwidth}
%     \vspace{-1em}
%     \centering
%     \resizebox{\linewidth}{!}{% 
%     \begin{tabular}{lc}
%     \toprule
%     \textbf{RBM-1M-OOD} & \textbf{Kendall $\tau_\alpha$} \\
%     \midrule
%     naive VLM baseline & 0.295 \\
%     + success criteria (\cref{sec:method:assessor}) & 0.477 \\
%     + decoupling (\cref{sec:method:describer}) & 0.419 \\
%     + grounding (\method) & \textbf{0.562} \\
%     % w/o decomposition & 0.460 \\
%     % w/o grounding     & 0.422 \\
%     % \textbf{METHOD}   & 0.538 \\
%     \bottomrule
%     \end{tabular}}
%     \caption{\textbf{Ablations.}\todo{replace with full ablation table}}
%     \label{tab:ablation_study}
%     \vspace{-1em}
% \end{wraptable}
% \begin{wraptable}{R}{0.25\textwidth}
\begin{table}[tb]
    \centering
    \resizebox{\linewidth}{!}{% 
    \begin{tabular}{cccc}
    \toprule
    \shortstack{\small\textbf{Success Criteria}\\ \small(\cref{sec:method:assessor})} & 
    \shortstack{\small\textbf{Decoupling}\\ \small(\cref{sec:method:describer})} & 
    \shortstack{\small\textbf{Grounding}\\ \small(\cref{sec:method:describer})} & 
    \shortstack{\small\textbf{Kendall $\tau_\alpha$} $\uparrow$} \\
    \midrule
    \xmark & \xmark & \xmark    & 0.486 \\
\midrule
    \cmark & \xmark & \xmark    & 0.561 \\
\midrule

    \xmark & \cmark & \xmark    & 0.343 \\
    % \xmark & \cmark & Task      & 0.557 \\
    \xmark & \cmark & \cmark    & 0.546 \\
\midrule
    
    \cmark & \cmark & \xmark    & 0.457 \\
    % \cmark & \cmark & Task     & 0.538 \\
    \cmark & \cmark & \cmark   & \textbf{0.626} \\
    \bottomrule
    \end{tabular}
    }
    \caption{\textbf{Ablations on \texttt{RBM-EVAL-OOD}.}
    % \vspace{-1em}
    }
    \label{tab:ablation_study}
\end{table}
% \end{wraptable}

\subsection*{\ref{q4} What components contribute to \method's performance?}
To understand how each design decision contributes to the performance of \Method, we perform a series of ablations in \Cref{tab:ablation_study} using Gemini 3.7 Flash. Introducing an explicit success criterion to \naive improves performance by $+0.18$ and closes the gap between zero-shot and post-trained methods. Combining the success criteria with decoupling the description and assessment stages, negatively impacts performance due to task-relevant objects not being identifiable or being missing completely from the language description.
When grounding the description in the objects present in the video, the performance improves by $+0.26$ over the \naive, outperforming post-trained models in some cases.

\subsection*{\ref{q5} Can rich feedback generated by \method improve over scalar rewards?}

We evaluate \Method as a source of rich feedback to improve Code-as-Policies (CaP)~\cite{liang2022code} on the LIBERO benchmark (goal suite)~\cite{liu2023libero}. The CaP planner takes multiple attempts at solving the task, generating code and executing it in the environment. We compare a history of code and binary success to extending it with rich feedback generated by \Method.
Allowing CaP to iterate over its solutions by giving it history improves performance from $11.4\%$ (no history) to $40.6\%$ (history). Adding \Method's rich language feedback achieves $51.4\%$ ($p=0.00076$ over history baseline), showcasing how rich semantic feedback provides a more effective signal than scalar rewards for downstream refinement.
See \Cref{fig:cap} for an example of feedback improving the behavior.

\section{Limitations and Discussion}
\label{sec:limitations}
In this work, we introduce \method, a methodology for leveraging frontier models in reward modeling and rich feedback. We counter confirmation bias and address hallucinated success by decoupling description and assessment. We demonstrate the efficacy of \method on several real-world robotic manipulation tasks. 
While effective, our approach introduces several limitations. First, it inherits the systemic constraints of off-the-shelf frontier models, including high inference latency, token cost, and restricted context windows. Because these models are typically accessed via closed-source APIs, they introduce stochasticity and limited granular control. They also remain susceptible to occasional semantic hallucinations, which can degrade description fidelity and subsequent reward assessment.
Second, the generated reward is bounded by the granularity of natural language descriptions. This poses challenges in domains requiring highly nuanced or continuous reward structures that are difficult to verbalize, such as precise spatial relationships or absolute geometric distances.
Despite these challenges, \method marks a pivotal shift from fine-tuning to zero-shot robotics reward modeling that scales with rapid advances in frontier VLMs.

\section*{Acknowledgments}
We thank Marc Toussaint and the Learning and Intelligent Systems lab at TU Berlin for the initial discussions that eventually led to this project.
Part of this work was supported by TRI U3.0, Ai2, and the Cross-Pacific AI Initiative (X-PAI), which is supported by Amazon, NVIDIA, the University of Washington, and the University of Tsukuba.

% Generative AI tools and technologies, such as ChatGPT, may not be listed as authors of a paper submitted to ICRA 2027. The use of content generated by AI in an article (including but not limited to text, figures, images, and code) must be disclosed in the acknowledgments section. The AI system used must be identified, and specific sections of the article that use AI-generated content should be specified and accompanied by a brief explanation regarding the manner in which the AI system was used to generate the content. The use of AI systems for editing and grammar enhancement is common practice and, as such, is generally outside the intent of the above policy.
% https://2027.ieee-icra.org/contribute/call-for-icra-2027-papers-now-accepting-submissions/

\newpage
\printbibliography

\onecolumn
\section{Offline RL via Advantage Conditioning: Full Derivation}
\label{sec:appdx:offline_rl}

We provide the full derivation of the advantage-conditioned offline RL pipeline summarized in \Cref{sec:method:policy}. Unlike standard behavior cloning (BC) or filtered BC, which are strictly constrained by the average performance of the training data, RL can yield a better-than-demonstrator policy.

Suppose we are provided an offline dataset $\mathcal{D} = \{\tau_n\}_{n=1}^N$ containing $N$ trajectories. Each trajectory $\tau_\alpha$ is a sequence of observations and actions, $\tau = \{(o_1, a_1), (o_2, a_2), \dots, (o_T, a_T)\}$. Recall that we treat the progress scores $r_t = R(o_{1:T}, l)$ as a potential function and define the step-wise potential-based reward as the temporal difference $\tilde{r}_t = r_t - r_{t-1}$ with $r_0 = 0$, which effectively captures the instantaneous rate of task improvement at each step.

Because we operate in an offline setting, we can compute empirical state values and advantages entirely offline without the stability risks or compounding errors of bootstrapping. For any timestep $t$ within a trajectory $\tau_\alpha$, the discounted Monte Carlo return is calculated as $G_t = \sum_{k=t}^{T} \gamma^{k-t} \tilde{r}_k$, where $\gamma \in [0, 1)$ is the discount factor. To isolate how much better a specific sequence of actions is relative to the entire dataset, we establish a global value baseline $V$ as the empirical mean of the discounted Monte Carlo returns across all timesteps and all trajectories in $\mathcal{D}$:
\begin{equation*}
    V = \frac{1}{\sum_{n=1}^N T_n} \sum_{n=1}^N \sum_{t=1}^{T_n} G_{t,n}.
\end{equation*}
Computing $V$ over the full dataset (rather than per-trajectory or per-task) lets the resulting advantages rank state-action pairs against the dataset-wide distribution of outcomes, which is what CFGRL conditions on. Using this global baseline, the offline advantage $A(s_t, a_t)$ for each state-action pair is computed as the centered return: $A(s_t, a_t) = G_t - V$.

These computed advantages can be directly used for downstream offline RL algorithms. We leverage the CFGRL framework~\citep{frans2025diffusion} to scale actions or action tokens by training a conditional policy $\pi_\theta(a_t \mid s_t, l, g)$ alongside an unconditional or baseline variant, utilizing the advantage as an explicit conditioning variable $g$ during optimization. Following RECAP~\citep{intelligence2025pi}, we adapt this advantage-guided framework directly to a $\pi_{0.5}$~\citep{intelligence2025pi_} and optimize the model via a standard supervised log-likelihood objective: $\mathcal{L}(\theta) = -\mathbb{E}_{(s_t, a_t, l) \in \mathcal{D}} \left[ \log \pi_\theta(a_t \mid s_t, l, g) \right]$.

\section{Online RL via Latent Space Noise Steering: Full Derivation}
\label{sec:appdx:online_rl}

We provide the full optimization details for the online DSRL pipeline, summarized in \Cref{sec:method:policy}. Unlike the offline setting, which benefits from static, dataset-wide baseline calculations, online policy learning requires the agent to continually explore and optimize its behavior through active environment interaction. DSRL~\citep{wagenmaker2025steering} learns a policy $\pi_\theta(w \mid s, l)$ that generates a lower-dimensional latent steering vector or noise variable $w$, which is then mapped through the frozen base model to produce the final robot actions, rather than optimizing actions directly in the high-dimensional action space.

Under this paradigm, the agent interacts with the environment online, and the absolute task progress score $r_t = R_{1:T}(o_{1:T}, l)$ generated by \method{} is used directly as the dense environmental reward. To optimize the policy over latent space interactions, we maintain a soft action-value function $Q_\phi(s, w, l)$ and a target state-value network $V_{\bar{\psi}}(s, l)$. The critic parameters $\phi$ are optimized by minimizing the standard temporal-difference (TD) error:
\begin{equation}
    \mathcal{L}(\phi) = \mathbb{E}_{(s_t, w_t, r_t, s_{t+1}) \sim \mathcal{B}} \left[ \Big( Q_\phi(s_t, w_t, l) - \big( r_t + \gamma V_{\bar{\psi}}(s_{t+1}, l) \big) \Big)^2 \right].
\end{equation}
Concurrently, the policy parameters $\theta$ are updated to maximize the expected future return from the critic alongside an entropy regularization term:
\begin{equation}
    \mathcal{L}(\theta) = -\mathbb{E}_{s_t \sim \mathcal{B}, \, w_t \sim \pi_\theta} \left[ Q_\phi(s_t, w_t, l) - \alpha \log \pi_\theta(w_t \mid s_t, l) \right].
\end{equation}
By mapping this online optimization loop directly onto the latent noise space of a frozen expert, this framework mitigates catastrophic forgetting and policy collapse, enabling sample-efficient online behavior optimization powered by \method{} rewards.

\newpage
\section{Additional Results}

\begin{table*}[h!]
    \centering

    \begin{tabular}{llc}
    \toprule
    \textbf{Benchmark} & \textbf{Model} & \textbf{Score} \\
    \midrule
    \texttt{RoboRewardBenchmark} & \Method (GPT-6-Astra) & \textbf{0.63} \\
                                 & \naive (GPT-5.5 (2026-04-23)~\cite{lee2026roboreward})$^*$ & 0.66 \\
                                 & \naive (GPT-5.5 xhigh reasoning)~\cite{lee2026roboreward} & \textbf{0.63} \\
    \midrule
    \texttt{D\&A-Bench}           & \Method (GPT-6-Astra) & \textbf{0.790} \\
                                 & \Method (Gemini 3 Flash) & 0.672 \\
    \midrule
    \texttt{RBM-EVAL-OOD}        & \Method (GPT-6-Astra) & \textbf{0.714} \\
                                 & \robometer & 0.648 \\
    \bottomrule
    \end{tabular}
    \captionof{table}{\textbf{\Method (GPT-6 Astra) vs. best reported on benchmark.}}
    \label{tab:sota_benchmark_evals}

    \vspace{1em}

    \begin{tabular}{cccccc}
    \toprule
    \naive & \gvl & \topreward & \roborewardfourb & \robometer & \Method \\
    \midrule
     0.92 & 1.34 & 1.80 & 0.85 & 0.75 & 0.70 \\
    \bottomrule
    \end{tabular}
    \captionof{table}{\texttt{RoboRewardBenchmark} (MAE $\downarrow$)}
    \label{tab:additional0}

    \vspace{1em}

    \begin{tabular}{lcccccc}
    \toprule
     & \naive & \gvl & \topreward & \roborewardfourb & \robometer & \Method \\
    \midrule
    \texttt{RBM-EVAL-OOD} & 0.950 & 0.855 & 0.796 & 0.842 & 0.934 & 0.885 \\
    \texttt{D\&A-Bench}  & 0.962 & 0.014 & 0.490 & 0.780 & 0.945 & 0.914 \\
    \bottomrule
    \end{tabular}
    \captionof{table}{\textbf{Additional results} (VOC $\uparrow$)}
    \label{tab:additional1}

    \vspace{1em}

\end{table*}

\section{Gains of \method over Base Model Capabilities - Frontier Models}
\label{sec:dna_gains_frontier}

\begin{figure*}[h!]
    \centering
    \begin{subfigure}[b]{0.23\linewidth}
        \centering
        \includegraphics[width=\linewidth]{figures/bar_qwen_3.6_plus.pdf}
        \caption{Qwen3.6-Plus}
        \label{fig:dna_gains_qwen}
    \end{subfigure}
            \hfill
    \begin{subfigure}[b]{0.23\linewidth}
        \centering
        \includegraphics[width=\linewidth]{figures/bar_gemini_3_flash.pdf}
        \caption{Gemini 3 Flash}
        \label{fig:dna_gains_3_flash}
    \end{subfigure}
    \hfill
    \begin{subfigure}[b]{0.23\linewidth}
        \centering
        \includegraphics[width=\linewidth]{figures/bar_gpt-6_astra.pdf}
        \caption{GPT-6 Astra}
        \label{fig:dna_gains_astra}
    \end{subfigure}
        \hfill
    \begin{subfigure}[b]{0.23\linewidth}
        \centering
        \includegraphics[width=\linewidth]{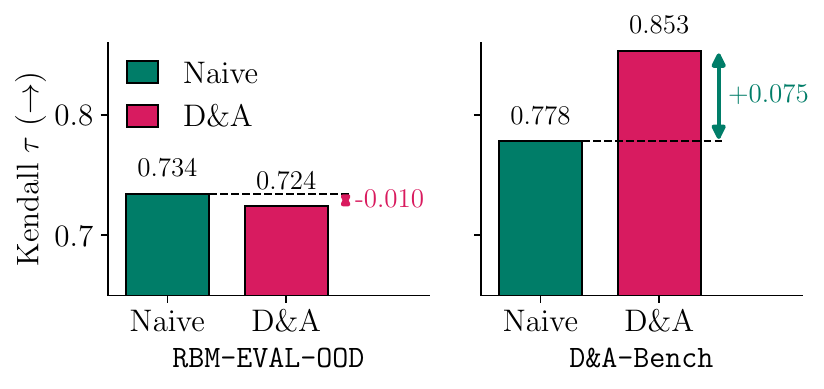}
        \caption{Gemini 4 Argon}
        \label{fig:dna_gains_argon}
    \end{subfigure}
    \label{fig:dna_gains_four}
    \caption{\textbf{\method gains on frontier models}}
\end{figure*}

On smaller models (Gemini 3 Flash, Qwen3.6-Plus), we observe the biggest improvements through \method, outperforming \roboreward on both benchmarks and \robometer on \texttt{D\&A-Bench} (\Cref{fig:dna_gains_qwen} and \Cref{fig:dna_gains_3_flash}).
On more recent VLMs, \method improves reward modeling capabilities, leading to SOTA results on both benchmarks (\Cref{fig:dna_gains_astra} and \Cref{fig:dna_gains_argon}).
The only exception is Gemini 4 Argon on \texttt{RBM-EVAL-OOD}, where \method is on par with \naive. Given \method's significant improvement on \texttt{D\&A-Bench}, we hypothesize that Gemini 4 Argon was trained on the \texttt{RBM-EVAL-OOD} dataset (Google specifically advertises Gemini 4 Argon as a value function\footnote{\url{https://developers.googleblog.com/gemini-4-argon-for-robotics/}}), diluting the benchmark results and preventing \method from improving.

\begin{figure*}[h!]
    \centering
    \begin{subfigure}[b]{0.45\linewidth}
        \centering
        \includegraphics[width=\linewidth]{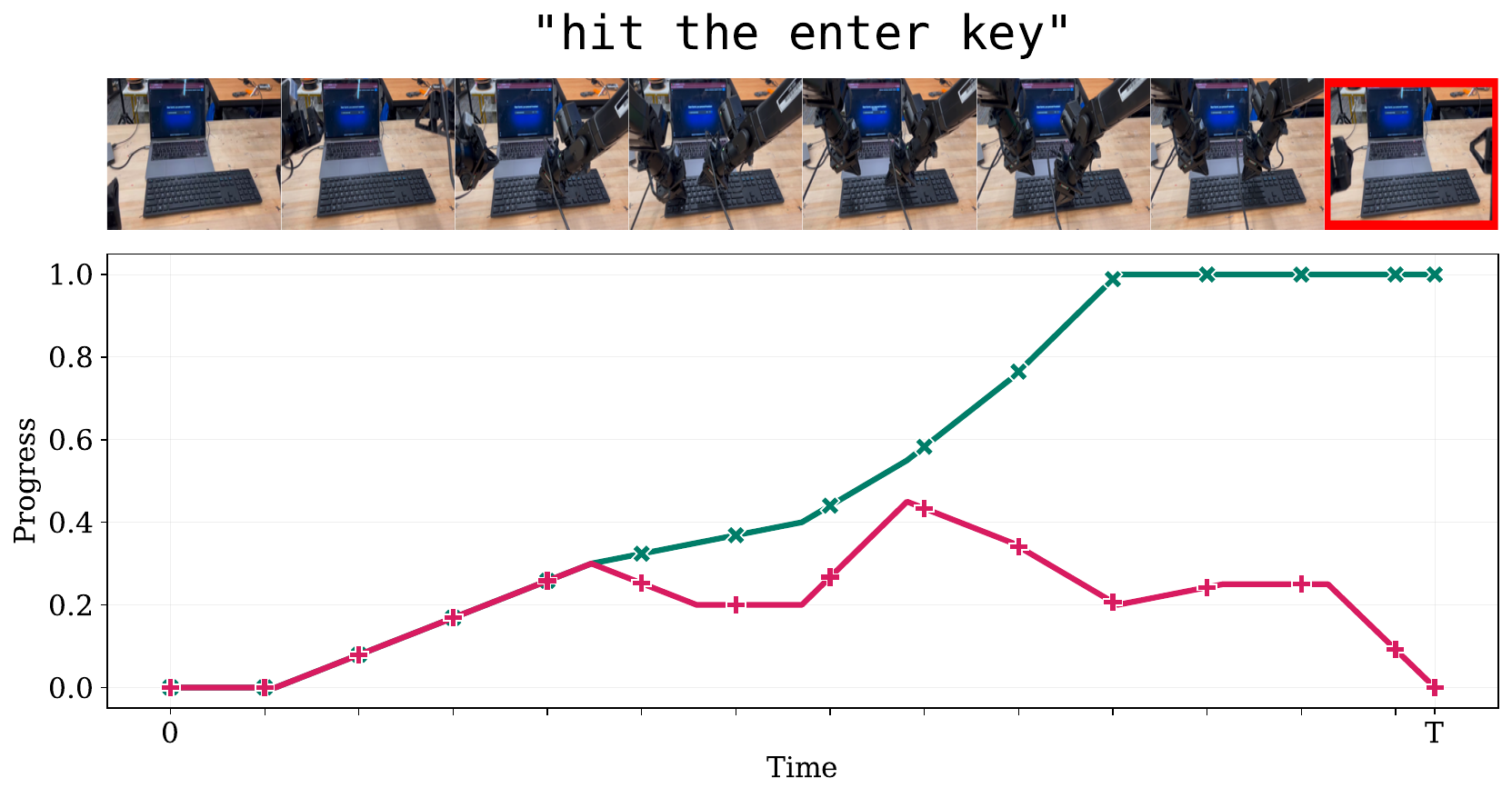}
        \caption{Enter task}
        \label{fig:dna_gains_astra_a_qual}
    \end{subfigure}
    \hfill
    \begin{subfigure}[b]{0.45\linewidth}
        \centering
        \includegraphics[width=\linewidth]{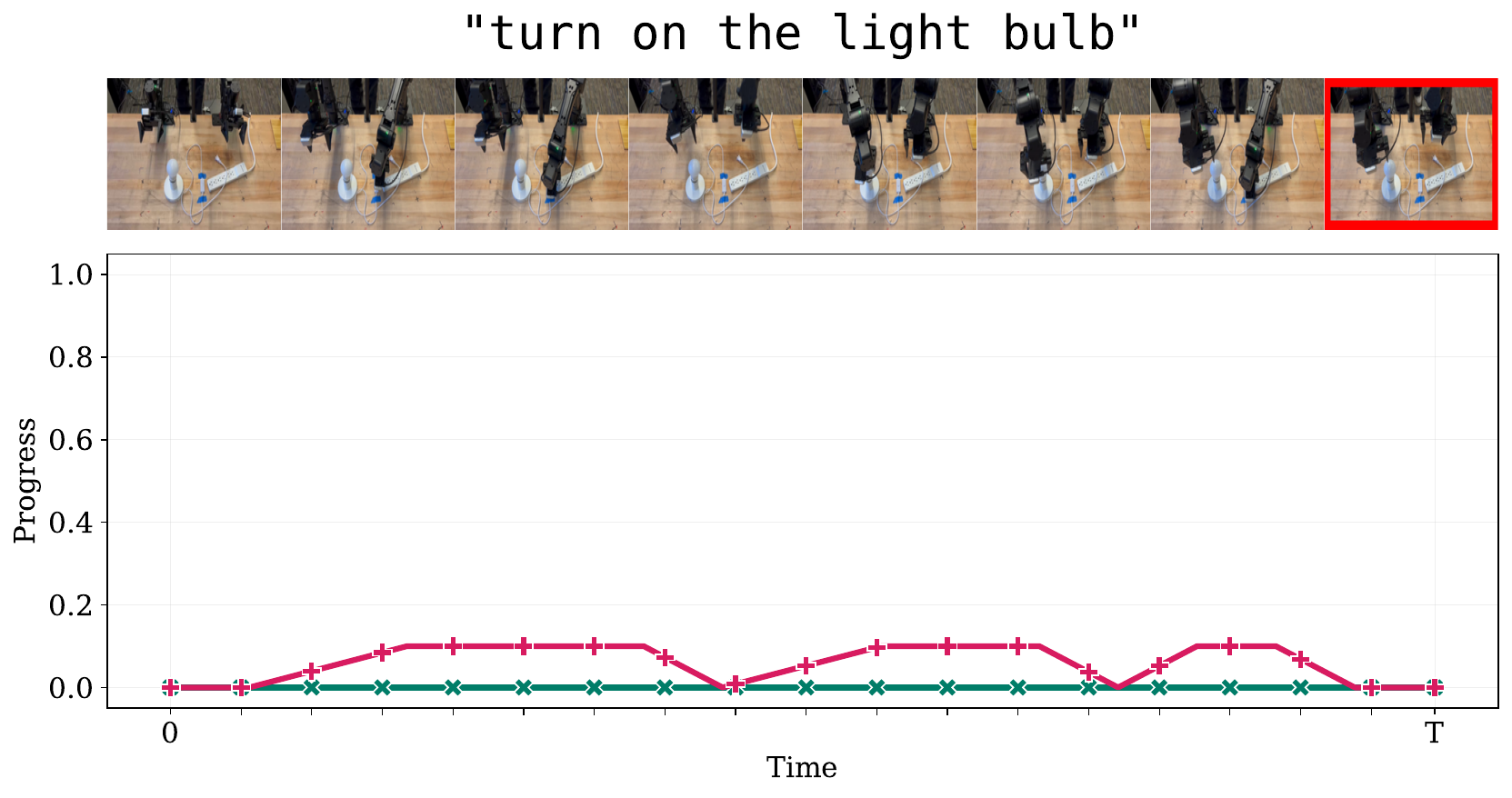}
        \caption{Light task}
        \label{fig:dna_gains_astra_b_qual}
    \end{subfigure}
    \label{fig:dna_gains_astra_qual}
    \caption{\textbf{\method qualitative gains on GPT-6 Astra}}
\end{figure*}

\begin{figure*}[h!]
    \centering
    \begin{subfigure}[b]{0.45\linewidth}
        \centering
        \includegraphics[width=\linewidth]{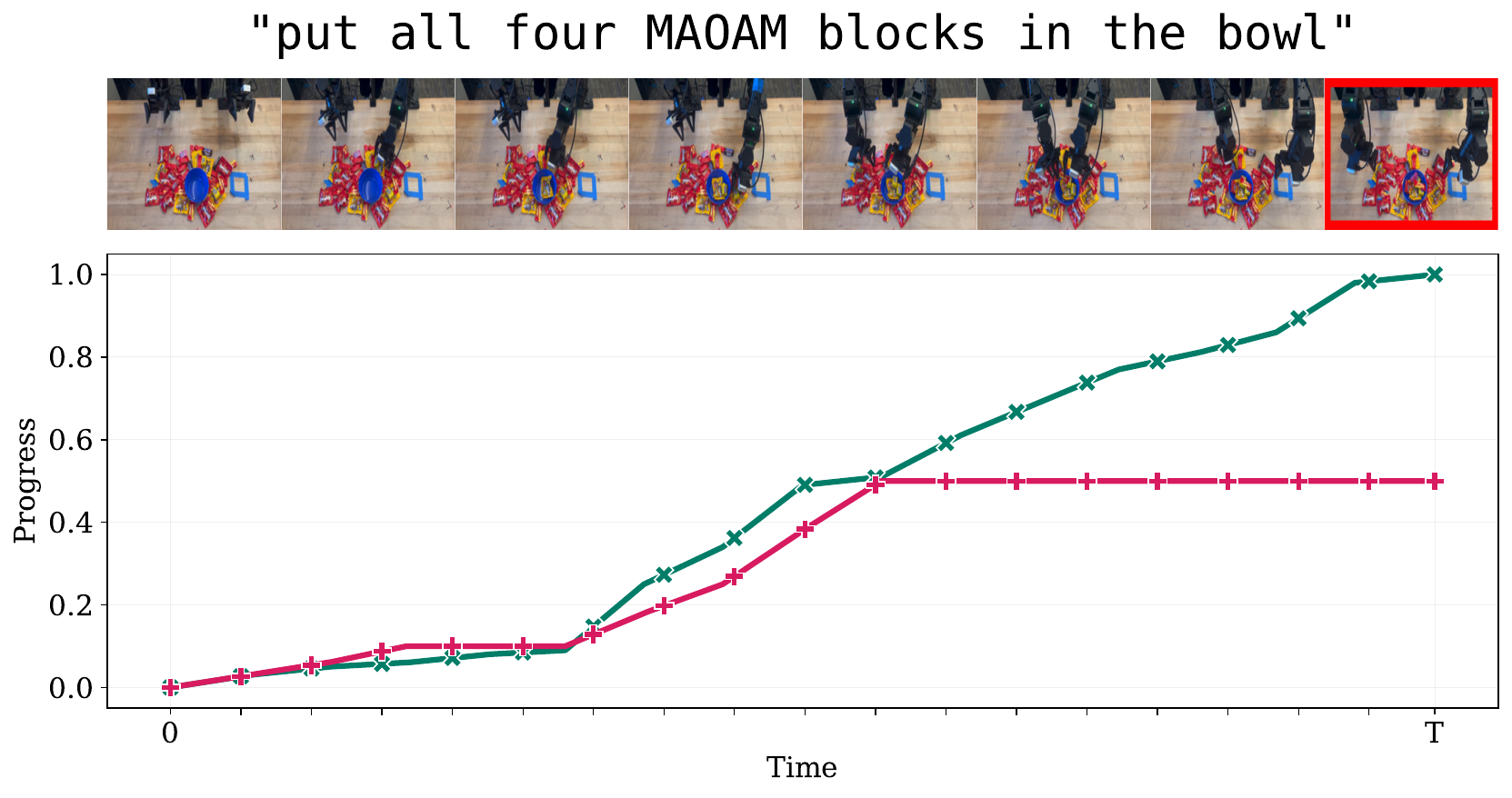}
        \caption{Maoam task}
        \label{fig:dna_gains_argon_a_qual}
    \end{subfigure}
    \hfill
    \begin{subfigure}[b]{0.45\linewidth}
        \centering
        \includegraphics[width=\linewidth]{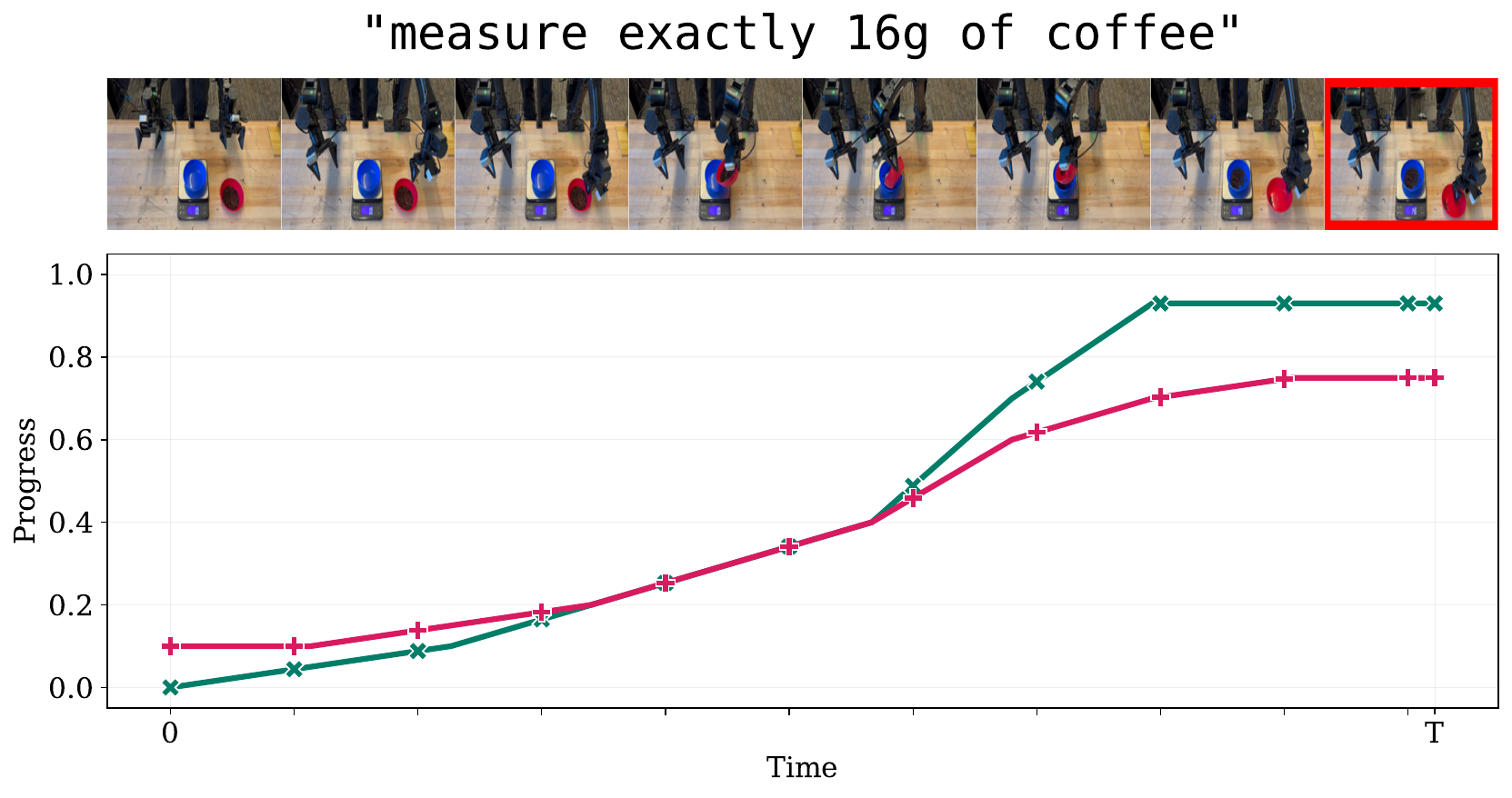}
        \caption{Coffee task}
        \label{fig:dna_gains_argon_b_qual}
    \end{subfigure}
    \label{fig:dna_gains_argon_qual}
    \caption{\textbf{\method qualitative gains on Gemini 4 Argon}}
\end{figure*}

Qualitatively, \Cref{fig:dna_gains_astra_a_qual} and \Cref{fig:dna_gains_argon_a_qual} both show the bias toward task success of \naive even when the task is not completed successfully, while \Cref{fig:dna_gains_astra_b_qual} and \Cref{fig:dna_gains_argon_b_qual} show \method's fine-grained reward assignment.

\section{Gains of \method over Base Model Capabilities - Gemini Model Family}
\label{sec:appdx:basemodel}

\begin{figure*}[!h]
    \centering
    \begin{minipage}{0.55\linewidth}
        \centering
        \begin{tabular}{lccc}
        \toprule
        \texttt{RBM-EVAL-OOD} & \naive & \Method & \Method \\
        \midrule
        Thinking Level & medium & medium & low \\
        \midrule
        Gemini 2.5 flash & 0.280 & 0.379 & 0.418 \\
        Gemini 2.5 pro   & 0.278 & 0.407 & 0.478 \\
        Gemini 3.5 flash & 0.499 & 0.517 & 0.609 \\
        Gemini 3.1 pro   & 0.515 & 0.620 & 0.591 \\
        \bottomrule
        \end{tabular}
        \captionof{table}{\textbf{Backbone and compute} (Kendall $\tau$ $\uparrow$)}
        \label{tab:scaling}
    \end{minipage}%
    \hfill
    \begin{minipage}{0.4\linewidth}
        \centering
        \includegraphics[width=\linewidth]{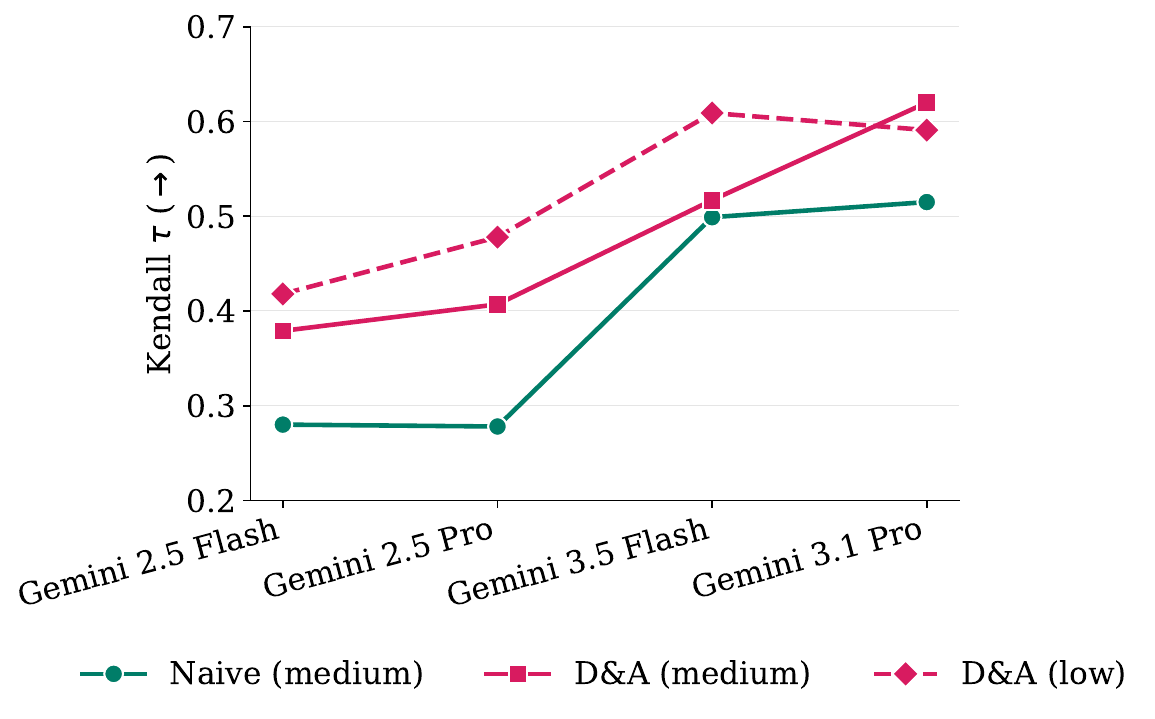}
        \caption{\textbf{\method gains persist across base models}}
        \label{fig:placeholder}
    \end{minipage}
\end{figure*}

The benefits of \Method persist as the base VLM becomes stronger, achieving consistent gains when improving the underlying Gemini models (\Cref{tab:scaling}). Crucially, these gains are not driven by increased test-time compute or reasoning, as reducing test-time compute by switching the thinking level to ''low`` matches results and can even improve performance (\Cref{tab:scaling}). Furthermore, baselines are evaluated on their strongest compatible backbones—such as Qwen-VL 8B Instruct for \topreward and Gemini 3 Flash for \gvl—rather than artificially weaker models. Existing baselines face major technical barriers on closed-source models: \topreward relies on token logprobs unavailable on proprietary platforms, \gvl performs poorly across closed models, and fine-tuning methods (\roboreward, \robometer) cannot be applied at all. As a result, \Method is the only approach that successfully scales to powerful closed-source models.

\newpage

\section{\method Cost, Wall Clock Time, and Token Usage}
\begin{table}[!h]
\centering
\label{tab:cost_analysis}
\begin{tabular}{lccc}
\toprule
\textbf{Model} & \textbf{Cost (\$)} & \textbf{Wall Clock Time (s)} & \textbf{Tokens} \\
\midrule
Gemini 3.1 Pro   & 0.035 & 23.7 & 7,358  \\
Gemini 3.5 Flash & 0.039 & 25.9 & 8,842  \\
Gemini 2.5 Pro   & 0.060 & 58.0 & 10,843 \\
Gemini 2.5 Flash & 0.015 & 36.9 & 11,177 \\
\bottomrule
\end{tabular}
\caption{Average compute, latency, and financial cost per trajectory evaluated across 567 trajectories on \texttt{RBM-EVAL-OOD}.}
\end{table}

\section{CaP Experiment Details}
\label{sec:appdx:cap}
We evaluate \method as a between-attempt feedback source for an open-loop code-as-policies planner on the LIBERO \texttt{libero\_goal} suite~\cite{liu2023libero}. The planner (gemini-3.7-flash, reasoning effort MEDIUM) observes the initial scene state and generates a code plan of absolute-waypoint primitives (\texttt{move\_to [x,y,z(,g)]}, gripper, done, $\leq40$ actions) which is then executed open-loop.
Each of the 500 initial conditions (10 tasks, 50 initial states each) is attempted 25 times. If required, \method scores each video (gemini-3.7-flash, reasoning effort MEDIUM, \texttt{agentview} rendered at $448\times448$, downsampled to 16 frames), and its output is appended to the planner's accumulated history for the next attempt.
Let $s_i(t)\in\{0,1\}$ denote ground-truth task success of initial condition $i$ on attempt $t$,
for $i=1,\dots,N$ ($N=500$) and $t=1,\dots,25$. We report Success@k, cumulative unique initial conditions solved at attempt $k$: $M(k) = \frac{1}{N}\sum_{i=1}^{N} \max_{1\le t\le k} s_i(t)$

\section{\texttt{D\&A-Bench} Benchmark and Qualitative Results}
\label{sec:appdx:dnabench}

Dataset: 14 real-world manipulation tasks manually collected on a bimanual YAM robot platform. Each task recorded as 2-4 iPhone-captured episodes (2 success, 2 failure; book task only 1 each), resized to 1280x720. Every task has one handwritten instruction plus 9 AI-generated paraphrases (Gemini 3 Flash, shown first frame of episode + original instruction, prompted for M paraphrases preserving meaning but varying phrasing/specificity/vocabulary), giving 10 instructions/task. Full grid: 14 tasks × 4 episodes × 10 instructions = 560 (episode, instruction) pairs scored per reward model. Below shows predictions of all model for the vanilla instruction. While \texttt{RBM-OOD-EVAL} subsamples the videos to 8 frames~\cite{liang2026robometer}, we subsample the videos by a factor of 60 (min 8 frames) to capture visual subtleties in the examples below.
We make the dataset publicly available at:  \url{https://huggingface.co/datasets/memmelma/dnabench}

% Requires in preamble:
%   \usepackage{graphicx}
%   \usepackage{subcaption}
%   \usepackage{caption}
% Fixed-size figures regardless of caption length: every subfigure image is
% forced to the same width via \qualimg (below), independent of how long the
% instruction text in its caption is (subfigure width is fixed to \qualimgw,
% not shrunk/grown by caption content). All source PDFs share the same
% aspect ratio, so a fixed width alone is enough to make every panel the
% same size — no height cap (that was the previous, too-small version:
% capping height at 4.2cm forced the width down well below \qualimgw,
% leaving big blank margins around each image). subfigure alignment [t]
% keeps image boxes top-aligned regardless of caption length.
\newlength{\qualimgw}
\setlength{\qualimgw}{0.49\columnwidth}

\newcommand{\qualimg}[1]{%
    \includegraphics[width=\qualimgw]{#1}%
}

\begin{figure*}[!h]
    \centering
    \includegraphics[width=0.6\columnwidth]{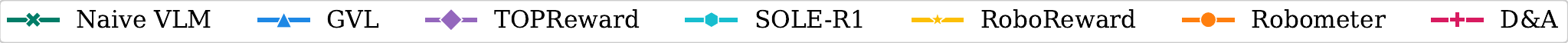}\\[4pt]
    \begin{subfigure}[t]{\qualimgw}
        \qualimg{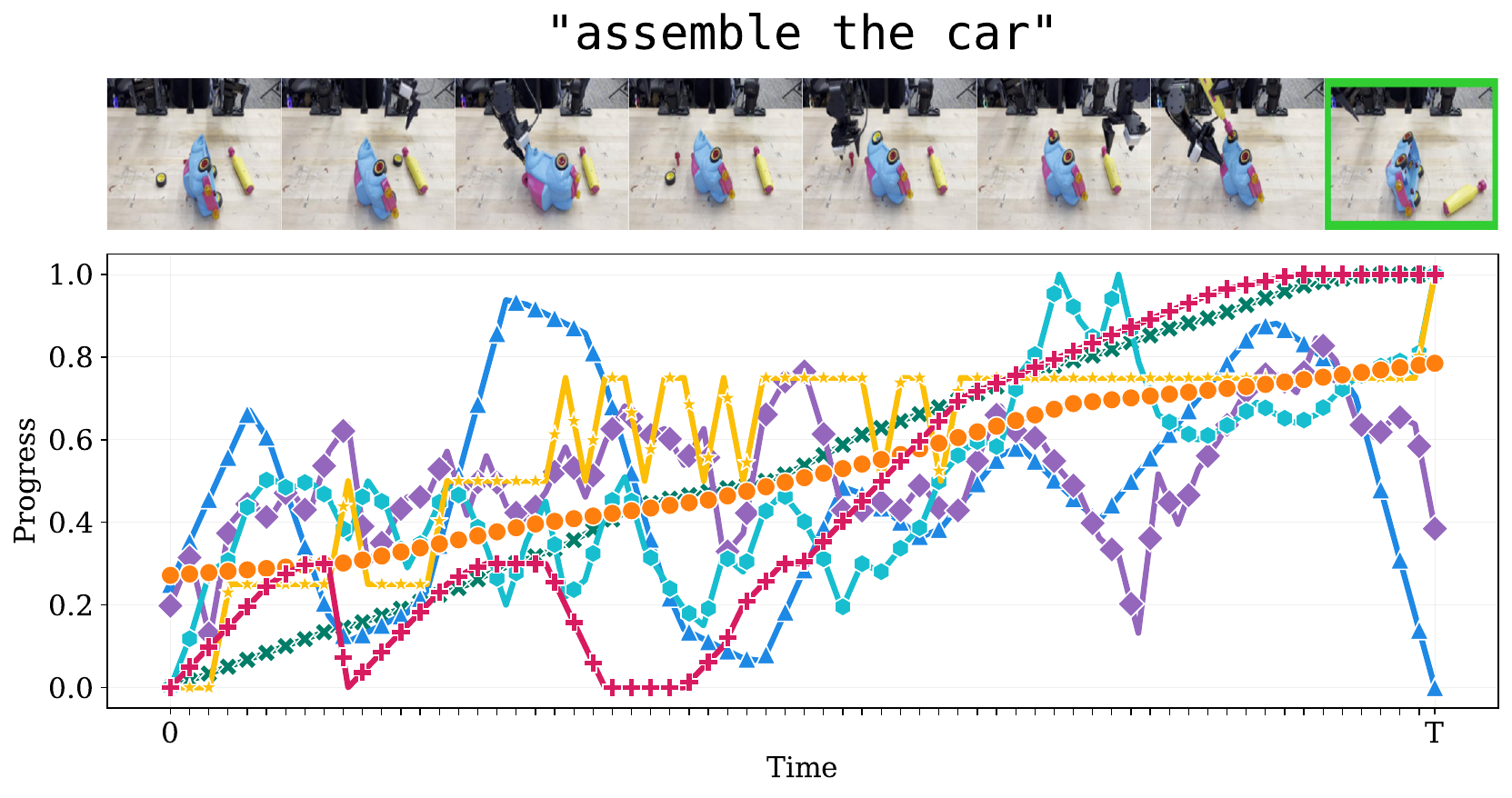}
        \caption{``assemble the car'' (success).}
    \end{subfigure}
    \hfill
    \begin{subfigure}[t]{\qualimgw}
        \qualimg{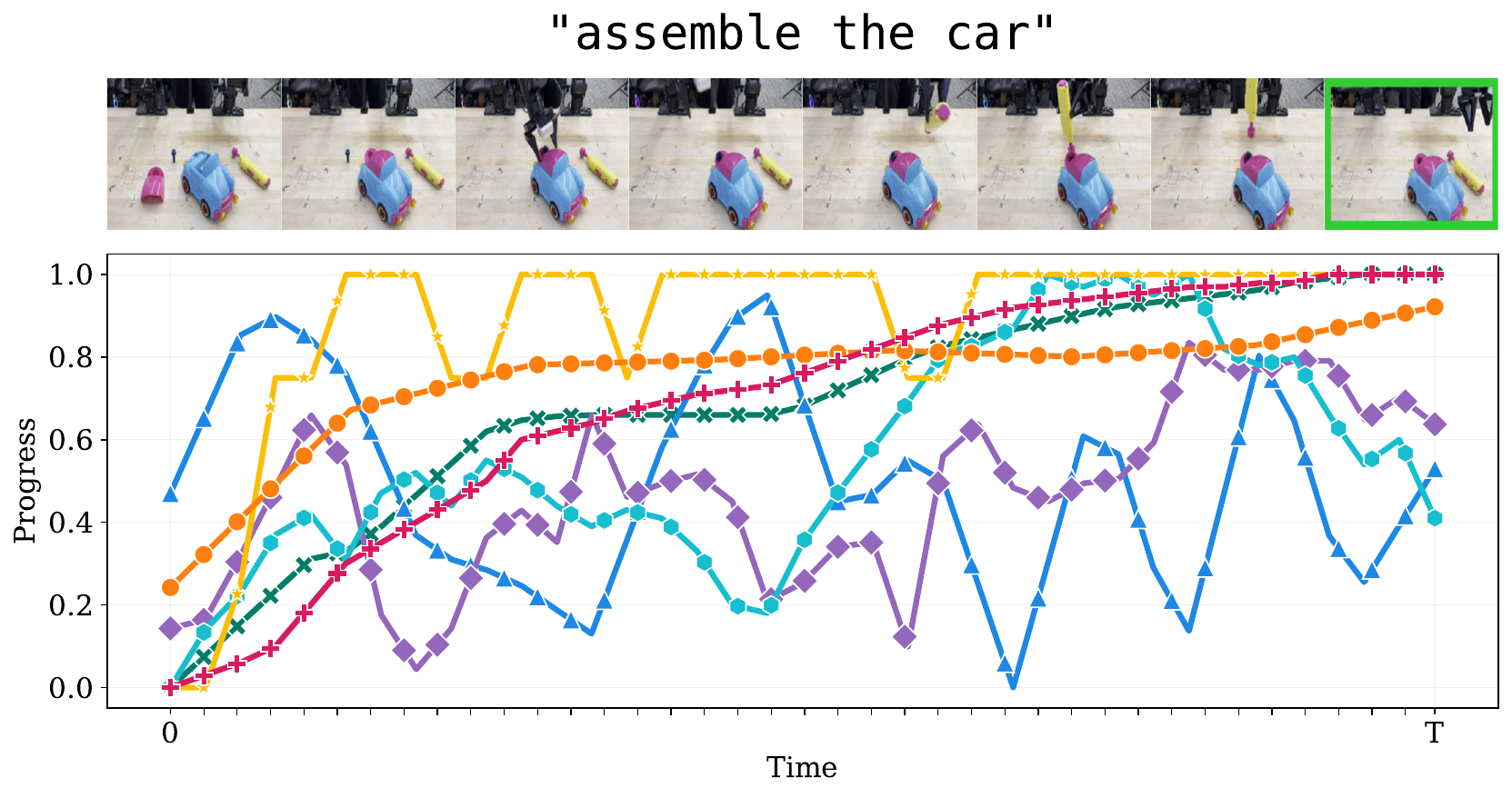}
        \caption{``assemble the car'' (success).}
    \end{subfigure}

    \vspace{0.5em}
    \begin{subfigure}[t]{\qualimgw}
        \qualimg{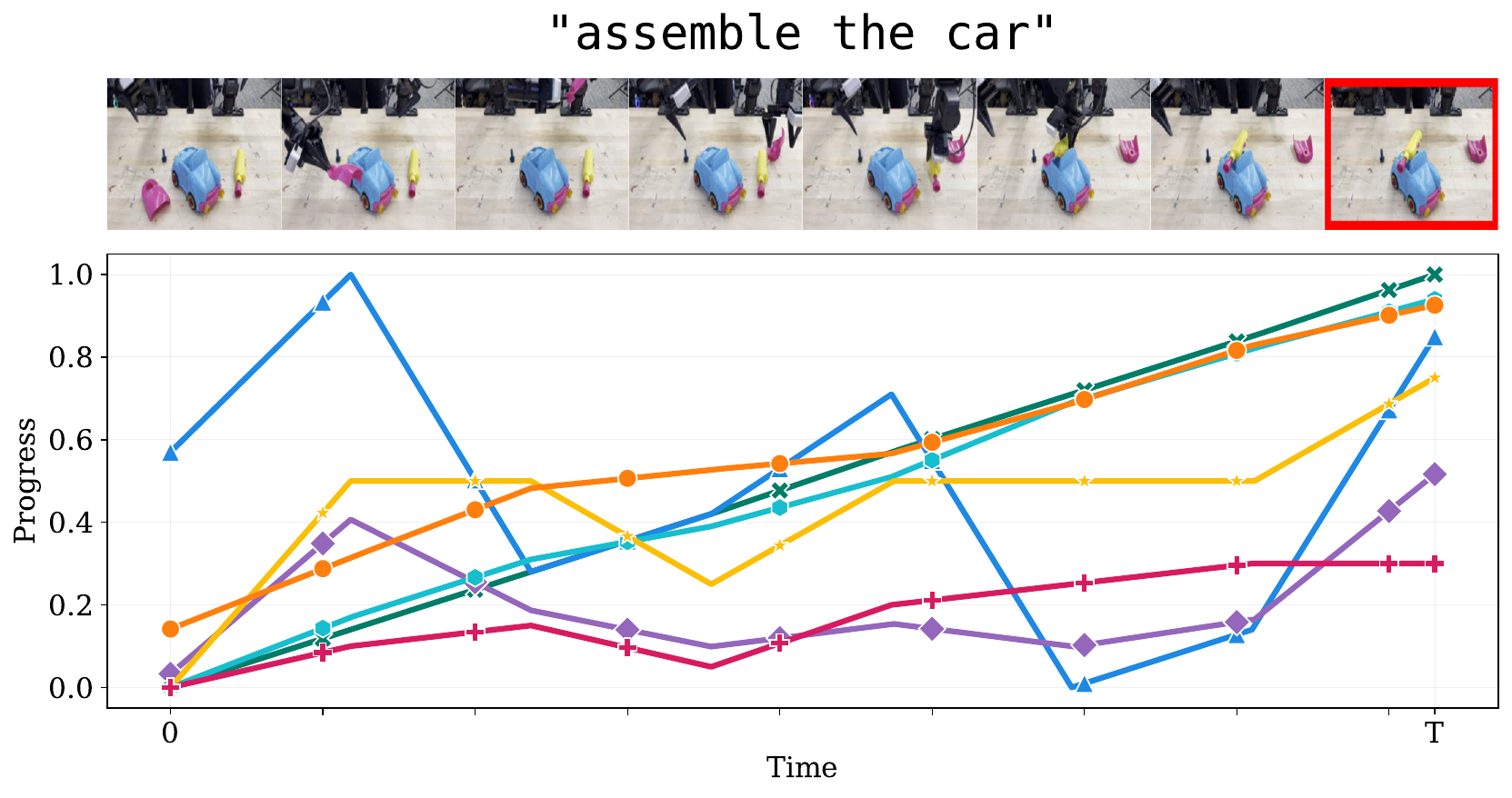}
        \caption{``assemble the car'' (failure).}
    \end{subfigure}
    \hfill
    \begin{subfigure}[t]{\qualimgw}
        \qualimg{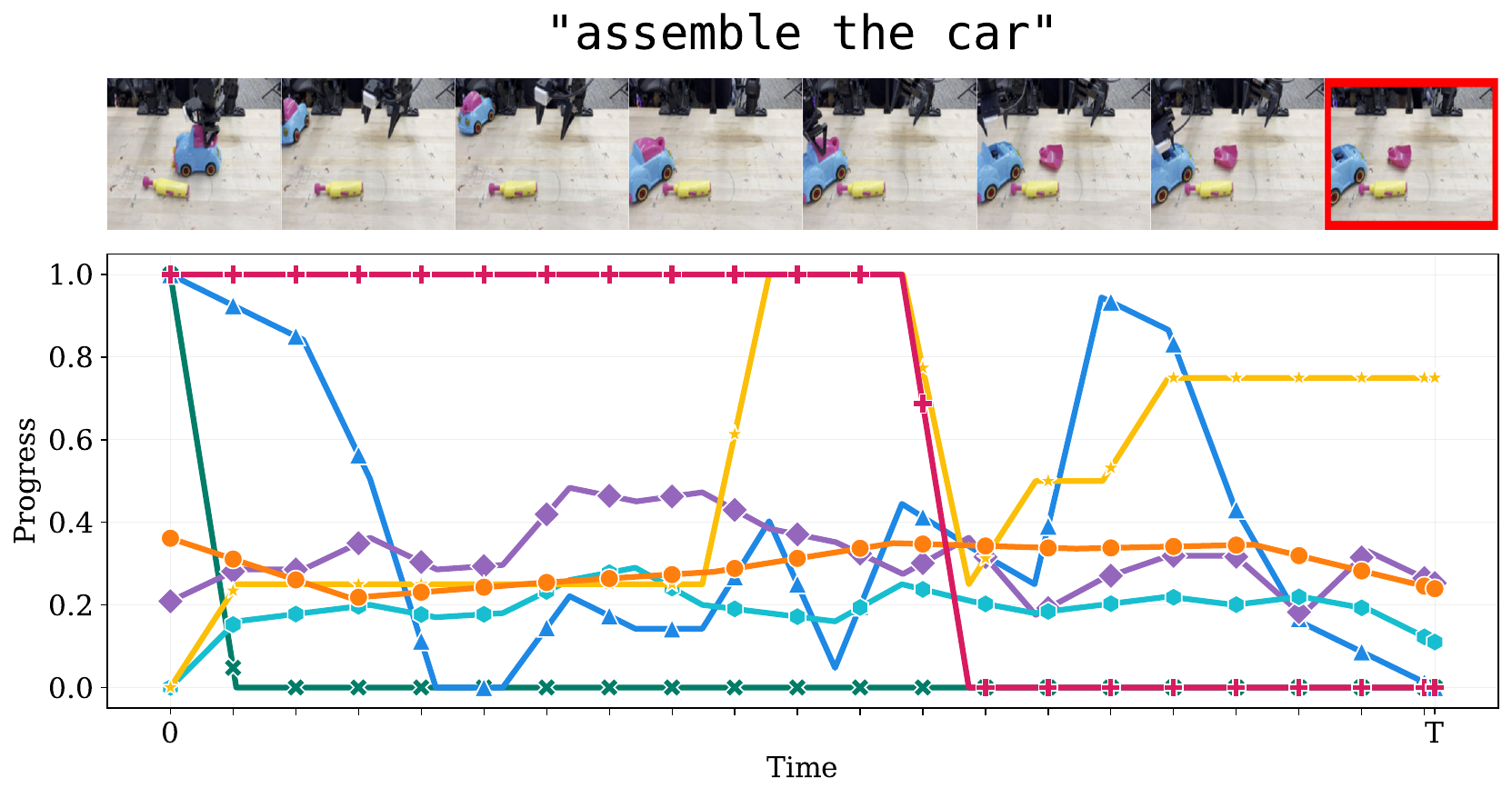}
        \caption{``assemble the car'' (failure).}
    \end{subfigure}
    \caption{\textbf{Qualitative per-trajectory predictions for ``assemble the car''.}}
    \label{fig:qualitative_car_task}
\end{figure*}

\begin{figure*}[!h]
    \centering
    \includegraphics[width=0.6\columnwidth]{plots/legend.pdf}\\[4pt]
    \begin{subfigure}[t]{\qualimgw}
        \qualimg{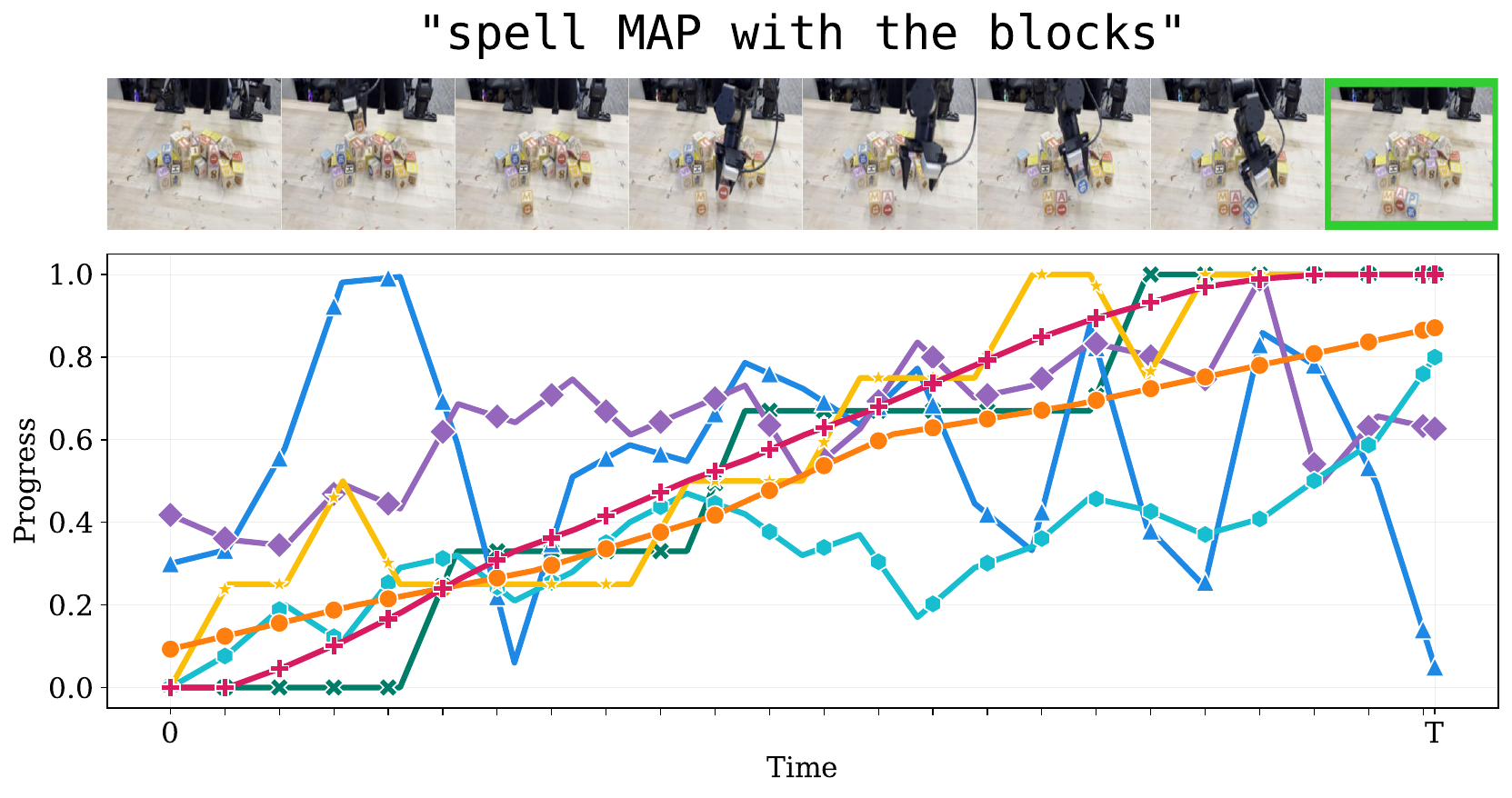}
        \caption{``spell MAP with the blocks'' (success).}
    \end{subfigure}
    \hfill
    \begin{subfigure}[t]{\qualimgw}
        \qualimg{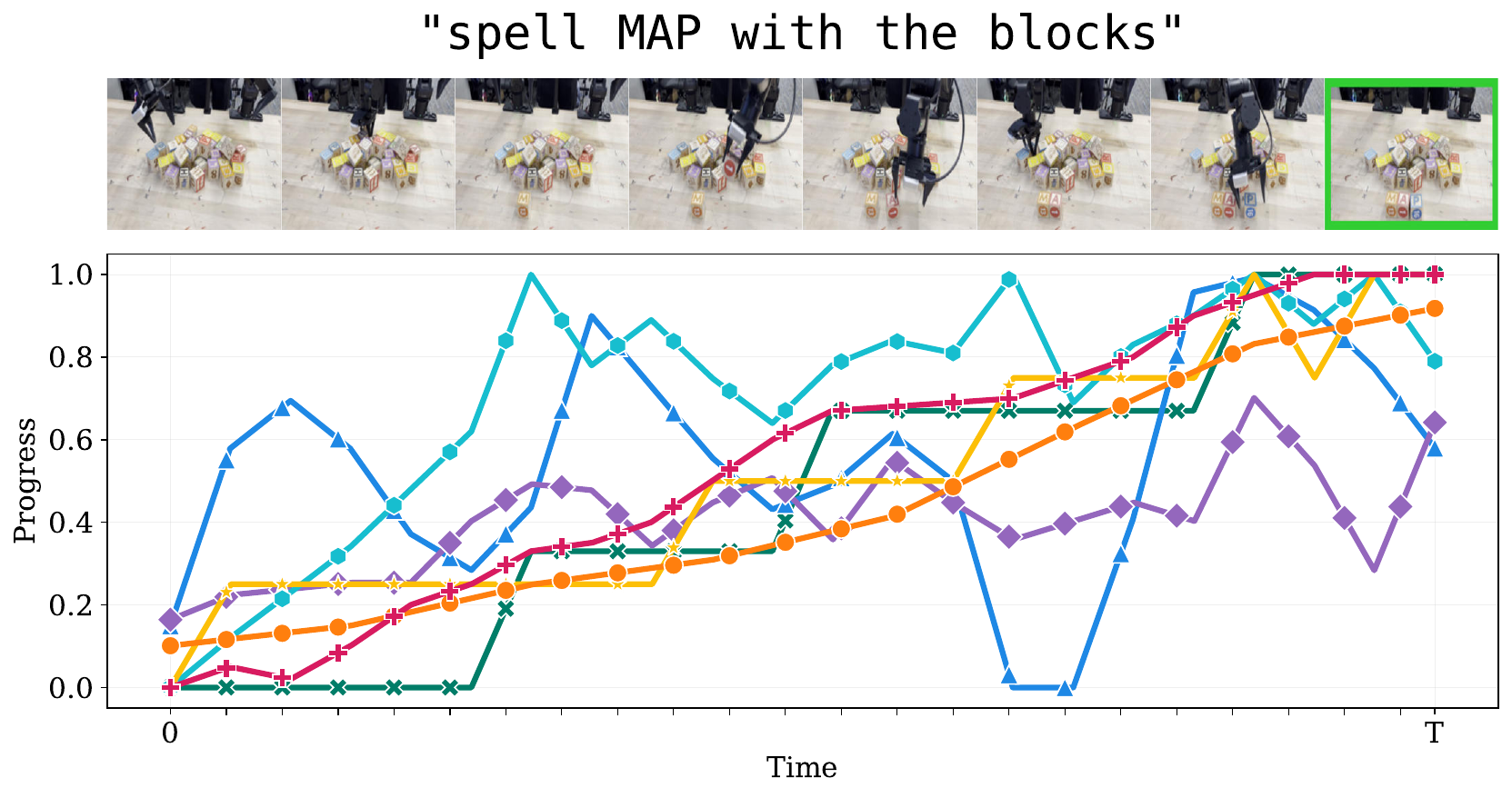}
        \caption{``spell MAP with the blocks'' (success).}
    \end{subfigure}

    \vspace{0.5em}
    \begin{subfigure}[t]{\qualimgw}
        \qualimg{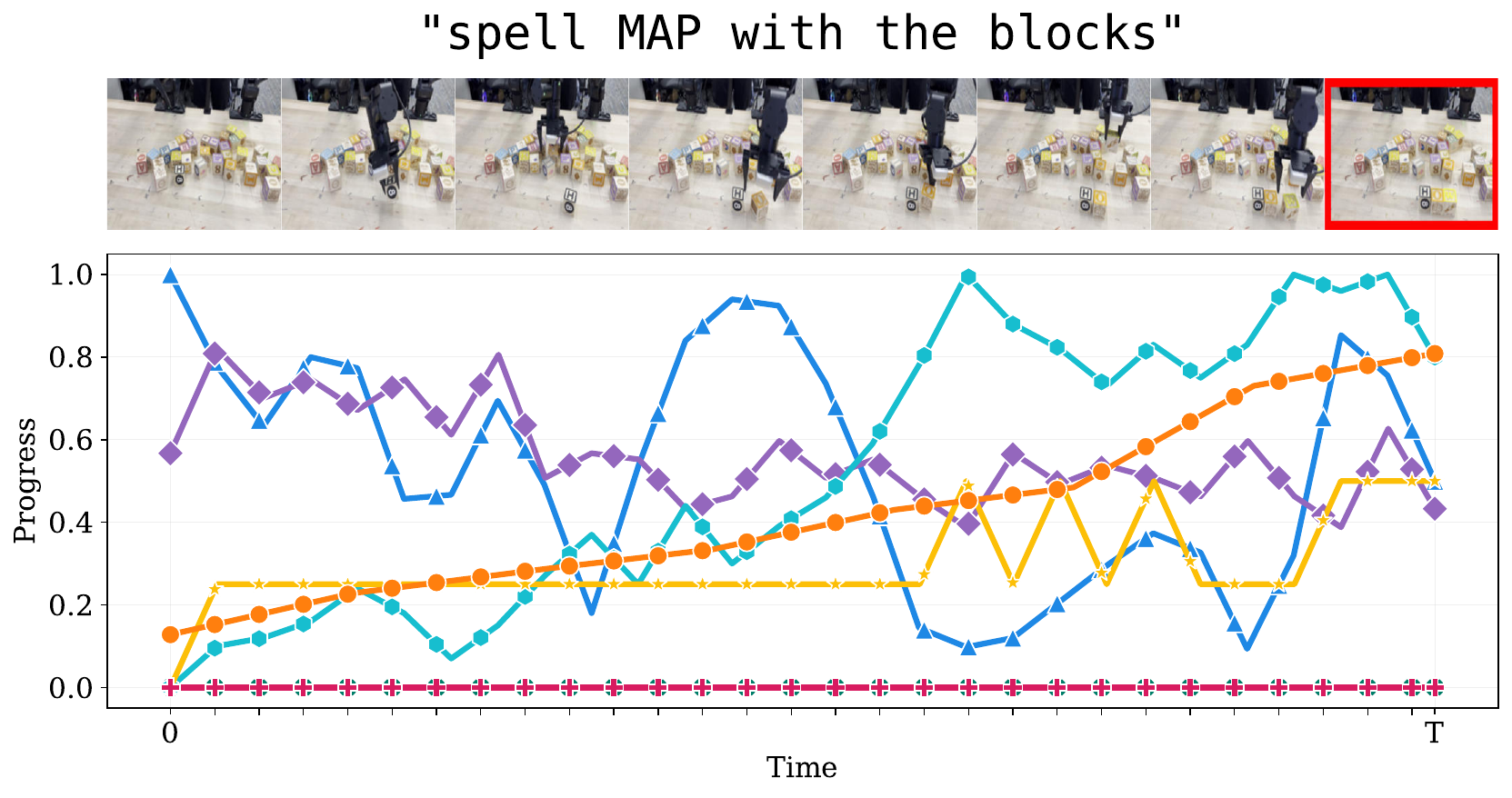}
        \caption{``spell MAP with the blocks'' (failure).}
    \end{subfigure}
    \hfill
    \begin{subfigure}[t]{\qualimgw}
        \qualimg{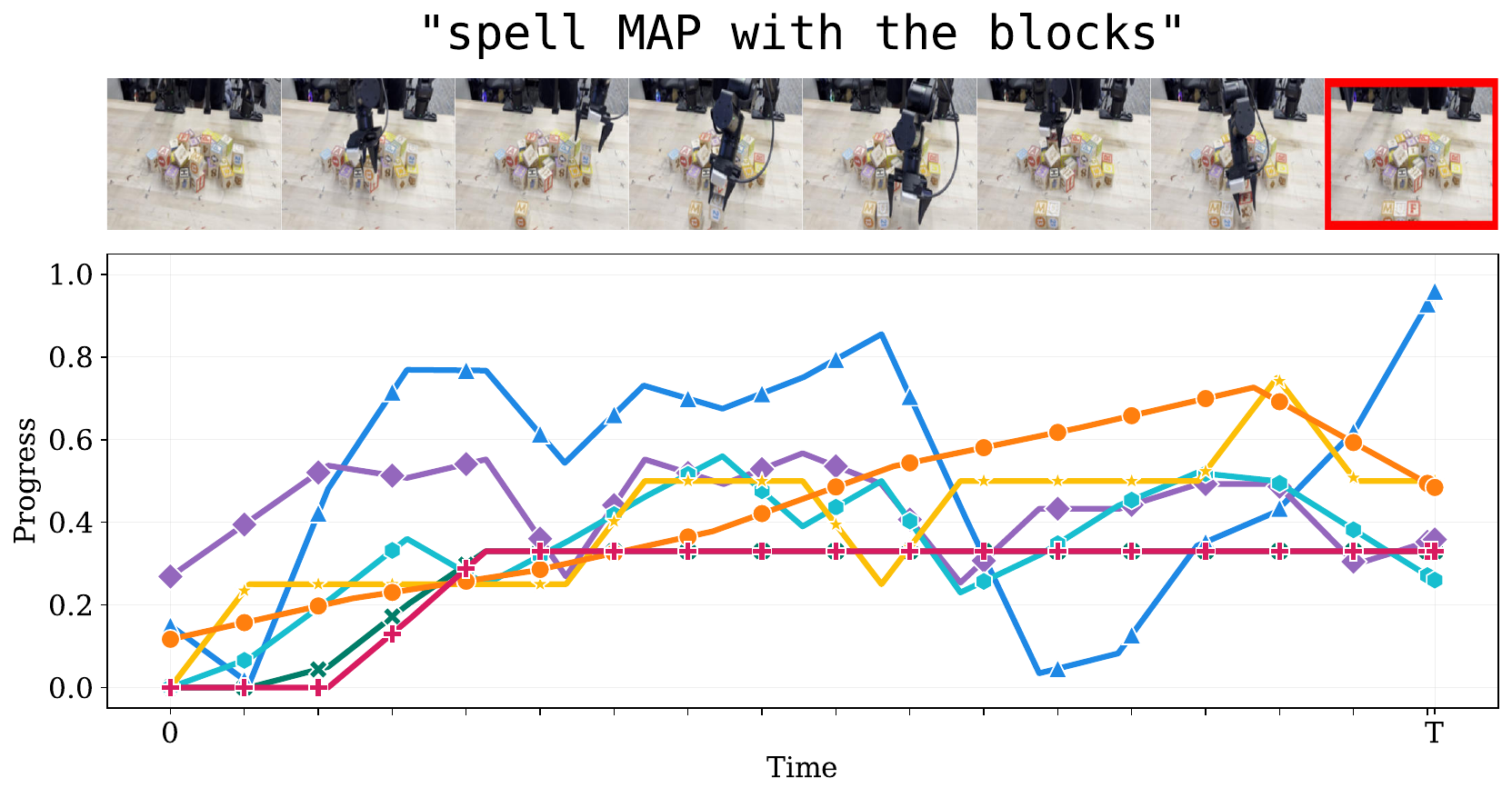}
        \caption{``spell MAP with the blocks'' (failure).}
    \end{subfigure}
    \caption{\textbf{Qualitative per-trajectory predictions for ``spell MAP with the blocks''.}}
    \label{fig:qualitative_spell_task}
\end{figure*}

\begin{figure*}[!h]
    \centering
    \includegraphics[width=0.6\columnwidth]{plots/legend.pdf}\\[4pt]
    \begin{subfigure}[t]{\qualimgw}
        \qualimg{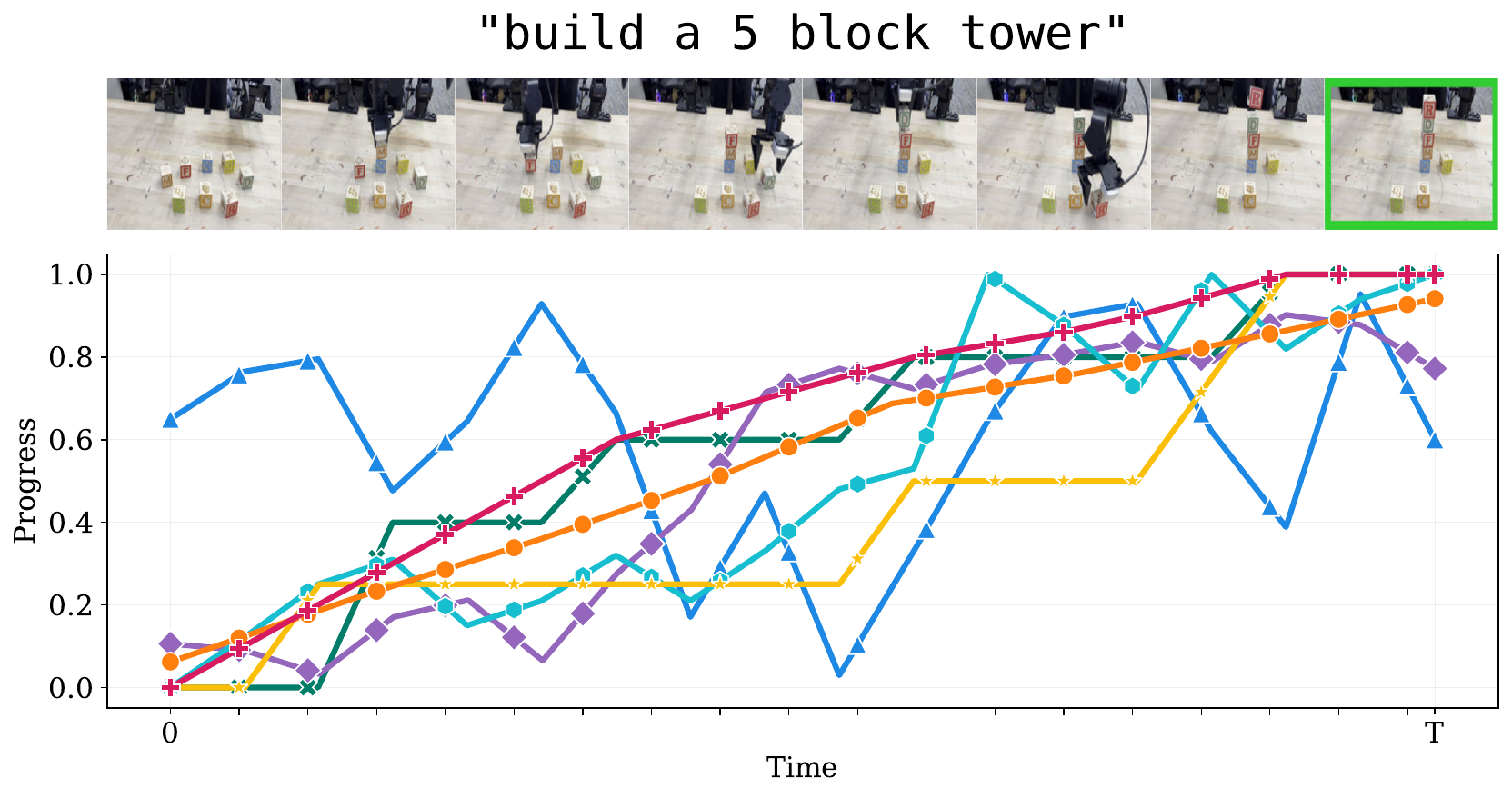}
        \caption{``build a 5 block tower'' (success).}
    \end{subfigure}
    \hfill
    \begin{subfigure}[t]{\qualimgw}
        \qualimg{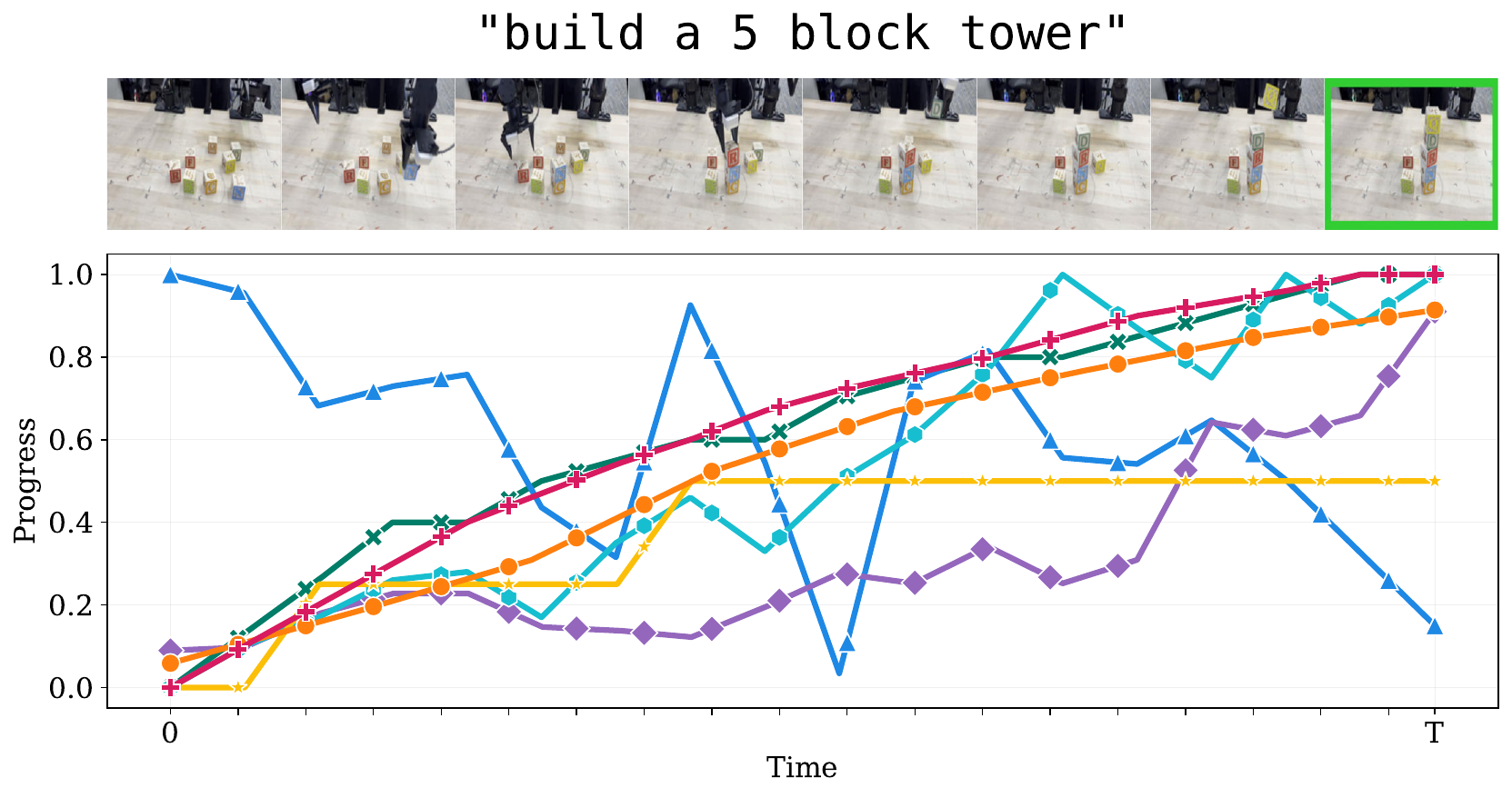}
        \caption{``build a 5 block tower'' (success).}
    \end{subfigure}

    \vspace{0.5em}
    \begin{subfigure}[t]{\qualimgw}
        \qualimg{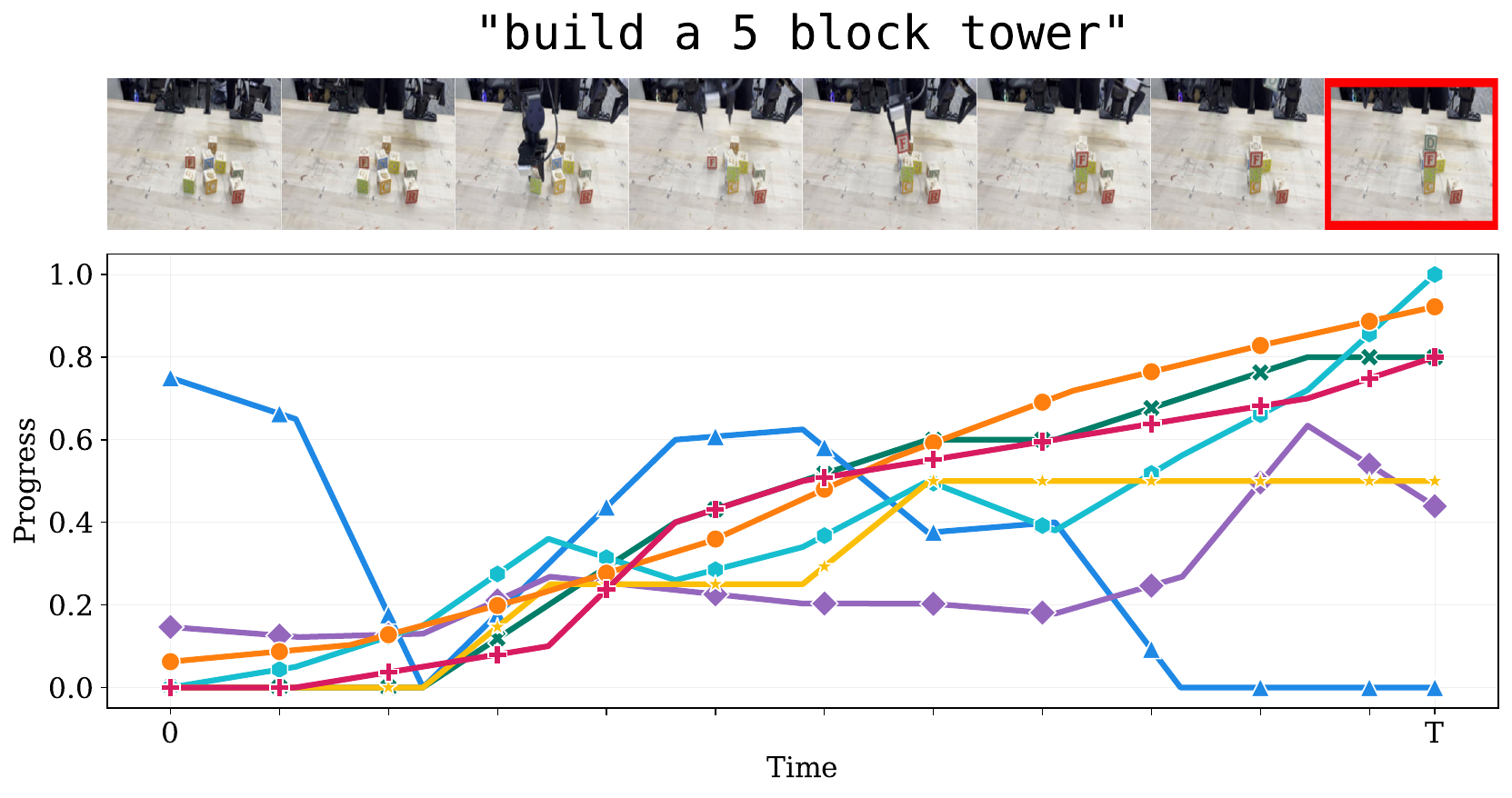}
        \caption{``build a 5 block tower'' (failure).}
    \end{subfigure}
    \hfill
    \begin{subfigure}[t]{\qualimgw}
        \qualimg{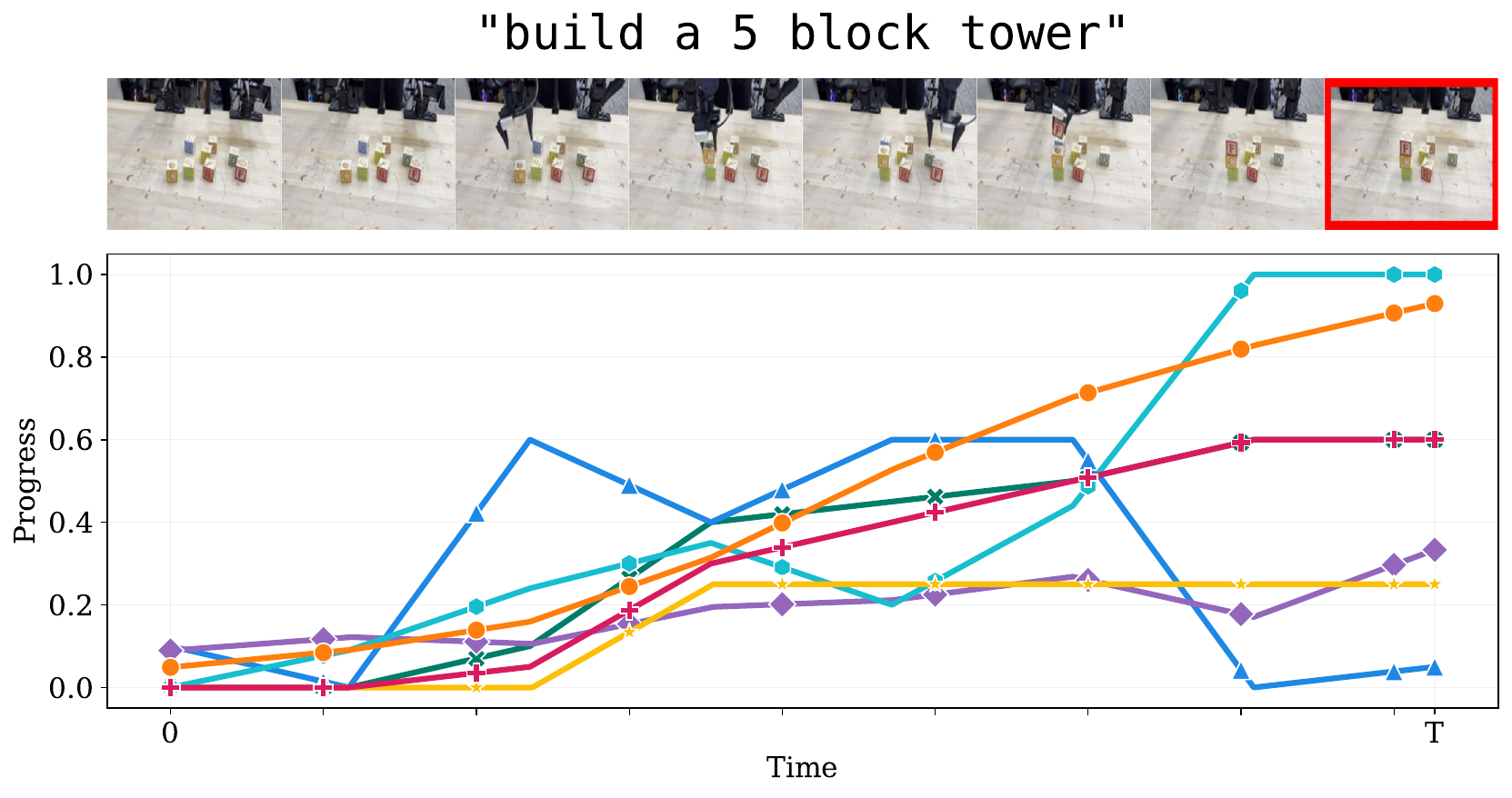}
        \caption{``build a 5 block tower'' (failure).}
    \end{subfigure}
    \caption{\textbf{Qualitative per-trajectory predictions for ``build a 5 block tower''.}}
    \label{fig:qualitative_tower_task}
\end{figure*}

\begin{figure*}[!h]
    \centering
    \includegraphics[width=0.6\columnwidth]{plots/legend.pdf}\\[4pt]
    \begin{subfigure}[t]{\qualimgw}
        \qualimg{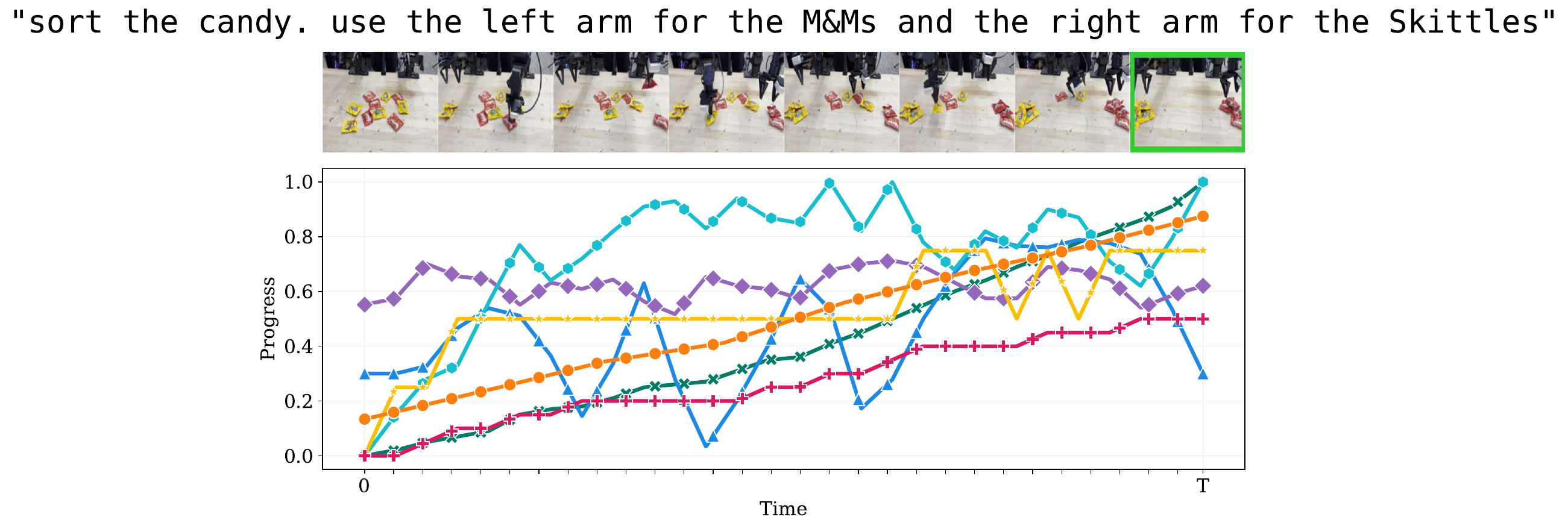}
        \caption{``sort the candy. use the left arm for the M\&Ms and the right arm for the Skittles'' (success).}
    \end{subfigure}
    \hfill
    \begin{subfigure}[t]{\qualimgw}
        \qualimg{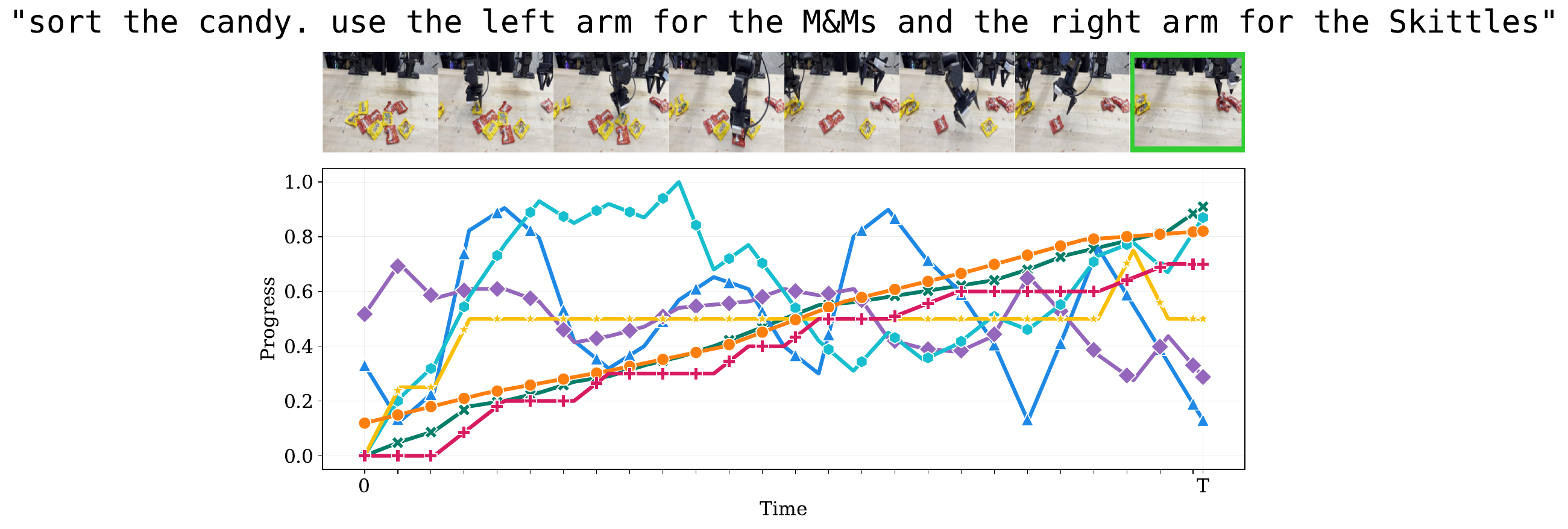}
        \caption{``sort the candy. use the left arm for the M\&Ms and the right arm for the Skittles'' (success).}
    \end{subfigure}

    \vspace{0.5em}
    \begin{subfigure}[t]{\qualimgw}
        \qualimg{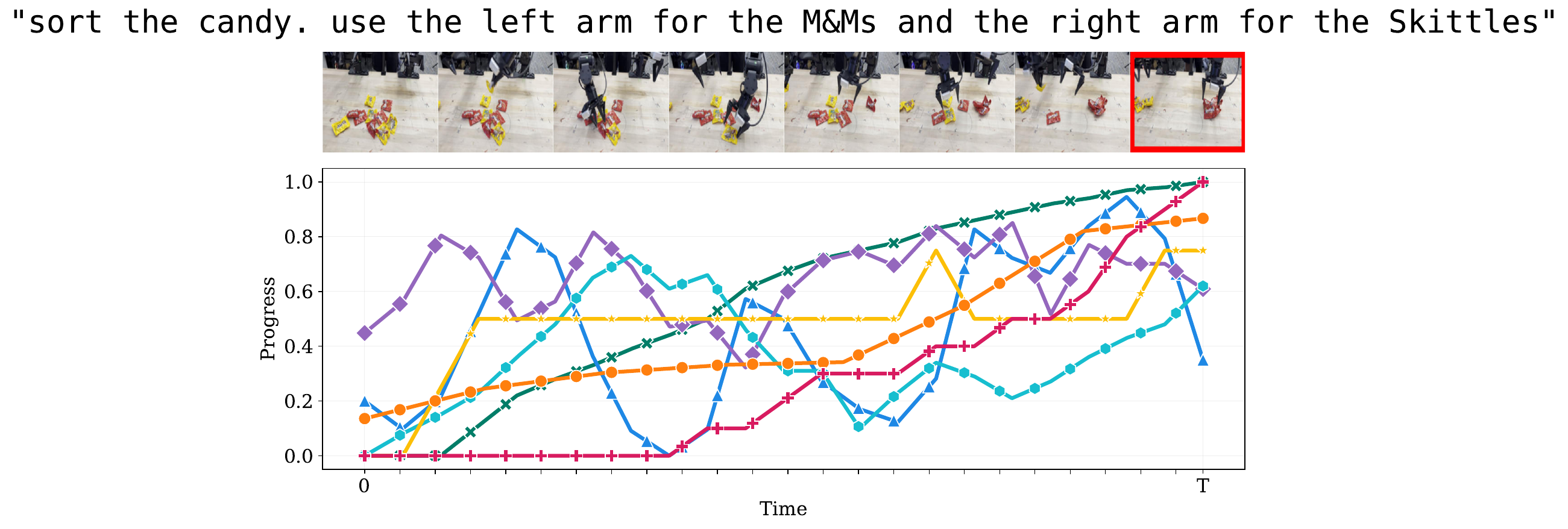}
        \caption{``sort the candy. use the left arm for the M\&Ms and the right arm for the Skittles'' (failure).}
    \end{subfigure}
    \hfill
    \begin{subfigure}[t]{\qualimgw}
        \qualimg{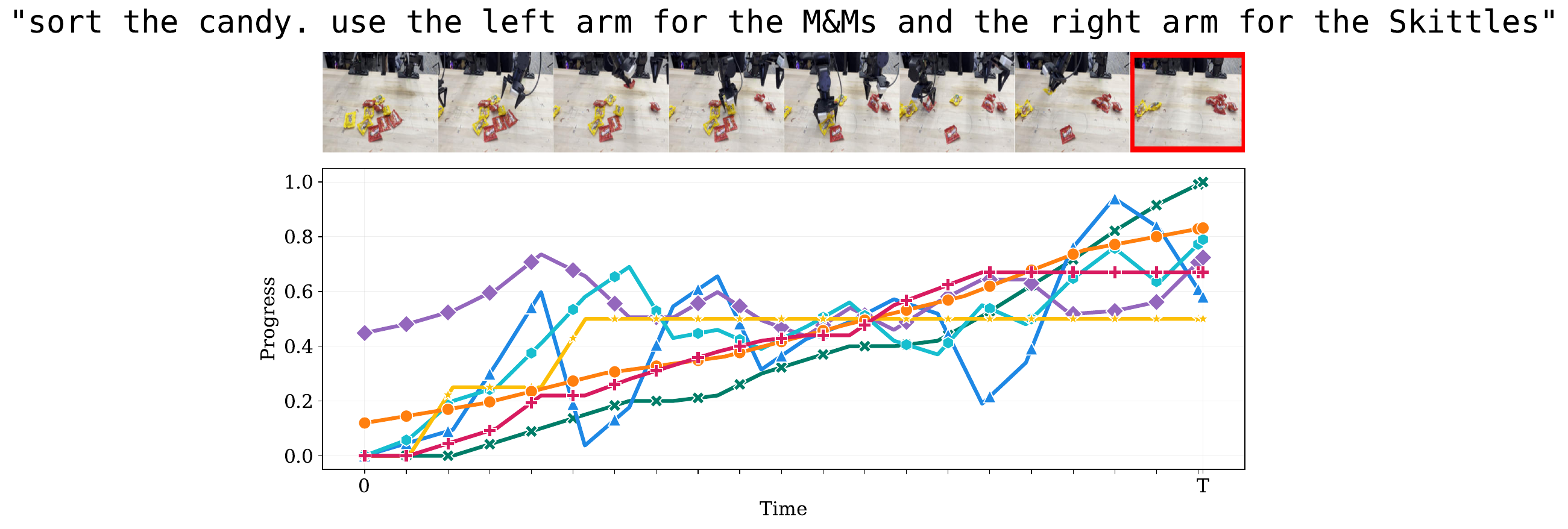}
        \caption{``sort the candy. use the left arm for the M\&Ms and the right arm for the Skittles'' (failure).}
    \end{subfigure}
    \caption{\textbf{Qualitative per-trajectory predictions for ``sort the candy''.}}
    \label{fig:qualitative_sorting_task}
\end{figure*}

\begin{figure*}[!h]
    \centering
    \includegraphics[width=0.6\columnwidth]{plots/legend.pdf}\\[4pt]
    \begin{subfigure}[t]{\qualimgw}
        \qualimg{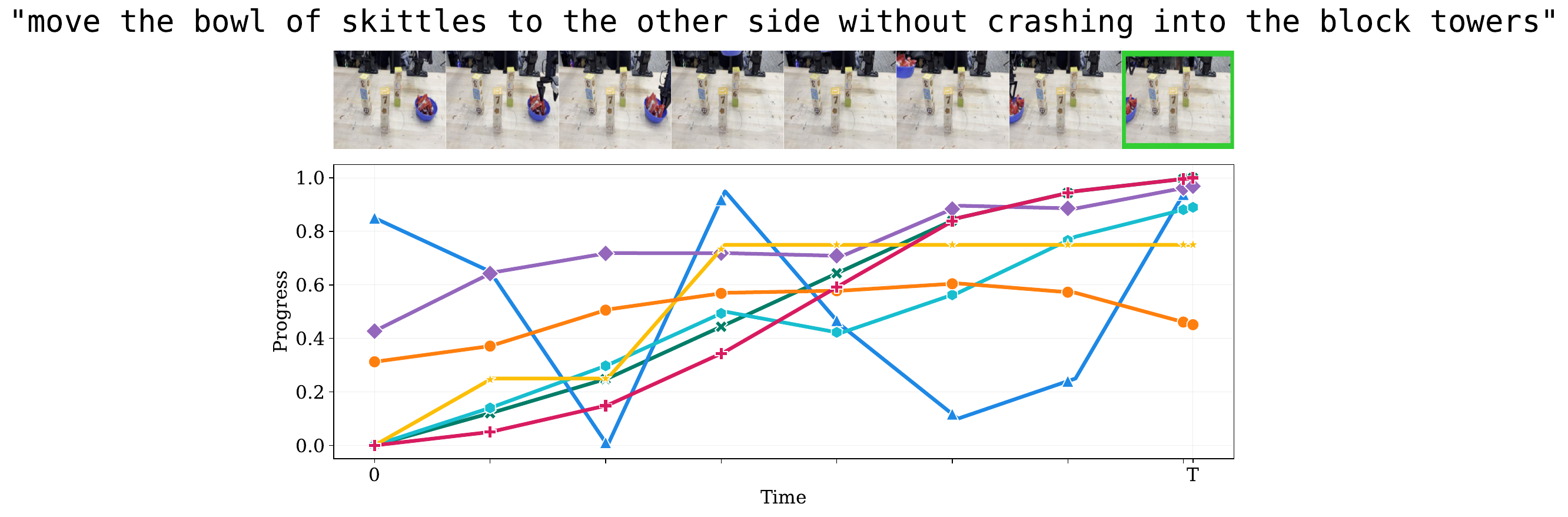}
        \caption{``move the bowl of skittles to the other side without crashing into the block towers'' (success).}
    \end{subfigure}
    \hfill
    \begin{subfigure}[t]{\qualimgw}
        \qualimg{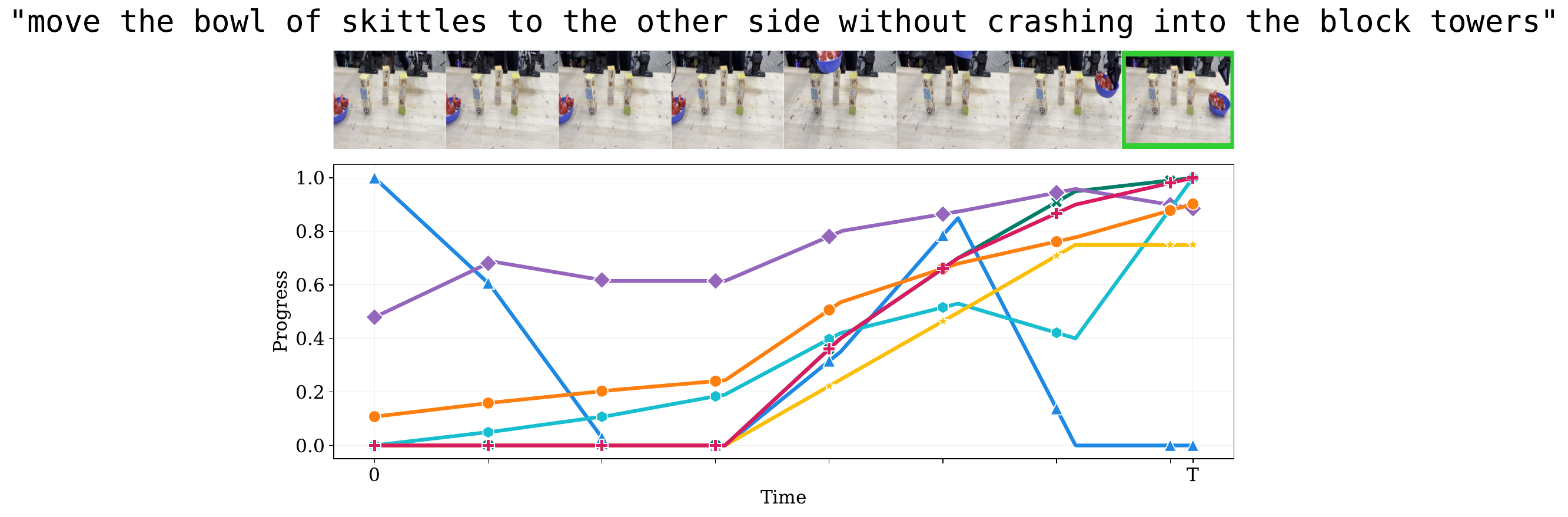}
        \caption{``move the bowl of skittles to the other side without crashing into the block towers'' (success).}
    \end{subfigure}

    \vspace{0.5em}
    \begin{subfigure}[t]{\qualimgw}
        \qualimg{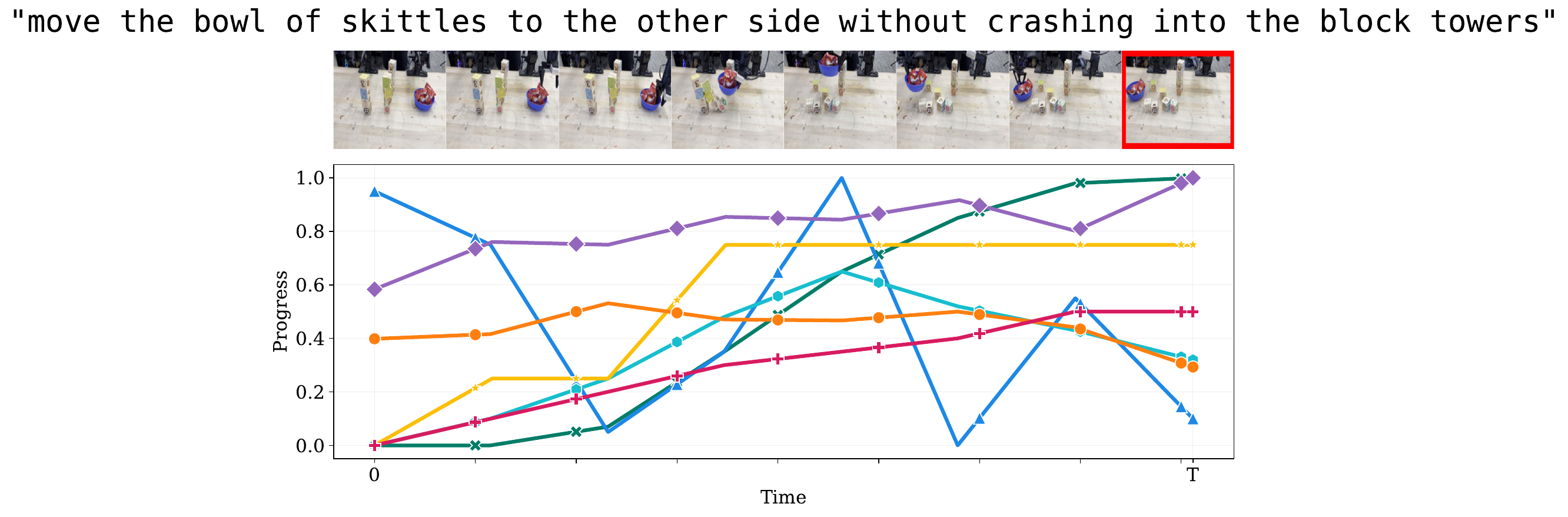}
        \caption{``move the bowl of skittles to the other side without crashing into the block towers'' (failure).}
    \end{subfigure}
    \hfill
    \begin{subfigure}[t]{\qualimgw}
        \qualimg{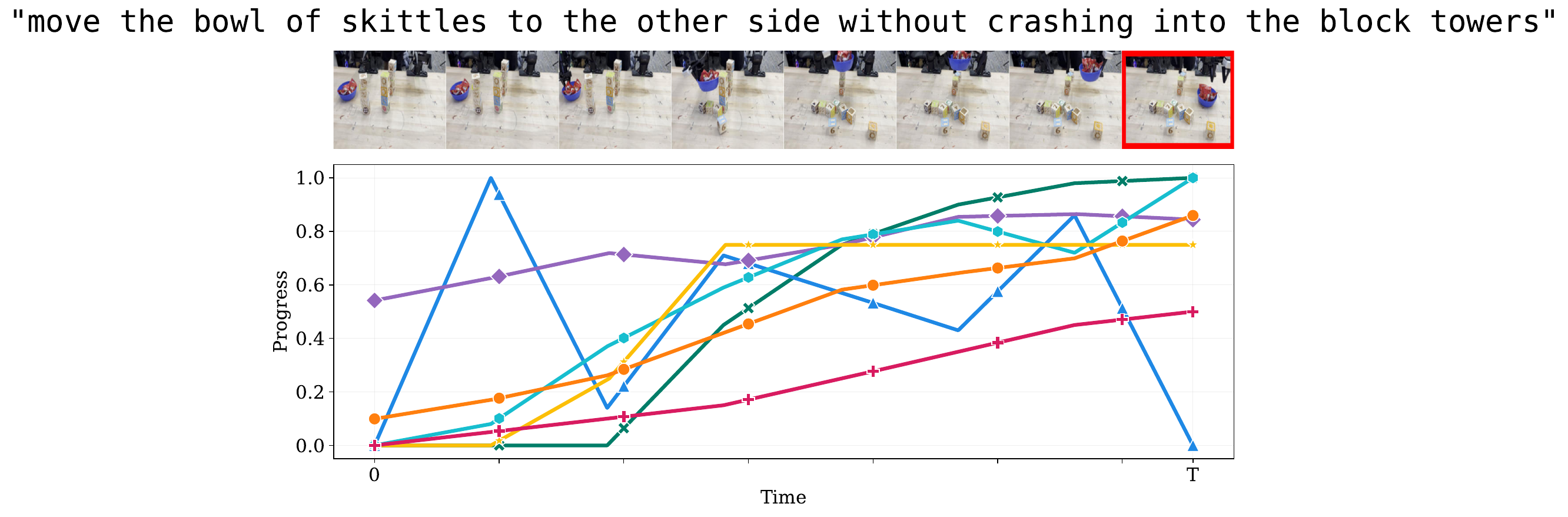}
        \caption{``move the bowl of skittles to the other side without crashing into the block towers'' (failure).}
    \end{subfigure}
    \caption{\textbf{Qualitative per-trajectory predictions for ``move the bowl of skittles without crashing''.}}
    \label{fig:qualitative_crash_task}
\end{figure*}

\begin{figure*}[!h]
    \centering
    \includegraphics[width=0.6\columnwidth]{plots/legend.pdf}\\[4pt]
    \begin{subfigure}[t]{\qualimgw}
        \qualimg{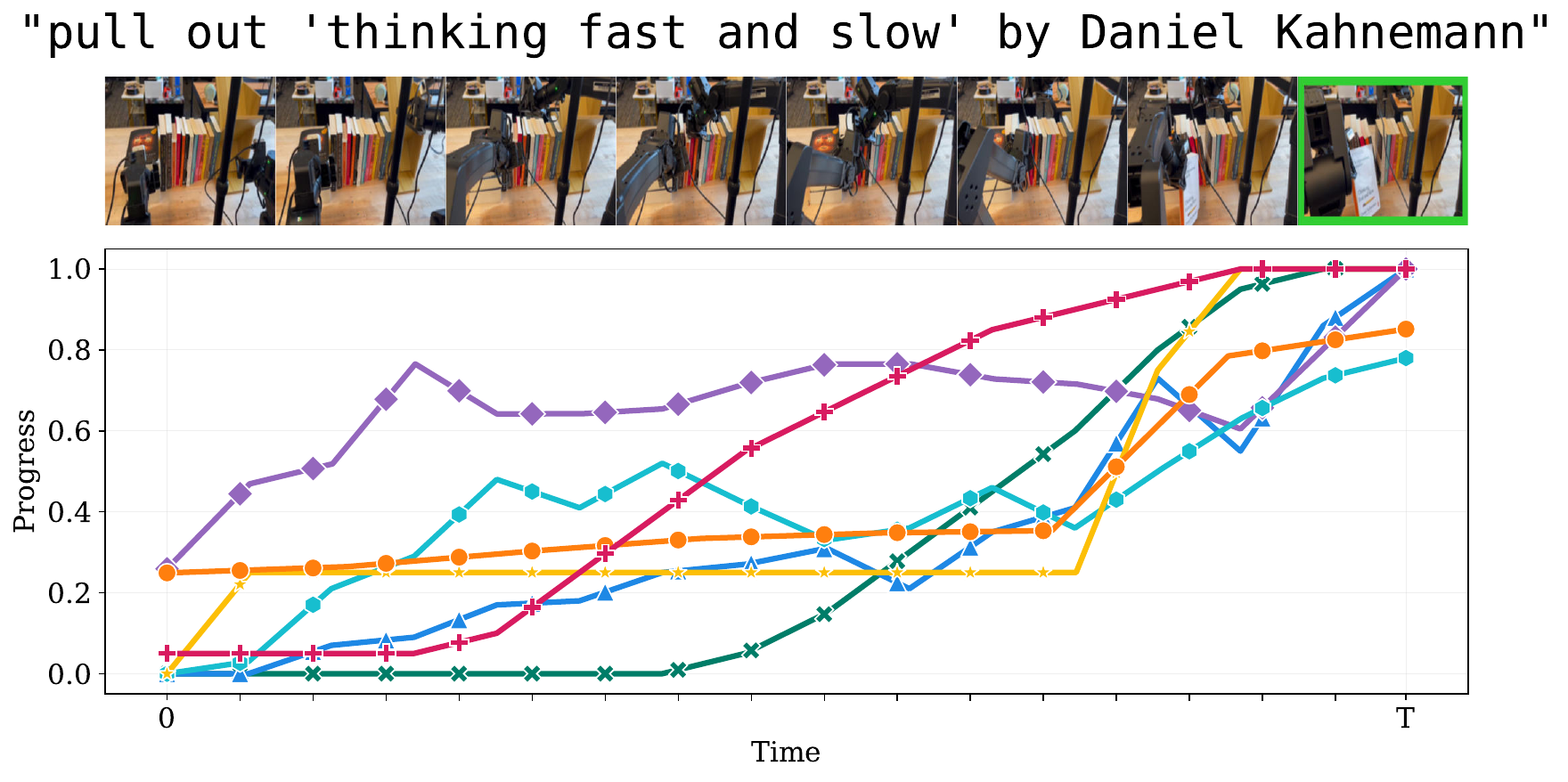}
        \caption{``pull out `Thinking, Fast and Slow' by Daniel Kahnemann'' (success).}
    \end{subfigure}
    \hfill
    \begin{subfigure}[t]{\qualimgw}
        \qualimg{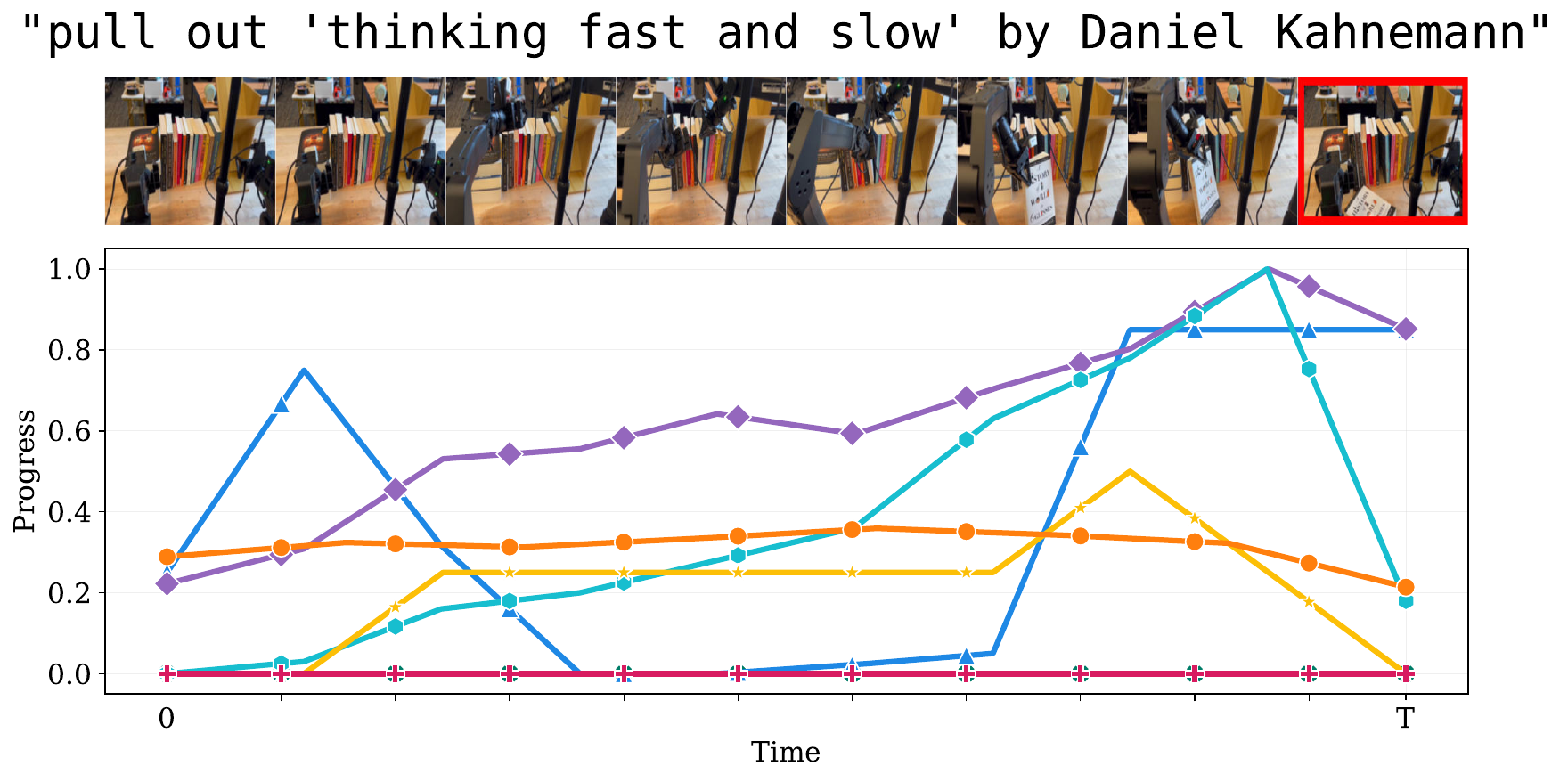}
        \caption{``pull out `Thinking, Fast and Slow' by Daniel Kahnemann'' (failure).}
    \end{subfigure}
    \caption{\textbf{Qualitative per-trajectory predictions for ``pull out `Thinking, Fast and Slow'\,''.}}
    \label{fig:qualitative_book_task}
\end{figure*}

\begin{figure*}[!h]
    \centering
    \includegraphics[width=0.6\columnwidth]{plots/legend.pdf}\\[4pt]
    \begin{subfigure}[t]{\qualimgw}
        \qualimg{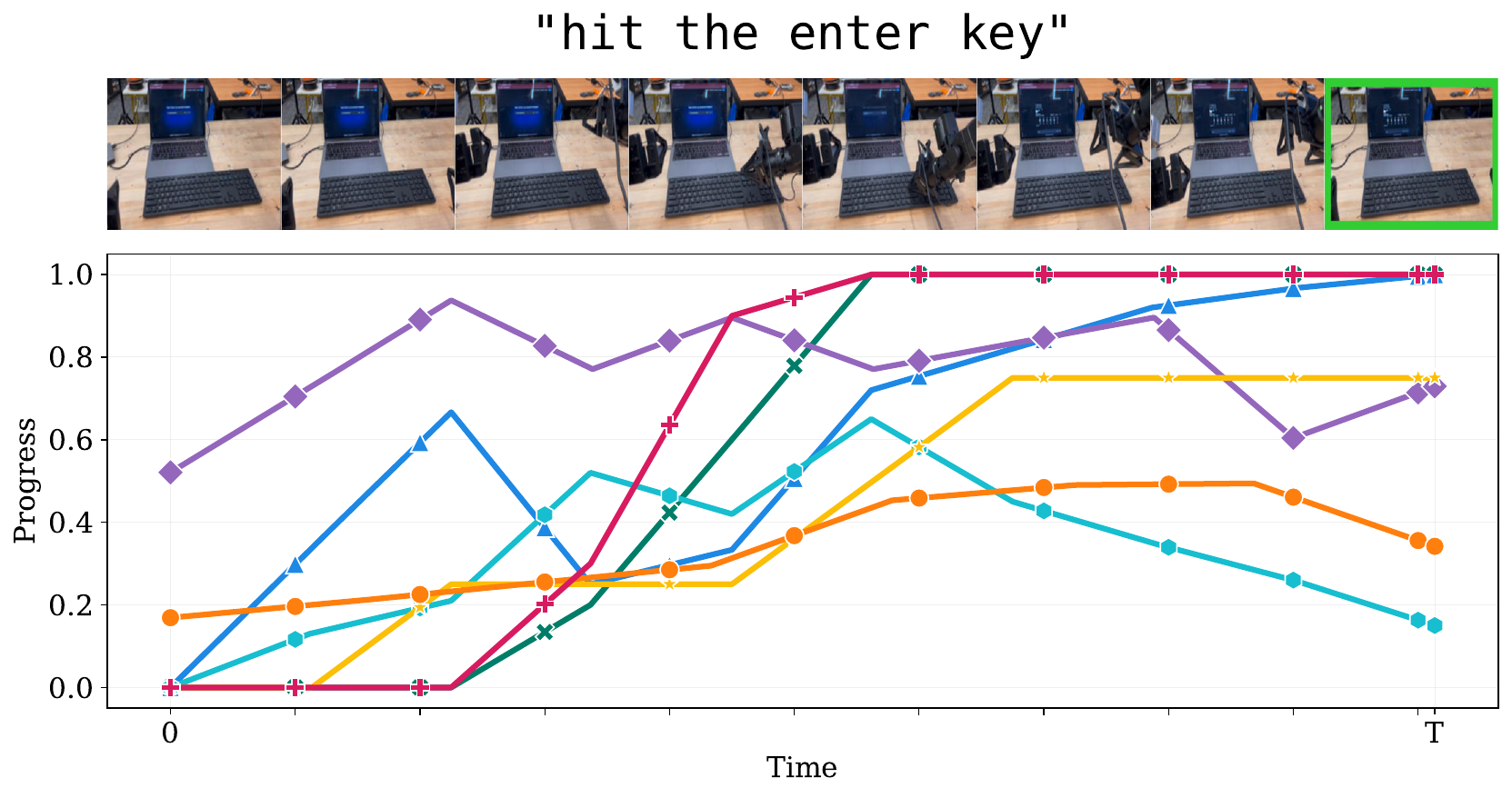}
        \caption{``hit the enter key'' (success).}
    \end{subfigure}
    \hfill
    \begin{subfigure}[t]{\qualimgw}
        \qualimg{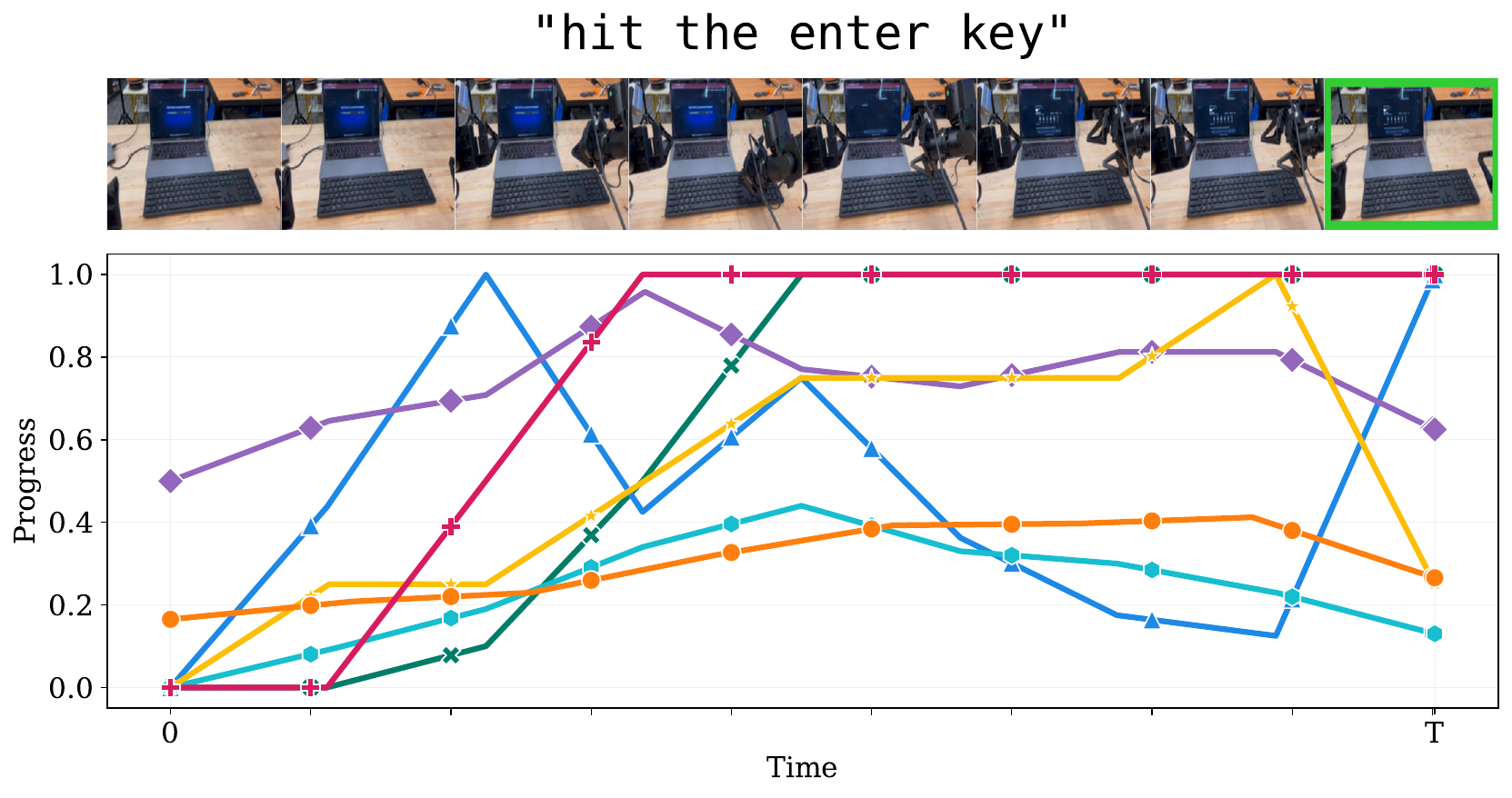}
        \caption{``hit the enter key'' (success).}
    \end{subfigure}

    \vspace{0.5em}
    \begin{subfigure}[t]{\qualimgw}
        \qualimg{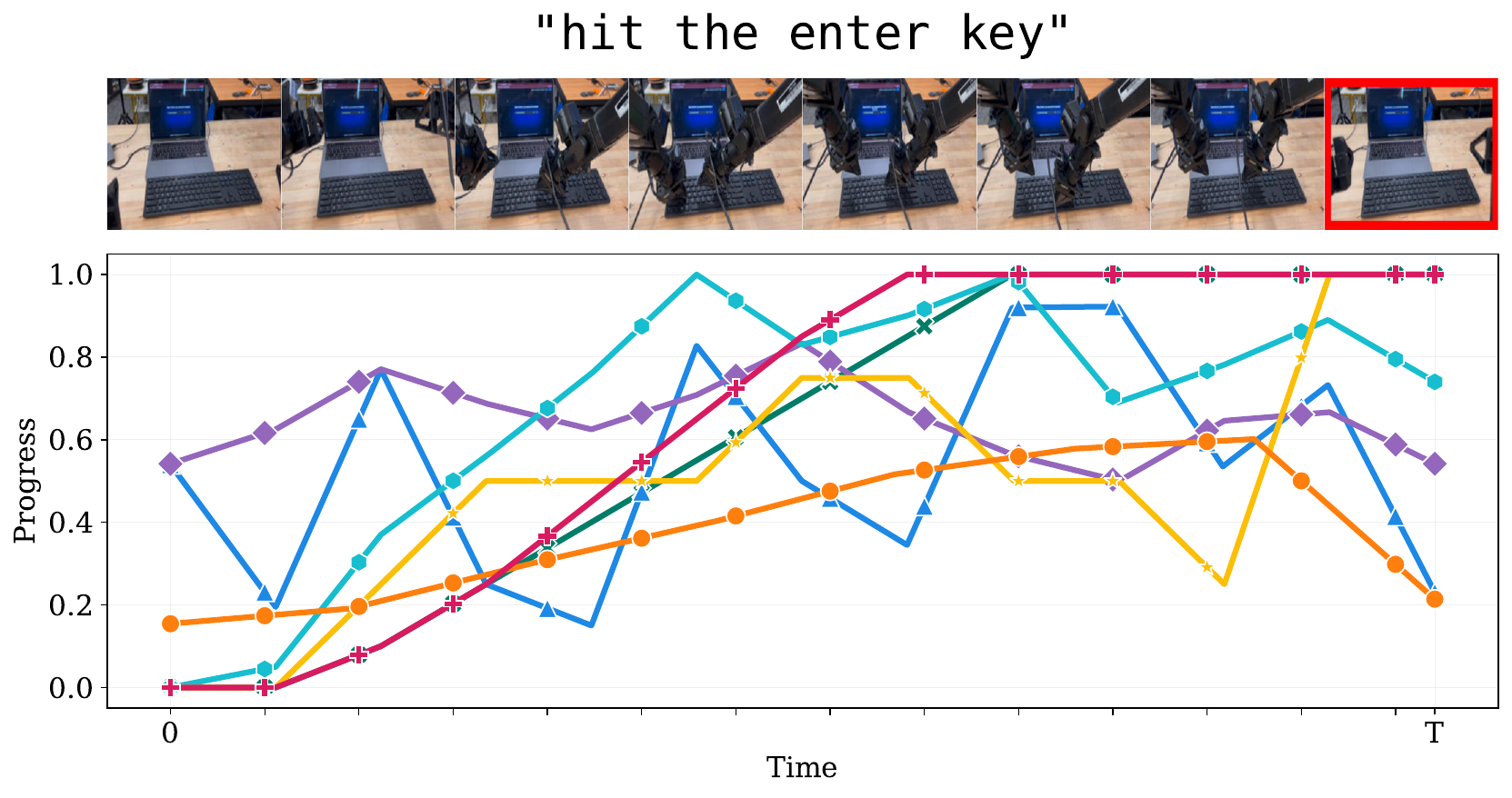}
        \caption{``hit the enter key'' (failure).}
    \end{subfigure}
    \hfill
    \begin{subfigure}[t]{\qualimgw}
        \qualimg{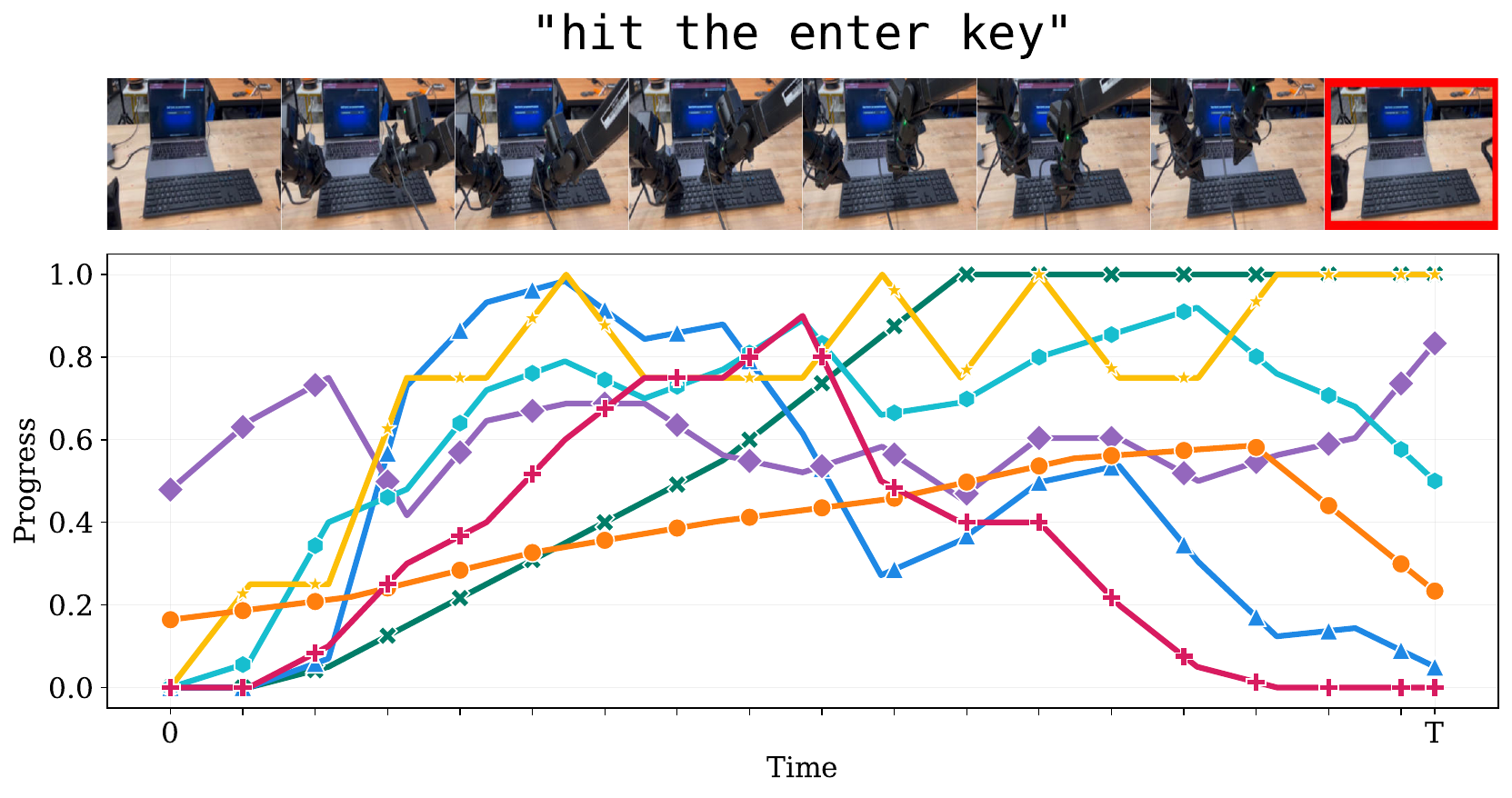}
        \caption{``hit the enter key'' (failure).}
    \end{subfigure}
    \caption{\textbf{Qualitative per-trajectory predictions for ``hit the enter key''.}}
    \label{fig:qualitative_enter_task}
\end{figure*}

\begin{figure*}[!h]
    \centering
    \includegraphics[width=0.6\columnwidth]{plots/legend.pdf}\\[4pt]
    \begin{subfigure}[t]{\qualimgw}
        \qualimg{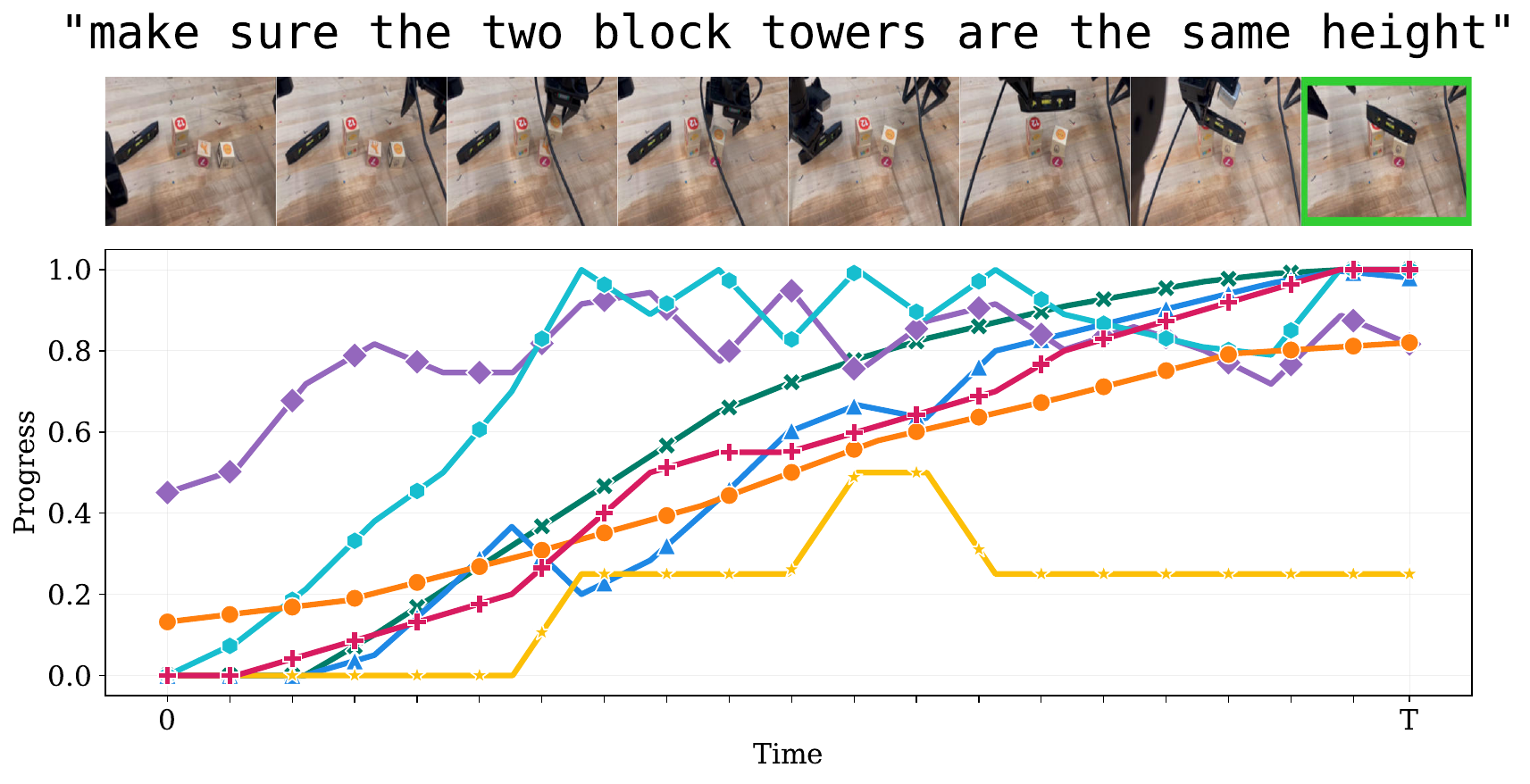}
        \caption{``make sure the two block towers are the same height'' (success).}
    \end{subfigure}
    \hfill
    \begin{subfigure}[t]{\qualimgw}
        \qualimg{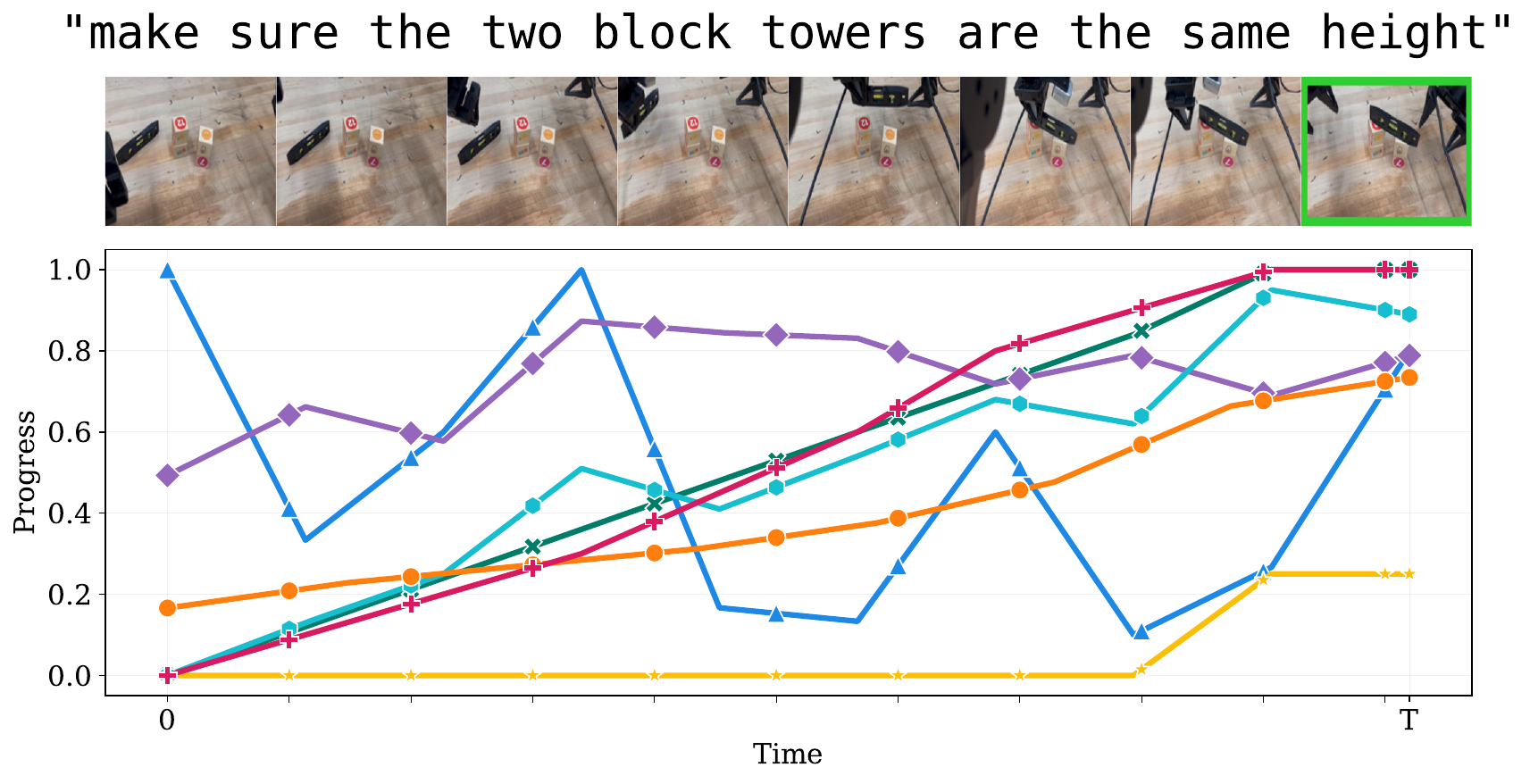}
        \caption{``make sure the two block towers are the same height'' (success).}
    \end{subfigure}

    \vspace{0.5em}
    \begin{subfigure}[t]{\qualimgw}
        \qualimg{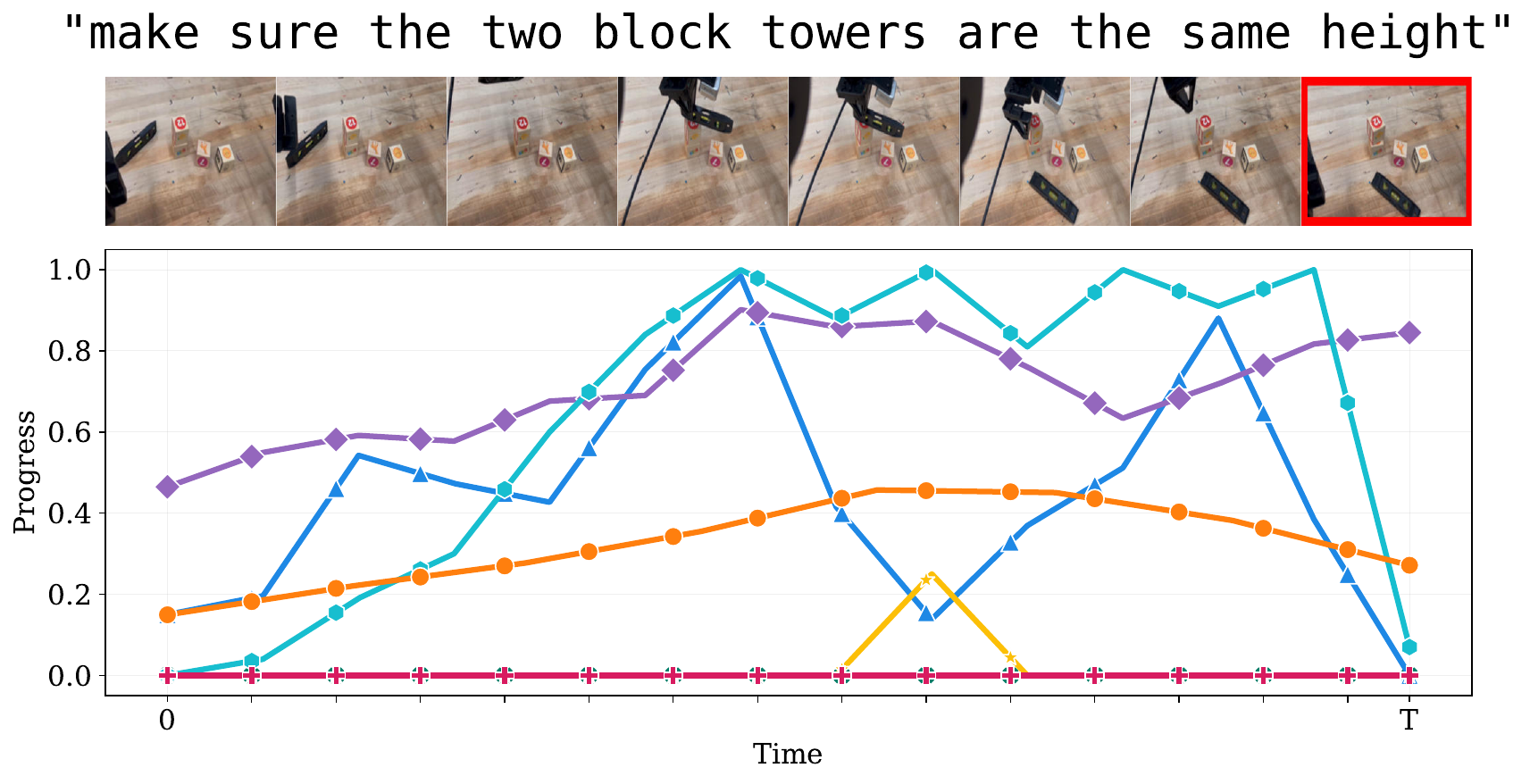}
        \caption{``make sure the two block towers are the same height'' (failure).}
    \end{subfigure}
    \hfill
    \begin{subfigure}[t]{\qualimgw}
        \qualimg{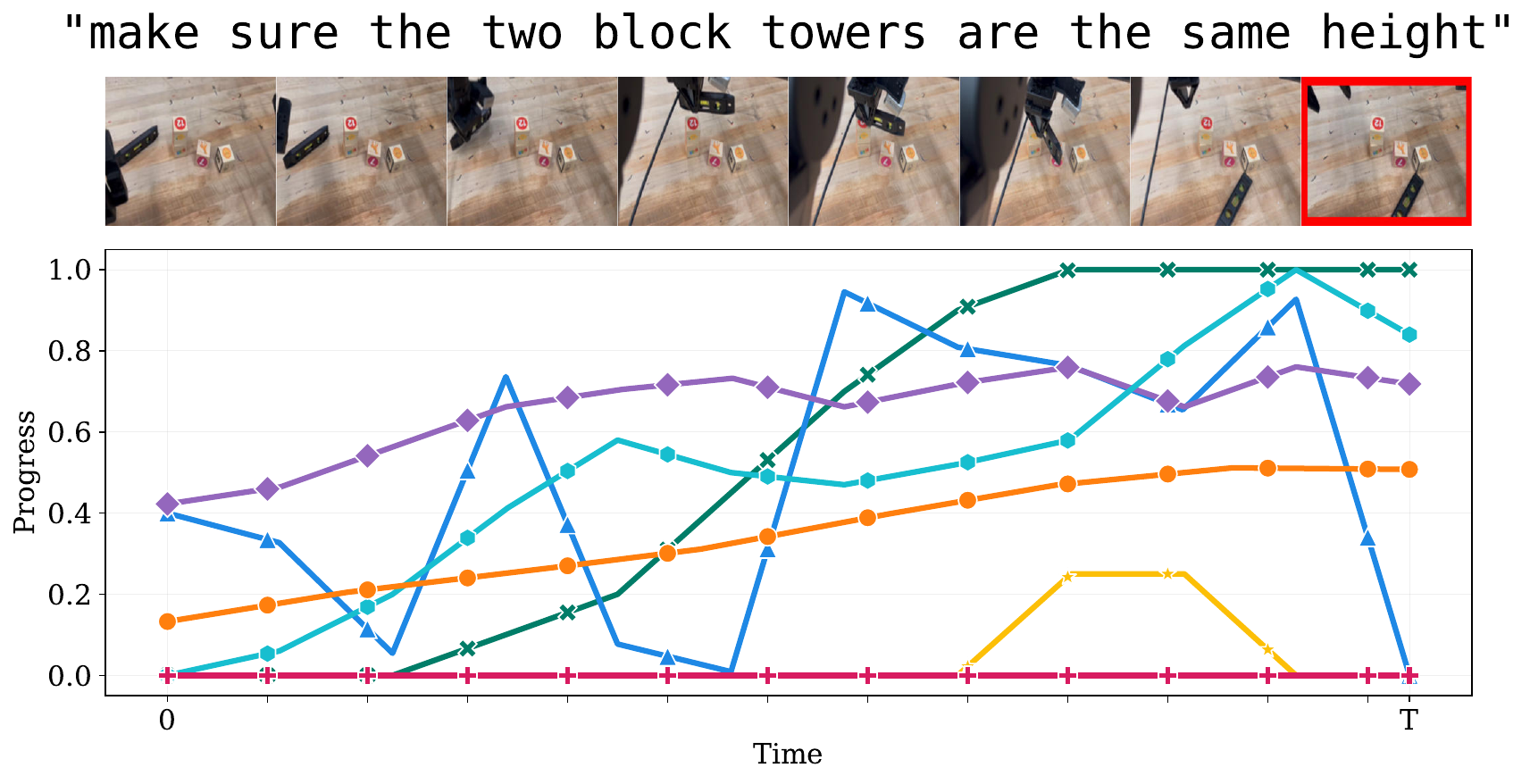}
        \caption{``make sure the two block towers are the same height'' (failure).}
    \end{subfigure}
    \caption{\textbf{Qualitative per-trajectory predictions for ``make sure the two block towers are the same height''.}}
    \label{fig:qualitative_level_task}
\end{figure*}

\begin{figure*}[!h]
    \centering
    \includegraphics[width=0.6\columnwidth]{plots/legend.pdf}\\[4pt]
    \begin{subfigure}[t]{\qualimgw}
        \qualimg{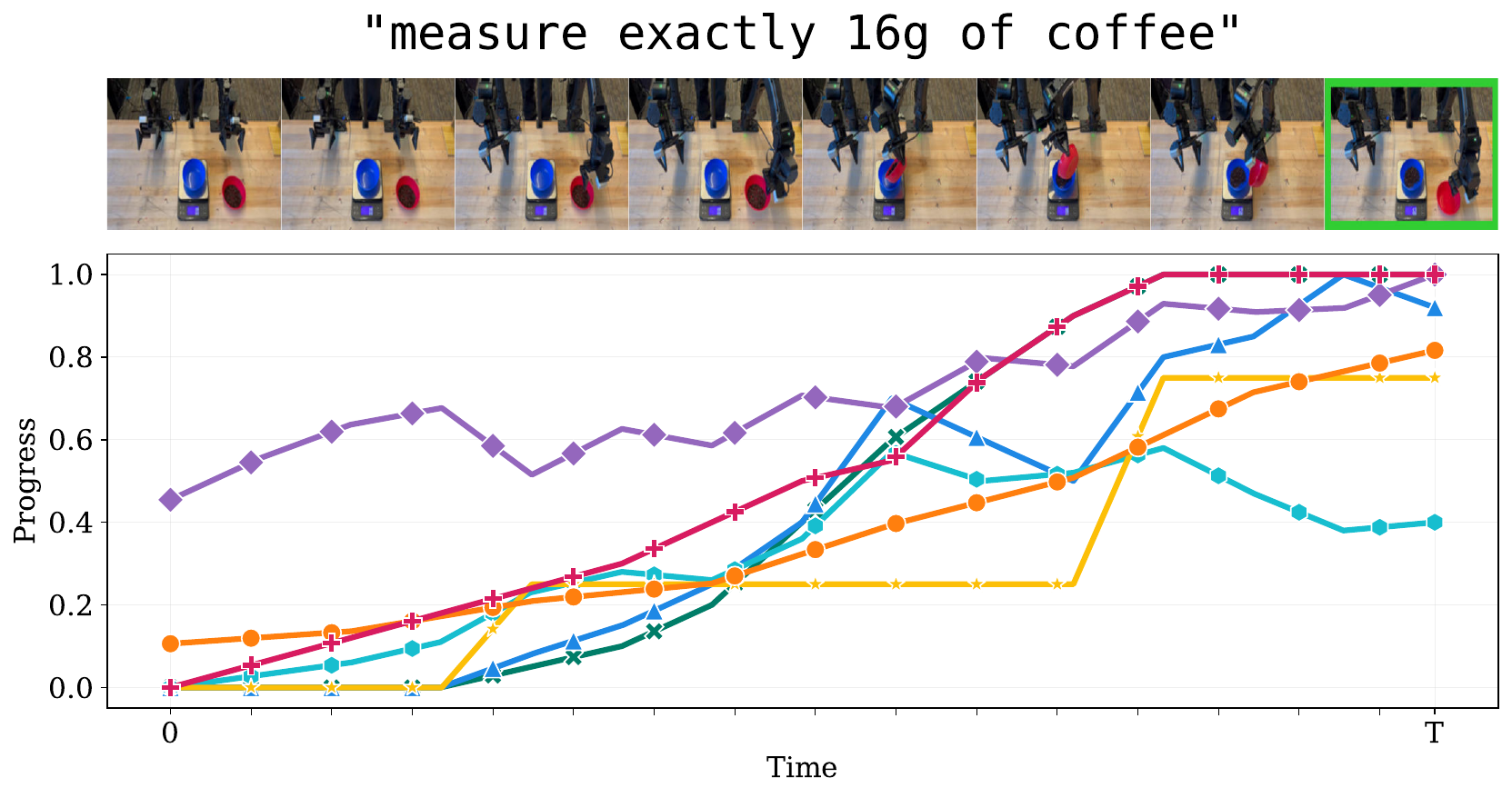}
        \caption{``measure exactly 16g of coffee'' (success).}
    \end{subfigure}
    \hfill
    \begin{subfigure}[t]{\qualimgw}
        \qualimg{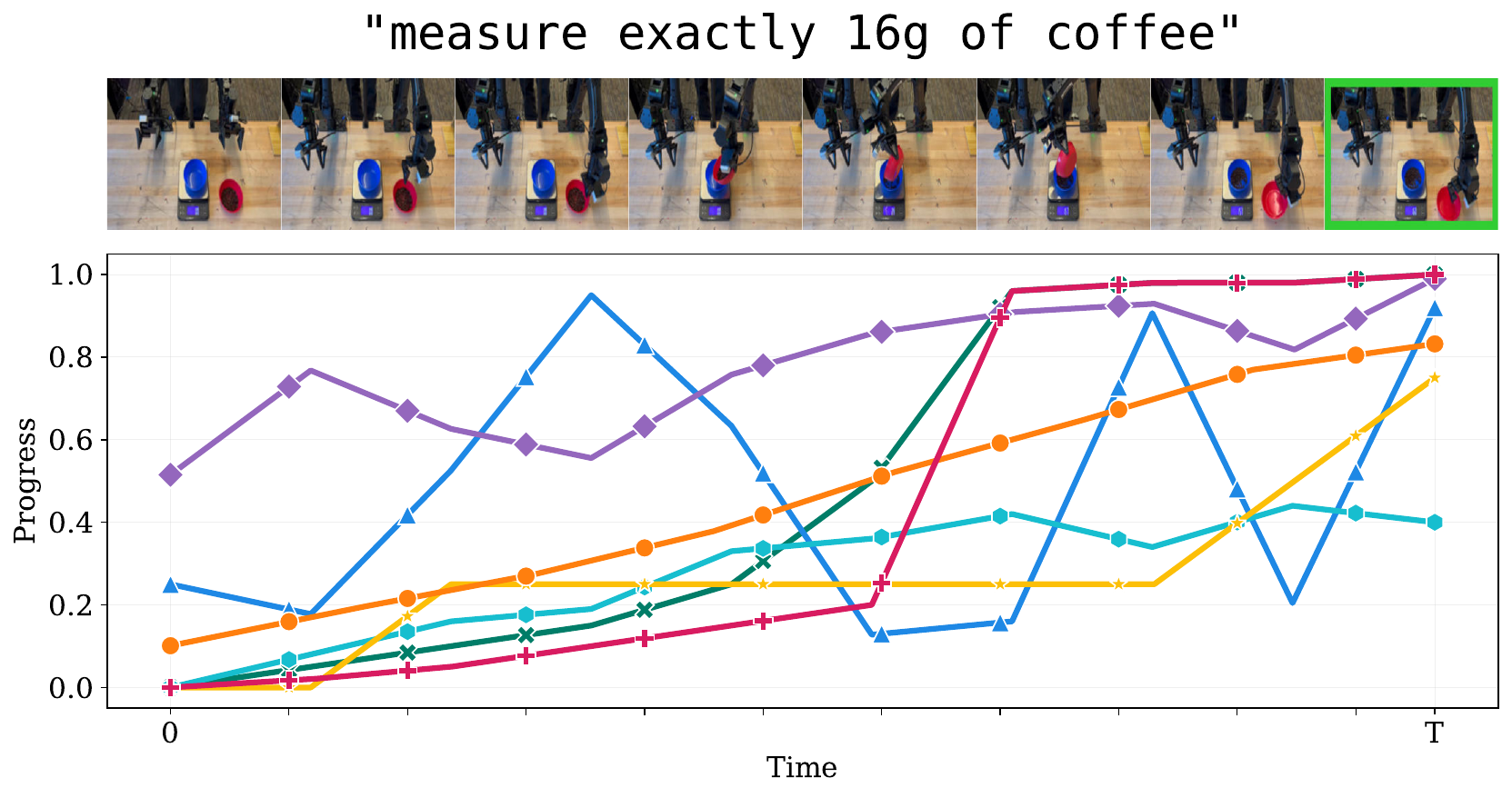}
        \caption{``measure exactly 16g of coffee'' (success).}
    \end{subfigure}

    \vspace{0.5em}
    \begin{subfigure}[t]{\qualimgw}
        \qualimg{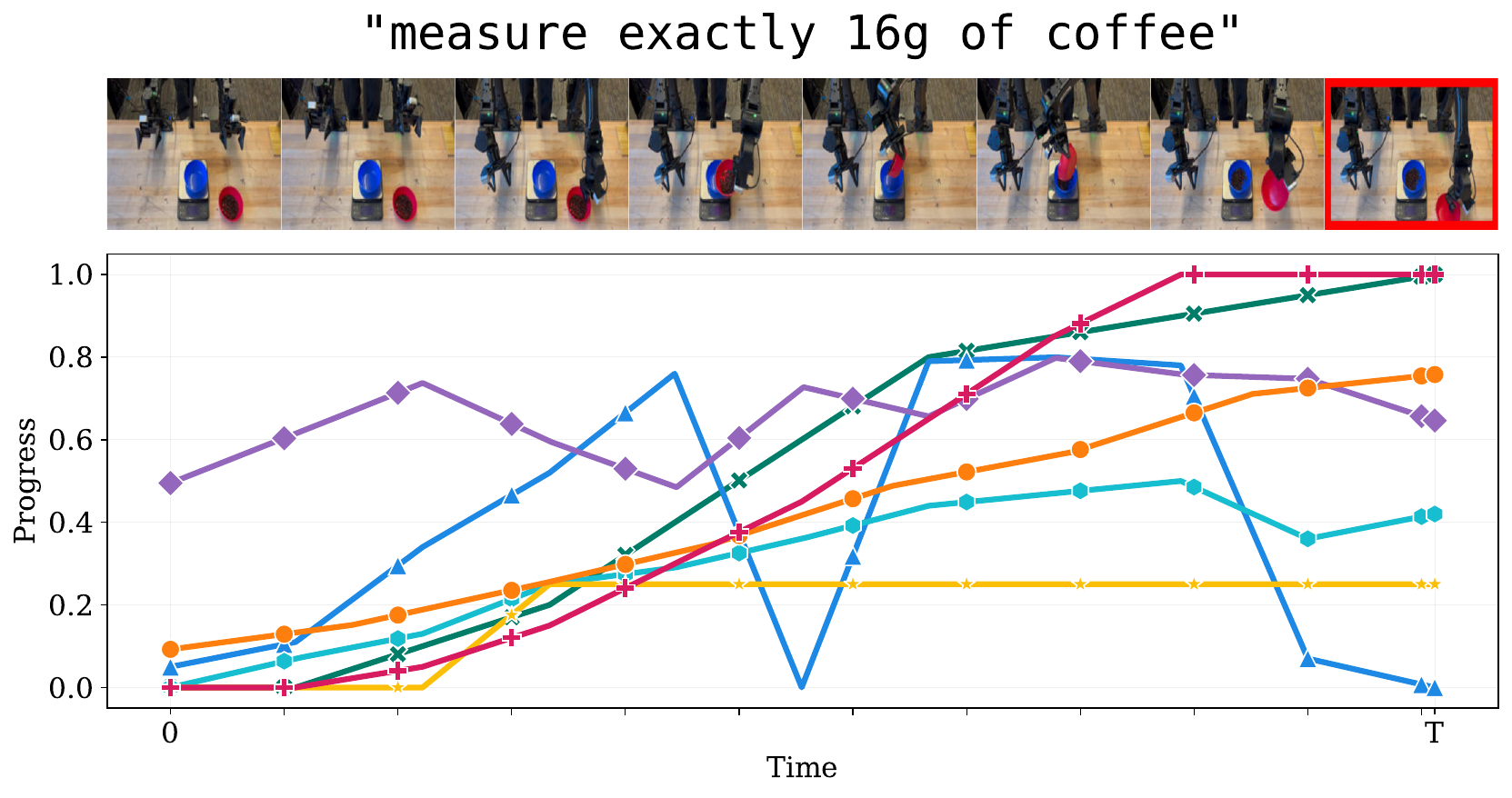}
        \caption{``measure exactly 16g of coffee'' (failure).}
    \end{subfigure}
    \hfill
    \begin{subfigure}[t]{\qualimgw}
        \qualimg{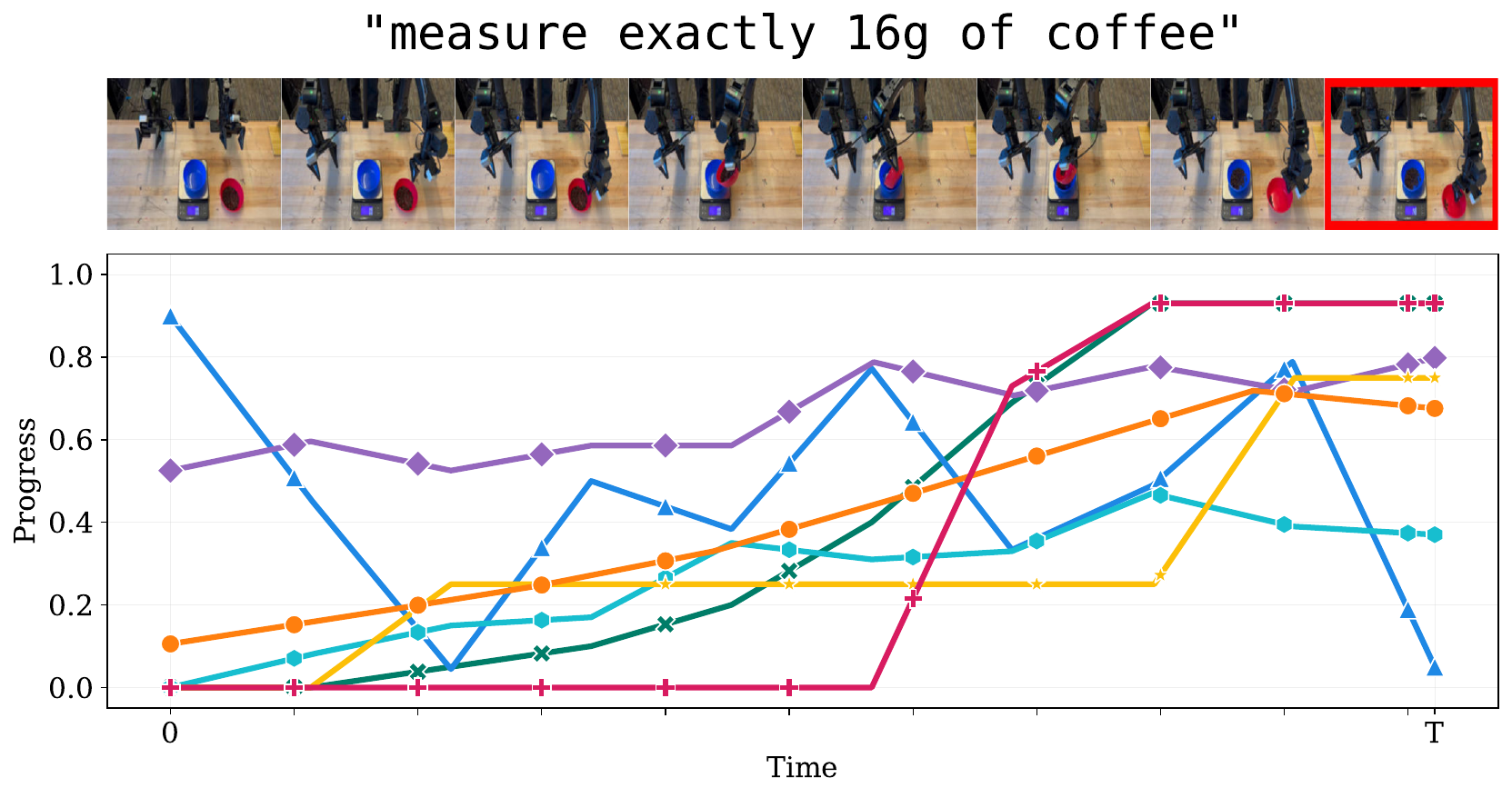}
        \caption{``measure exactly 16g of coffee'' (failure).}
    \end{subfigure}
    \caption{\textbf{Qualitative per-trajectory predictions for ``measure exactly 16g of coffee''.}}
    \label{fig:qualitative_coffee_task}
\end{figure*}

\begin{figure*}[!h]
    \centering
    \includegraphics[width=0.6\columnwidth]{plots/legend.pdf}\\[4pt]
    \begin{subfigure}[t]{\qualimgw}
        \qualimg{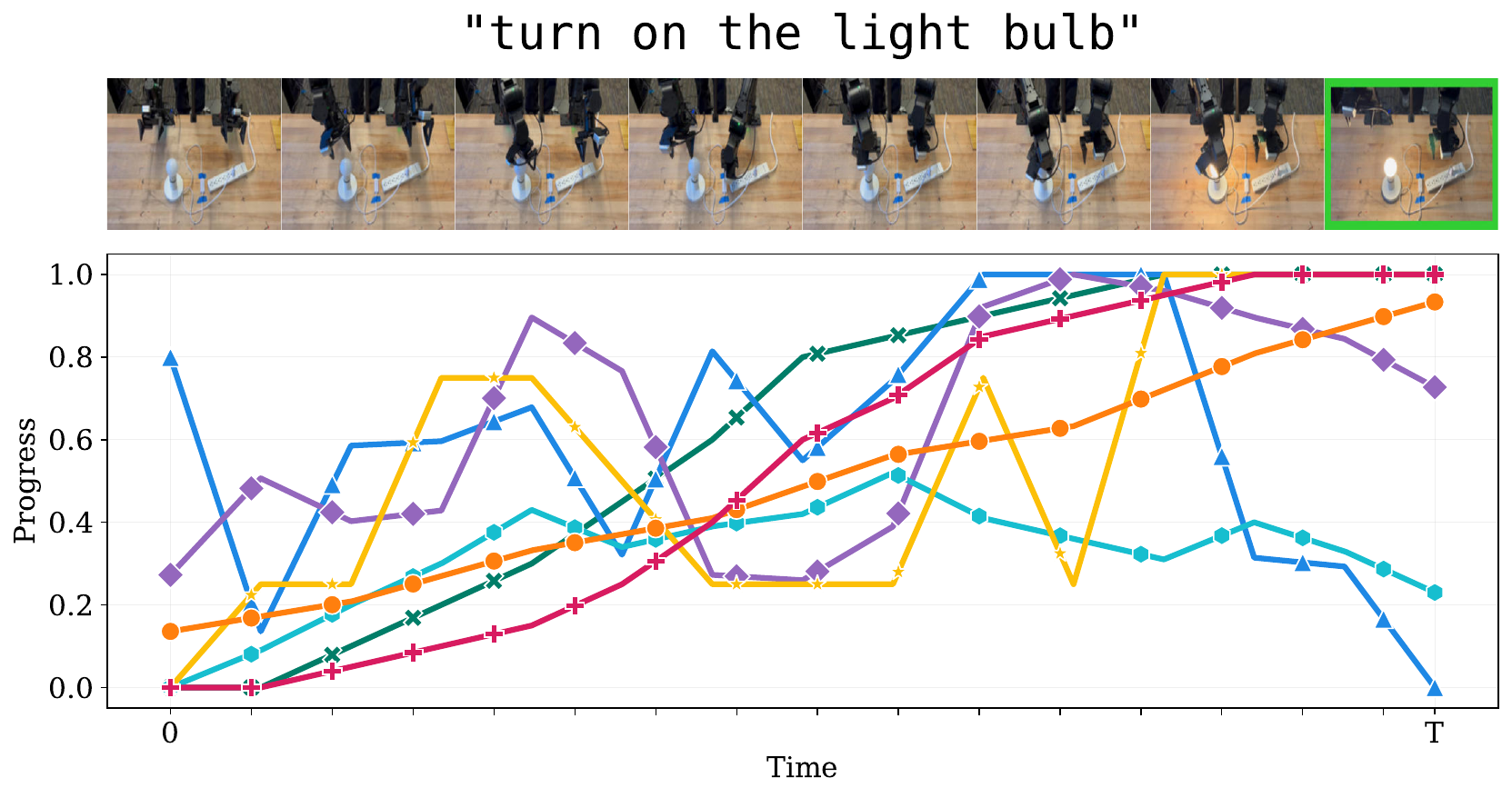}
        \caption{``turn on the light bulb'' (success).}
    \end{subfigure}
    \hfill
    \begin{subfigure}[t]{\qualimgw}
        \qualimg{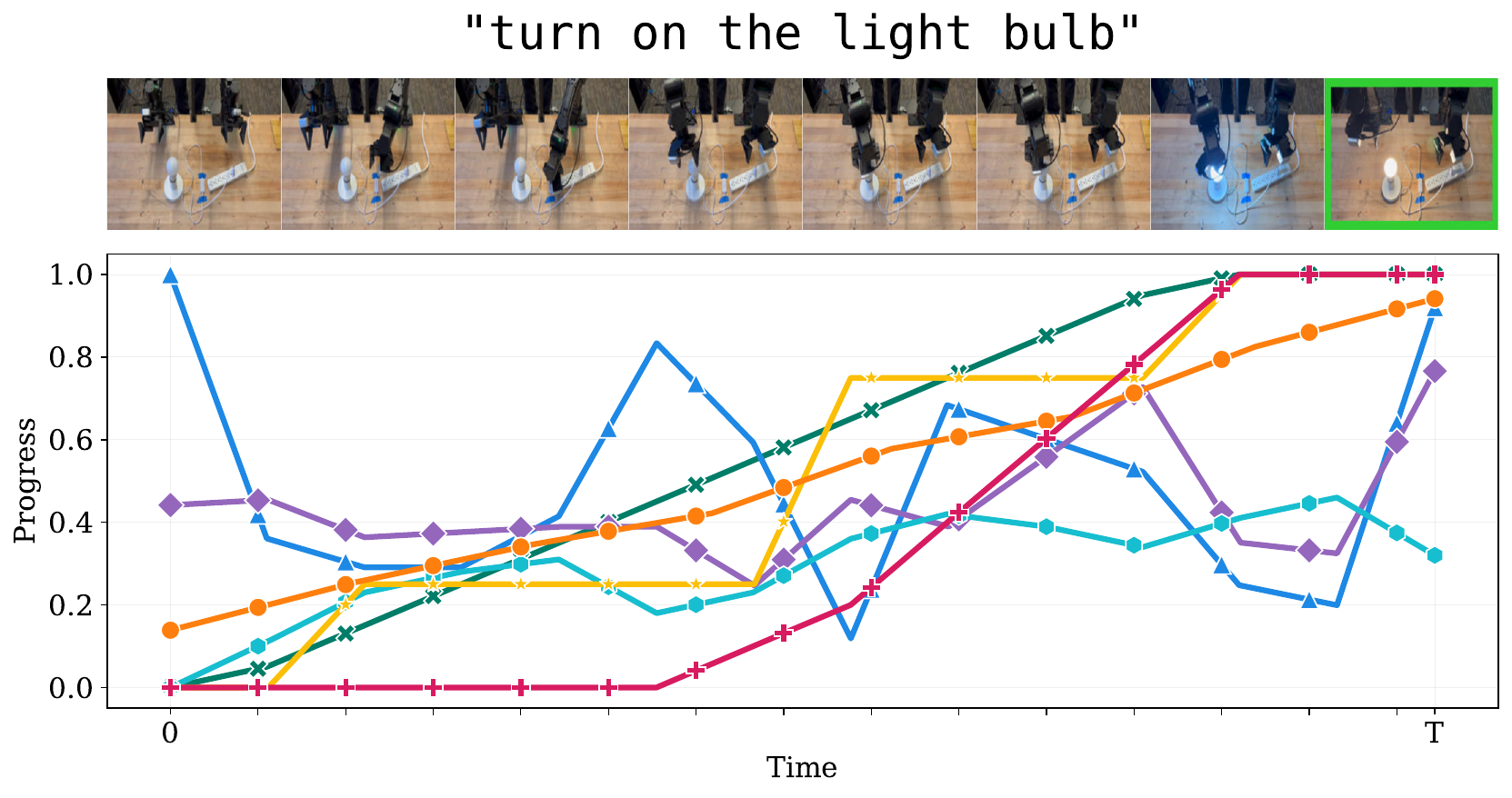}
        \caption{``turn on the light bulb'' (success).}
    \end{subfigure}

    \vspace{0.5em}
    \begin{subfigure}[t]{\qualimgw}
        \qualimg{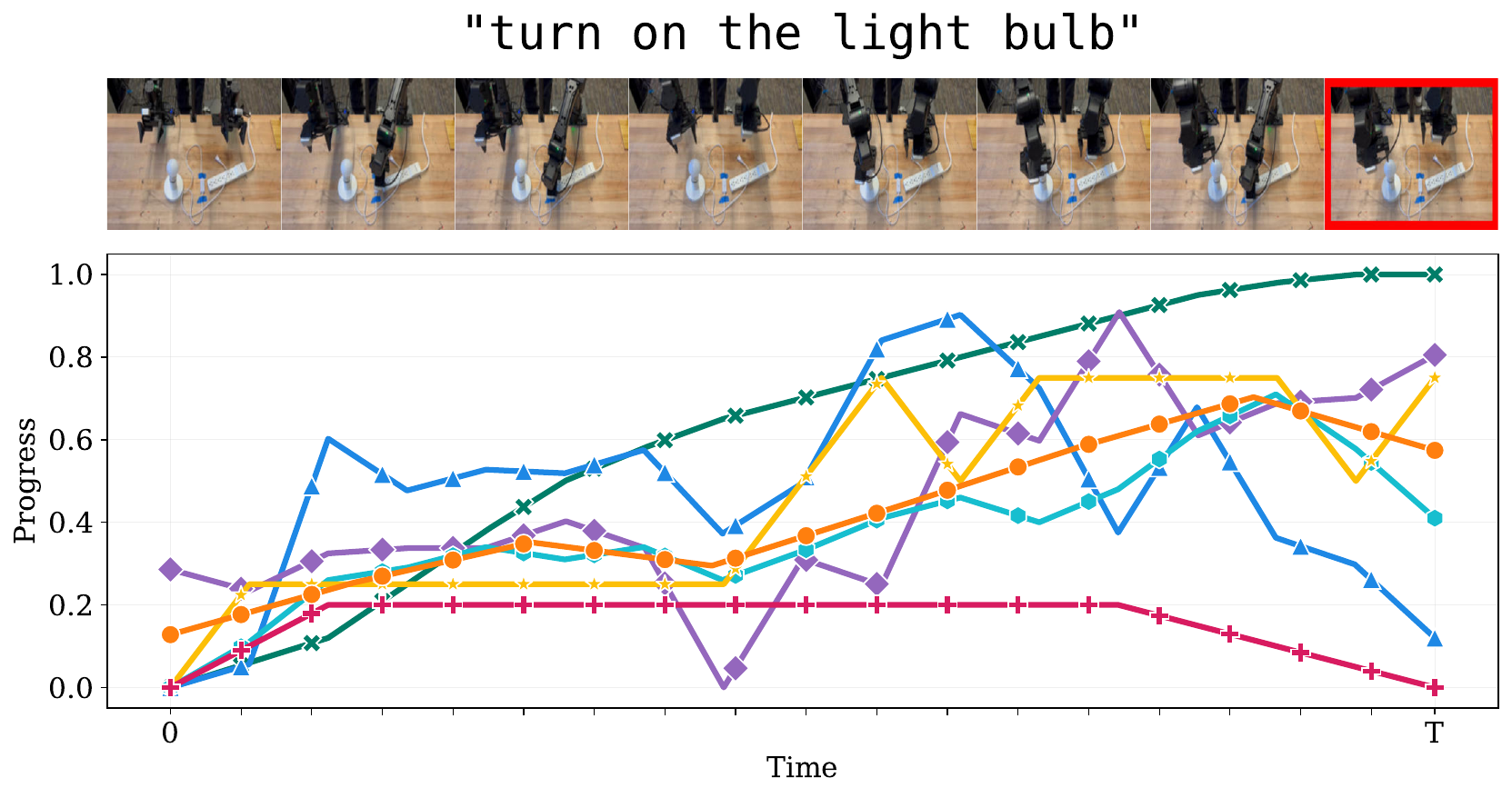}
        \caption{``turn on the light bulb'' (failure).}
    \end{subfigure}
    \hfill
    \begin{subfigure}[t]{\qualimgw}
        \qualimg{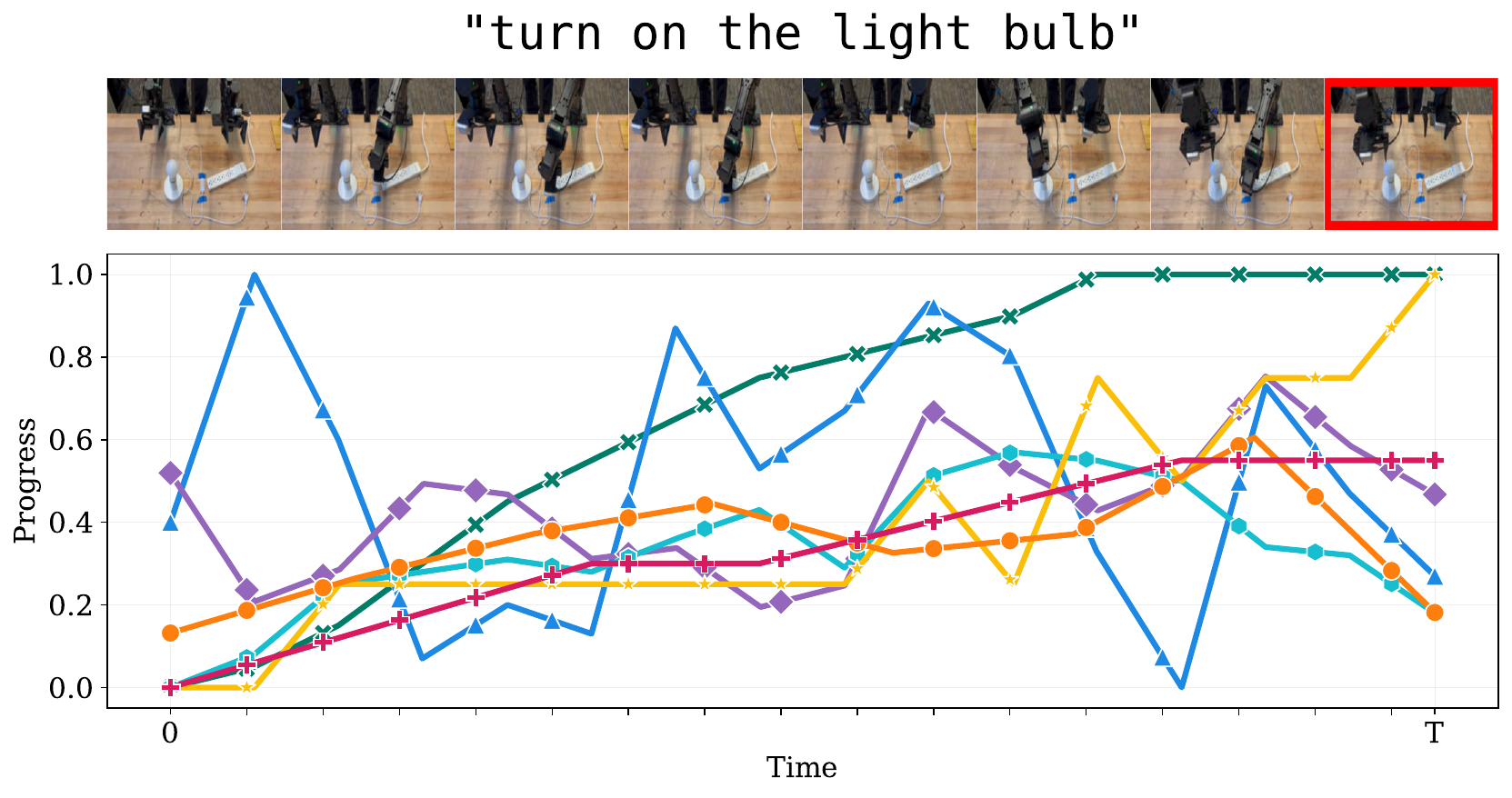}
        \caption{``turn on the light bulb'' (failure).}
    \end{subfigure}
    \caption{\textbf{Qualitative per-trajectory predictions for ``turn on the light bulb''.}}
    \label{fig:qualitative_light_task}
\end{figure*}

\begin{figure*}[!h]
    \centering
    \includegraphics[width=0.6\columnwidth]{plots/legend.pdf}\\[4pt]
    \begin{subfigure}[t]{\qualimgw}
        \qualimg{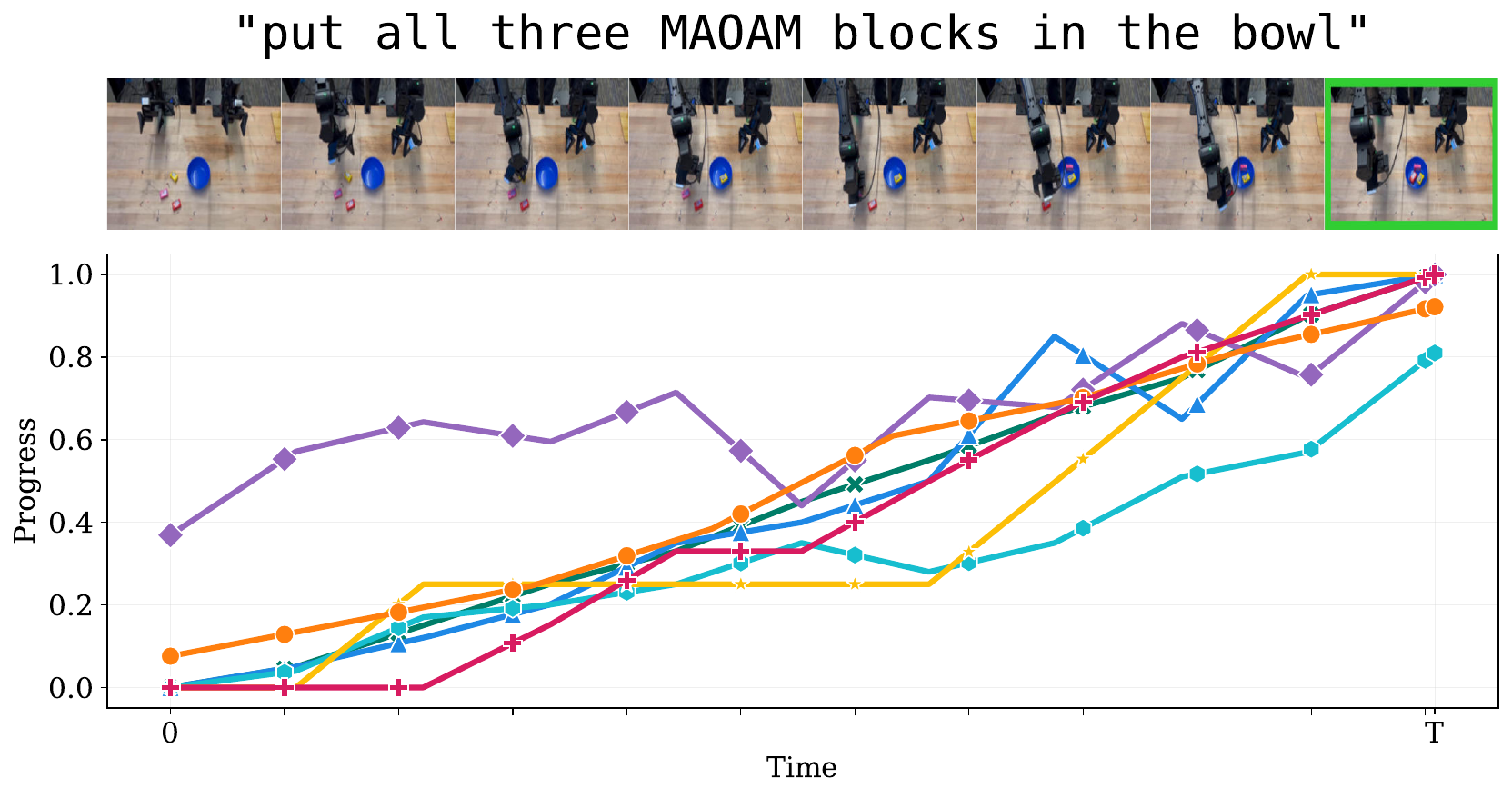}
        \caption{``put all three MAOAM blocks in the bowl'' (success).}
    \end{subfigure}
    \hfill
    \begin{subfigure}[t]{\qualimgw}
        \qualimg{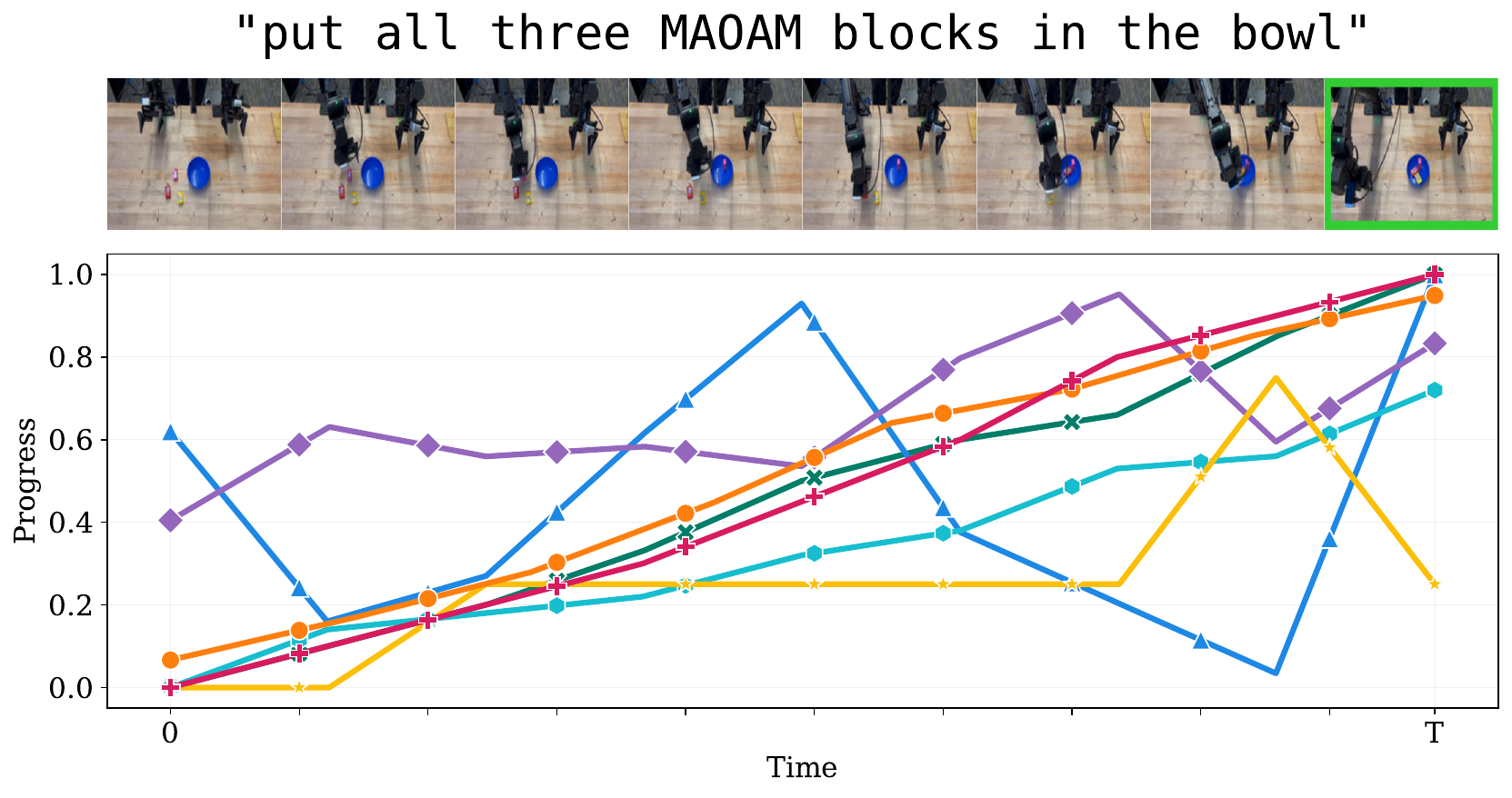}
        \caption{``put all three MAOAM blocks in the bowl'' (success).}
    \end{subfigure}

    \vspace{0.5em}
    \begin{subfigure}[t]{\qualimgw}
        \qualimg{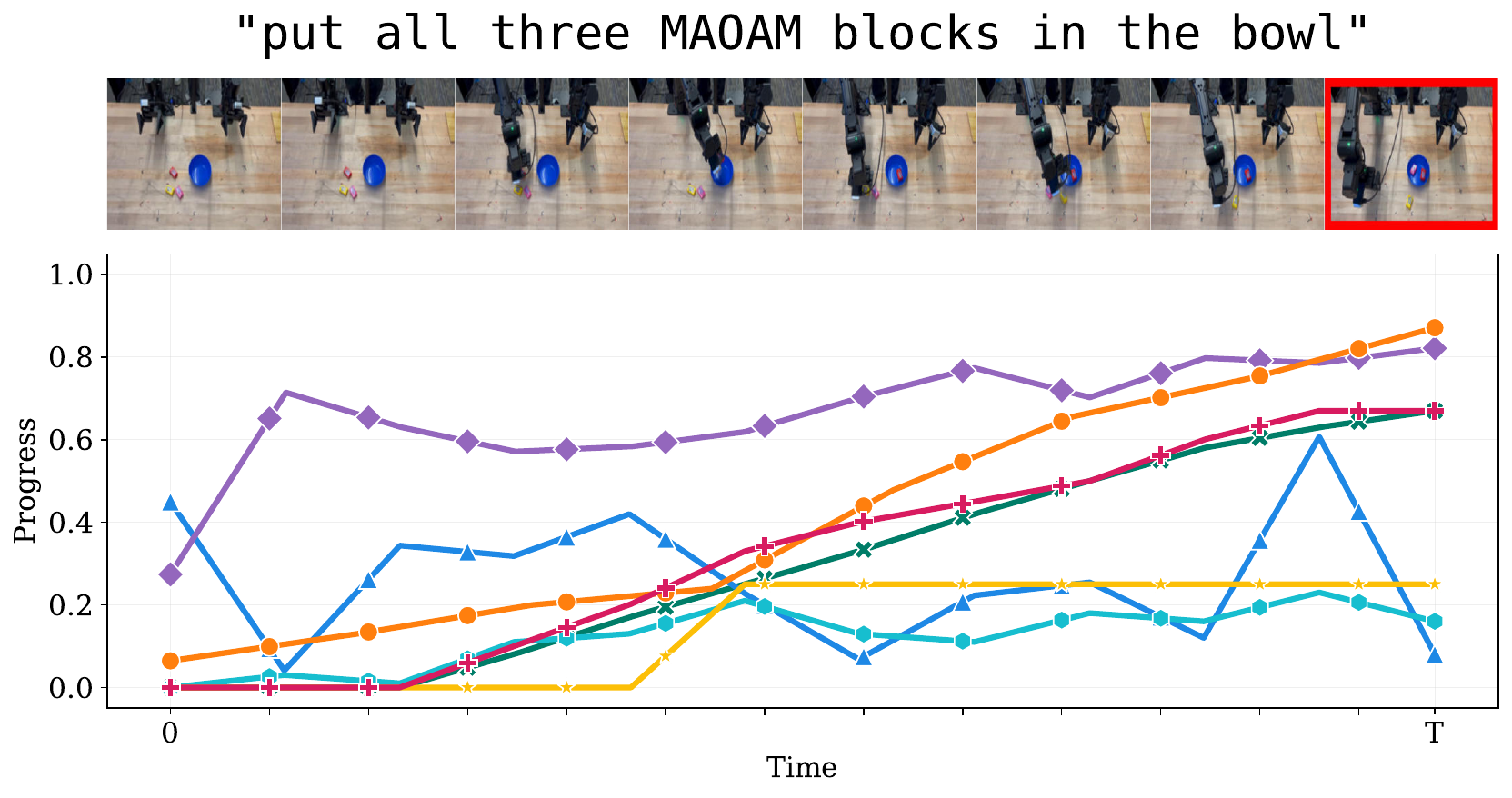}
        \caption{``put all three MAOAM blocks in the bowl'' (failure).}
    \end{subfigure}
    \hfill
    \begin{subfigure}[t]{\qualimgw}
        \qualimg{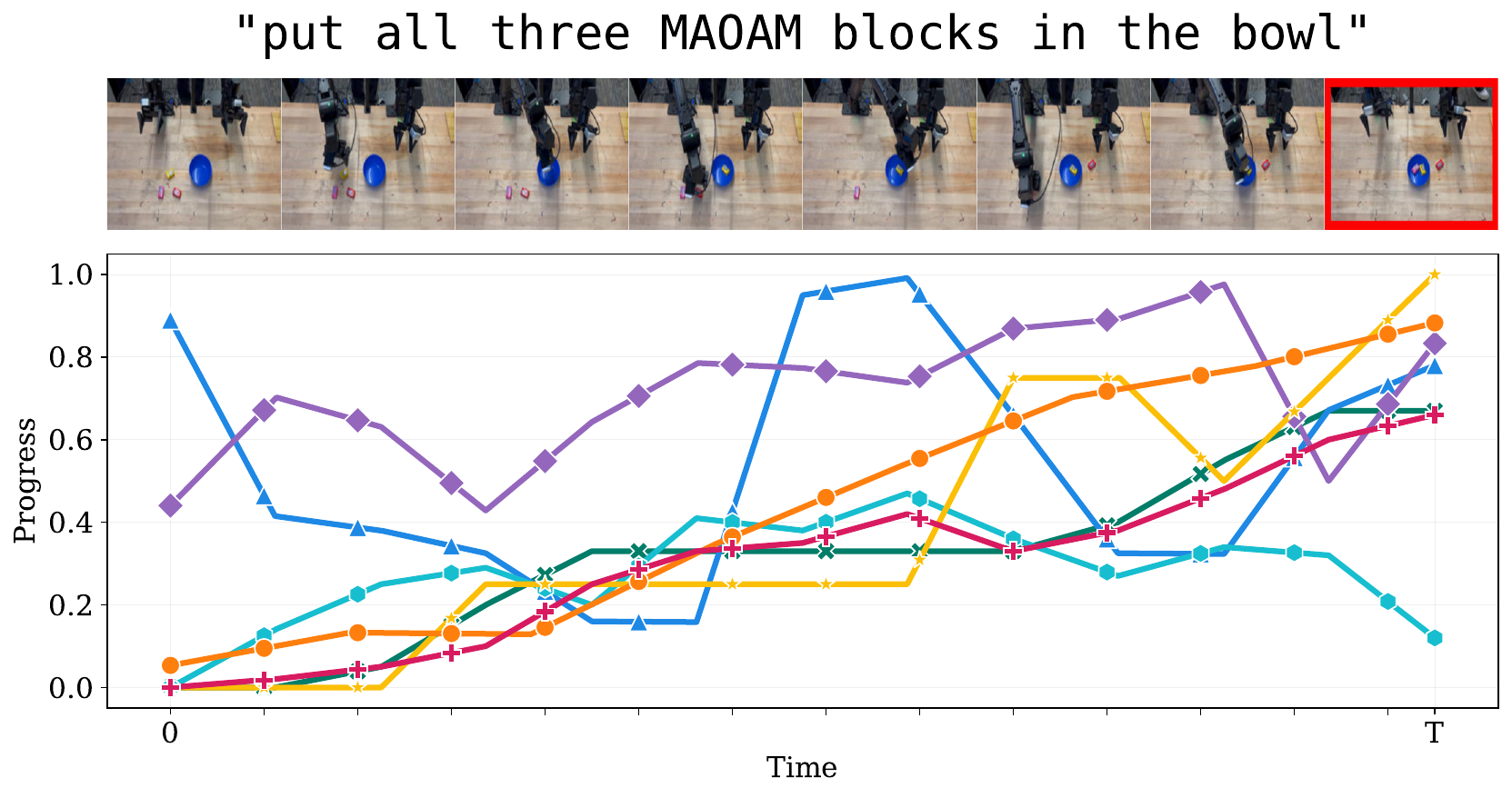}
        \caption{``put all three MAOAM blocks in the bowl'' (failure).}
    \end{subfigure}
    \caption{\textbf{Qualitative per-trajectory predictions for ``put all three MAOAM blocks in the bowl''.}}
    \label{fig:qualitative_maoam_task}
\end{figure*}

\begin{figure*}[!h]
    \centering
    \includegraphics[width=0.6\columnwidth]{plots/legend.pdf}\\[4pt]
    \begin{subfigure}[t]{\qualimgw}
        \qualimg{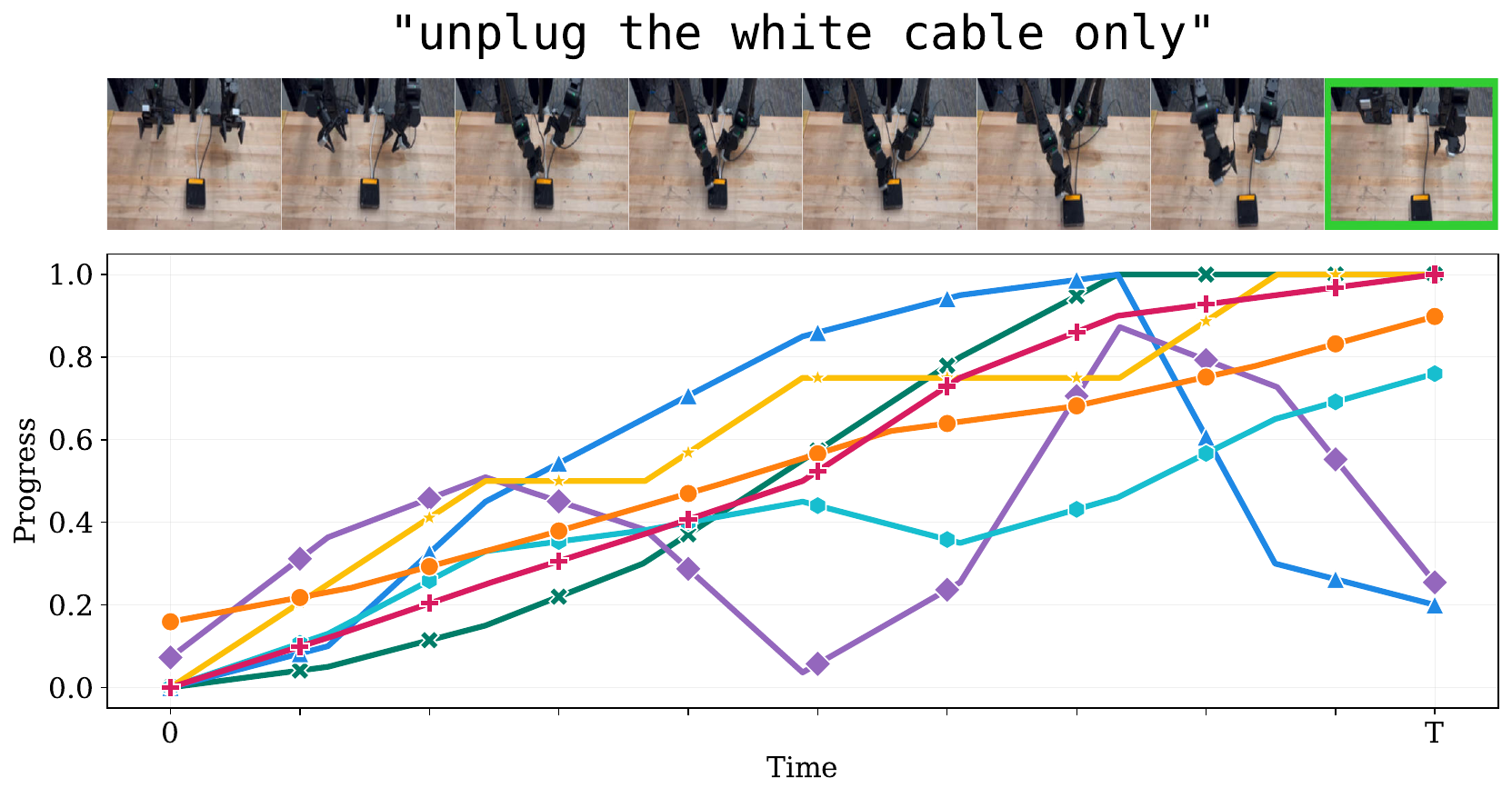}
        \caption{``unplug the white cable only'' (success).}
    \end{subfigure}
    \hfill
    \begin{subfigure}[t]{\qualimgw}
        \qualimg{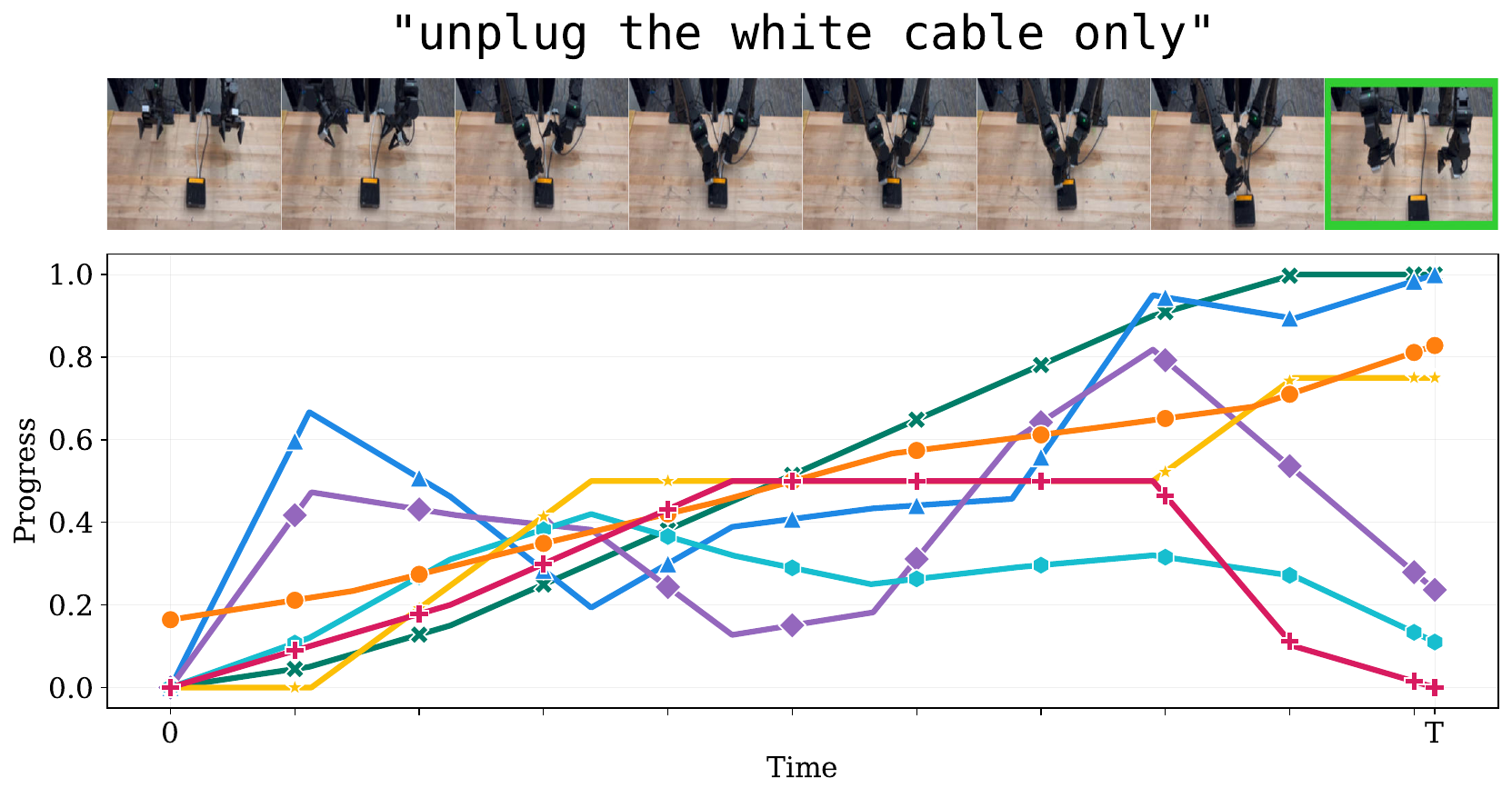}
        \caption{``unplug the white cable only'' (success).}
    \end{subfigure}

    \vspace{0.5em}
    \begin{subfigure}[t]{\qualimgw}
        \qualimg{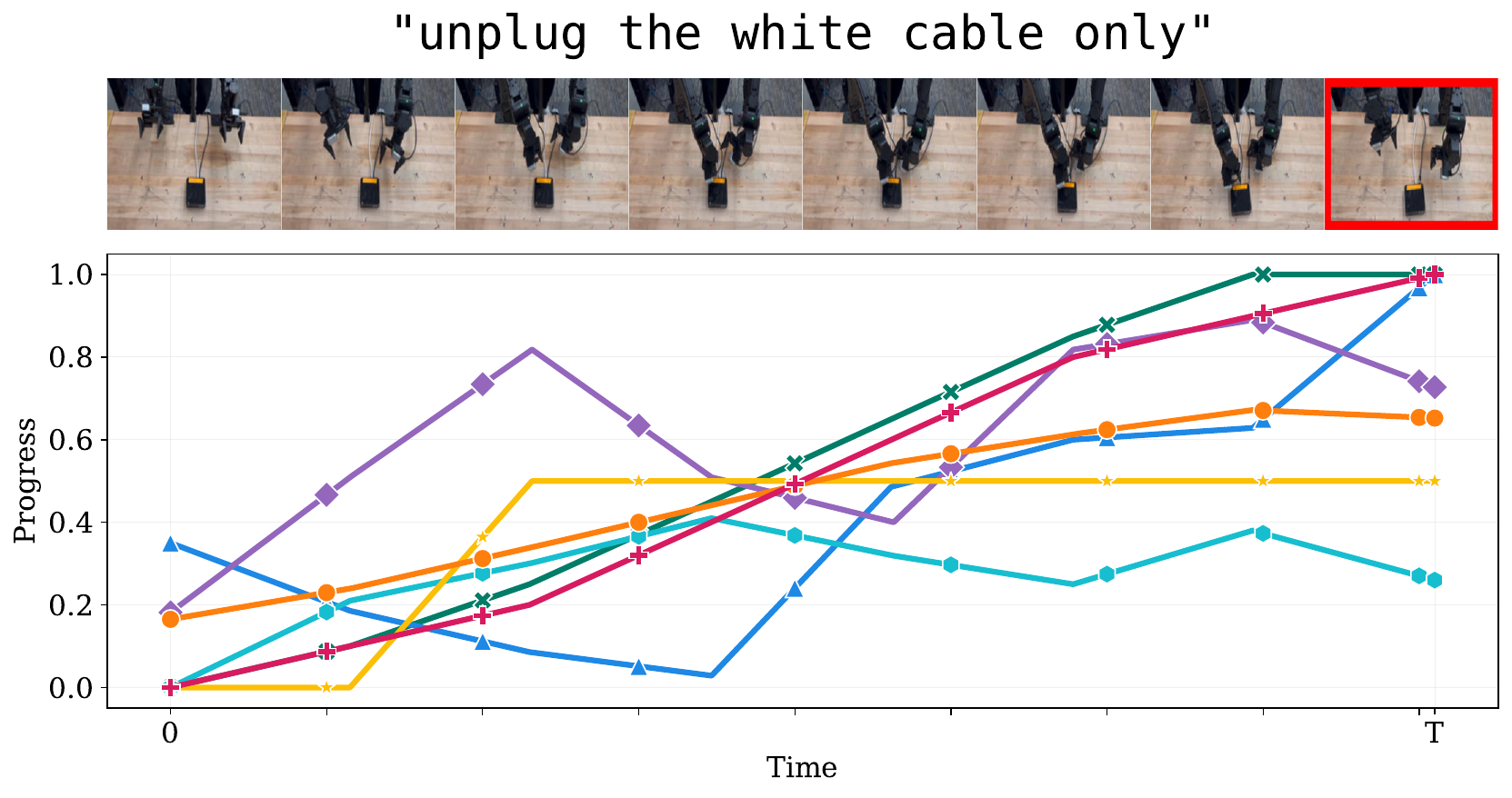}
        \caption{``unplug the white cable only'' (failure).}
    \end{subfigure}
    \hfill
    \begin{subfigure}[t]{\qualimgw}
        \qualimg{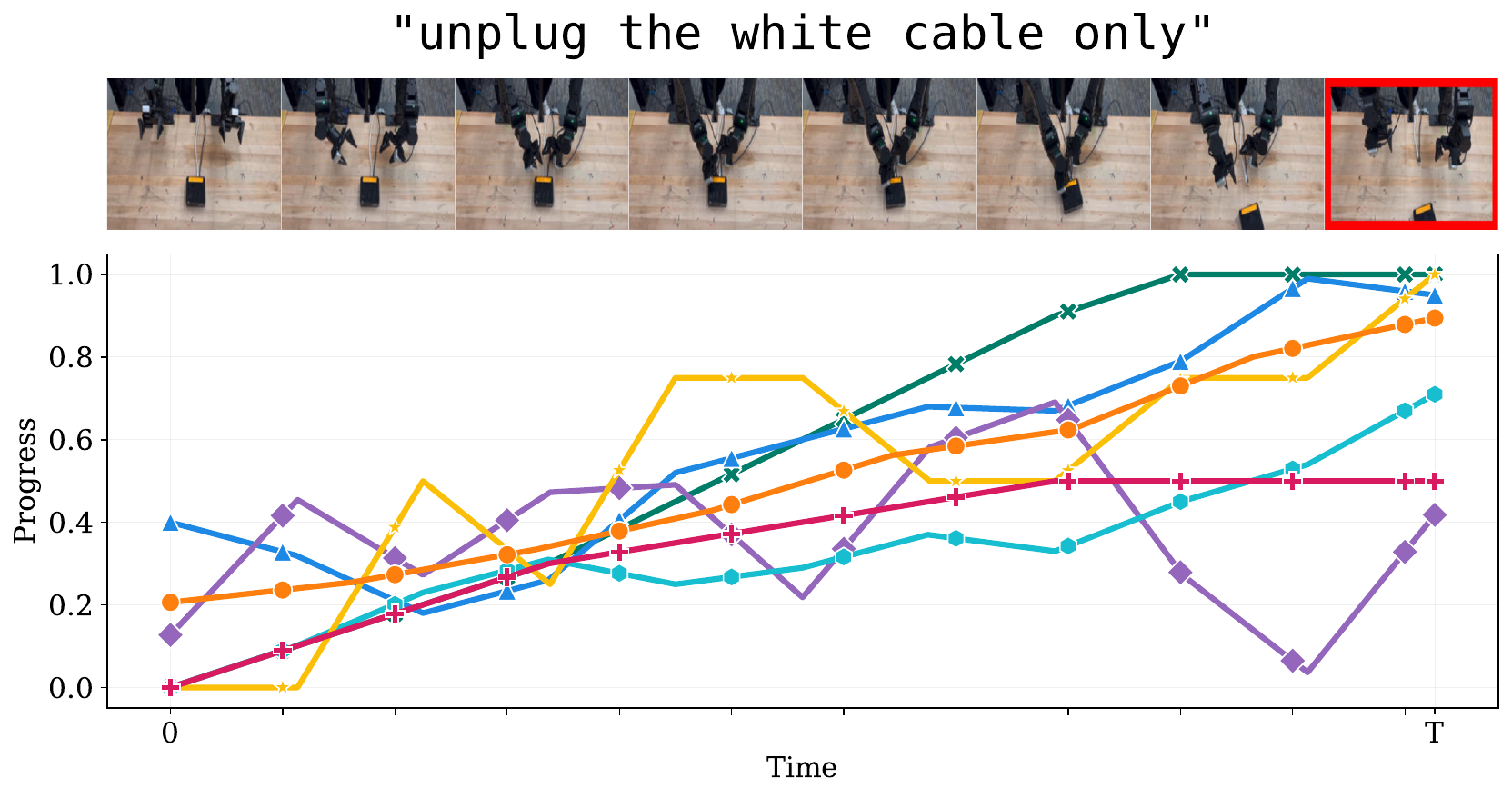}
        \caption{``unplug the white cable only'' (failure).}
    \end{subfigure}
    \caption{\textbf{Qualitative per-trajectory predictions for ``unplug the white cable only''.}}
    \label{fig:qualitative_cable_task}
\end{figure*}

\begin{figure*}[!h]
    \centering
    \includegraphics[width=0.6\columnwidth]{plots/legend.pdf}\\[4pt]
    \begin{subfigure}[t]{\qualimgw}
        \qualimg{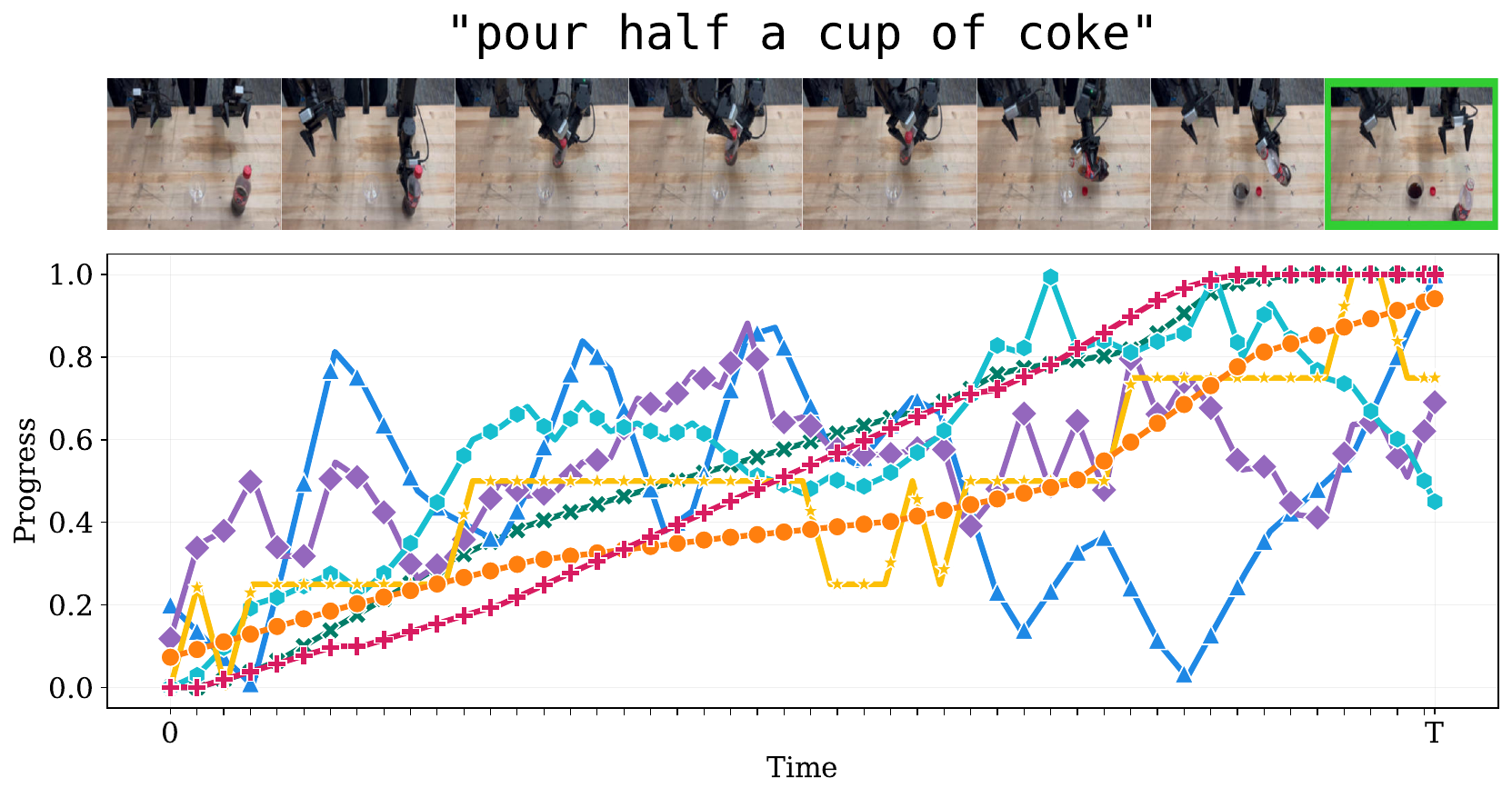}
        \caption{``pour half a cup of coke'' (success).}
    \end{subfigure}
    \hfill
    \begin{subfigure}[t]{\qualimgw}
        \qualimg{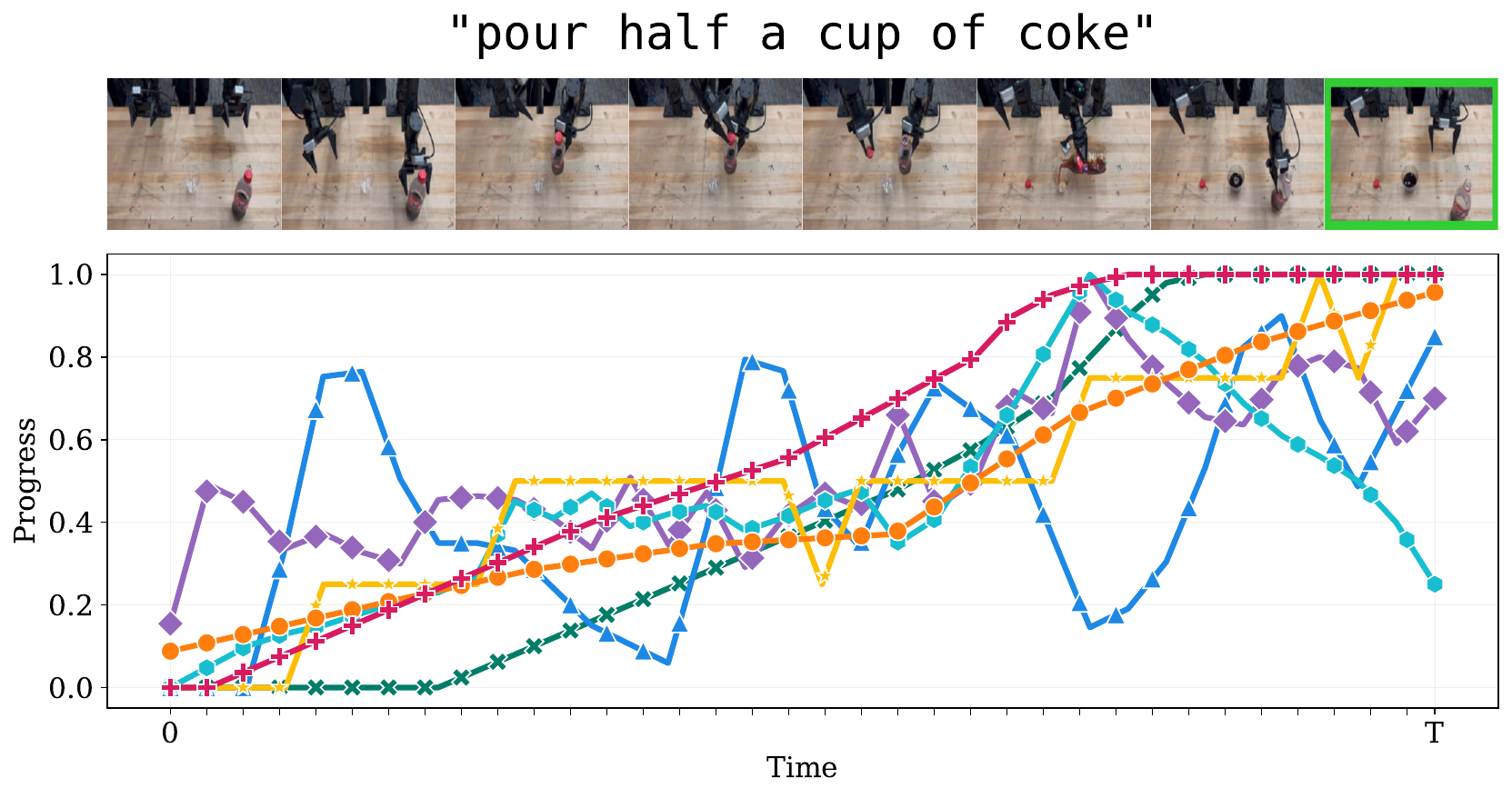}
        \caption{``pour half a cup of coke'' (success).}
    \end{subfigure}

    \vspace{0.5em}
    \begin{subfigure}[t]{\qualimgw}
        \qualimg{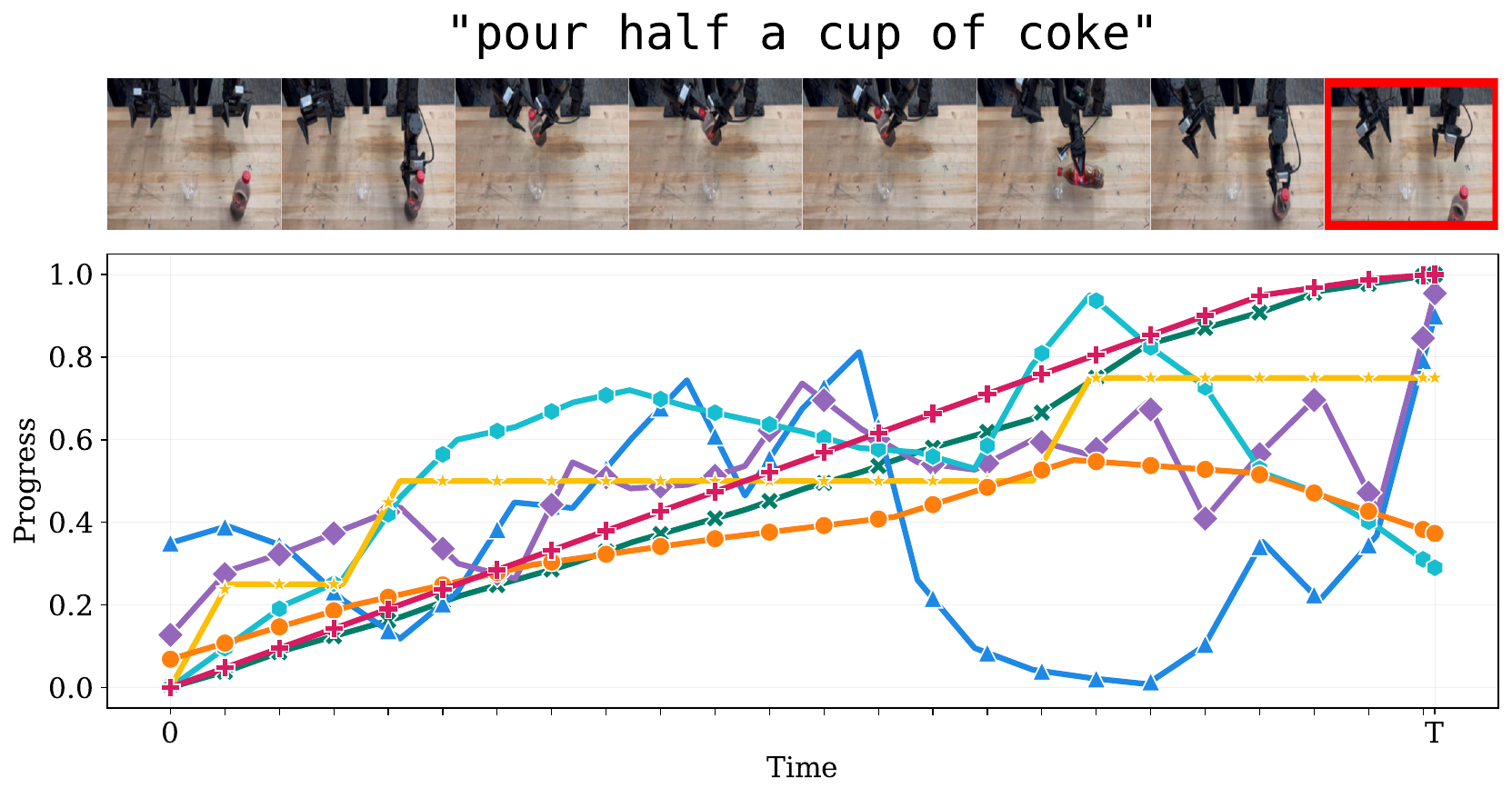}
        \caption{``pour half a cup of coke'' (failure).}
    \end{subfigure}
    \hfill
    \begin{subfigure}[t]{\qualimgw}
        \qualimg{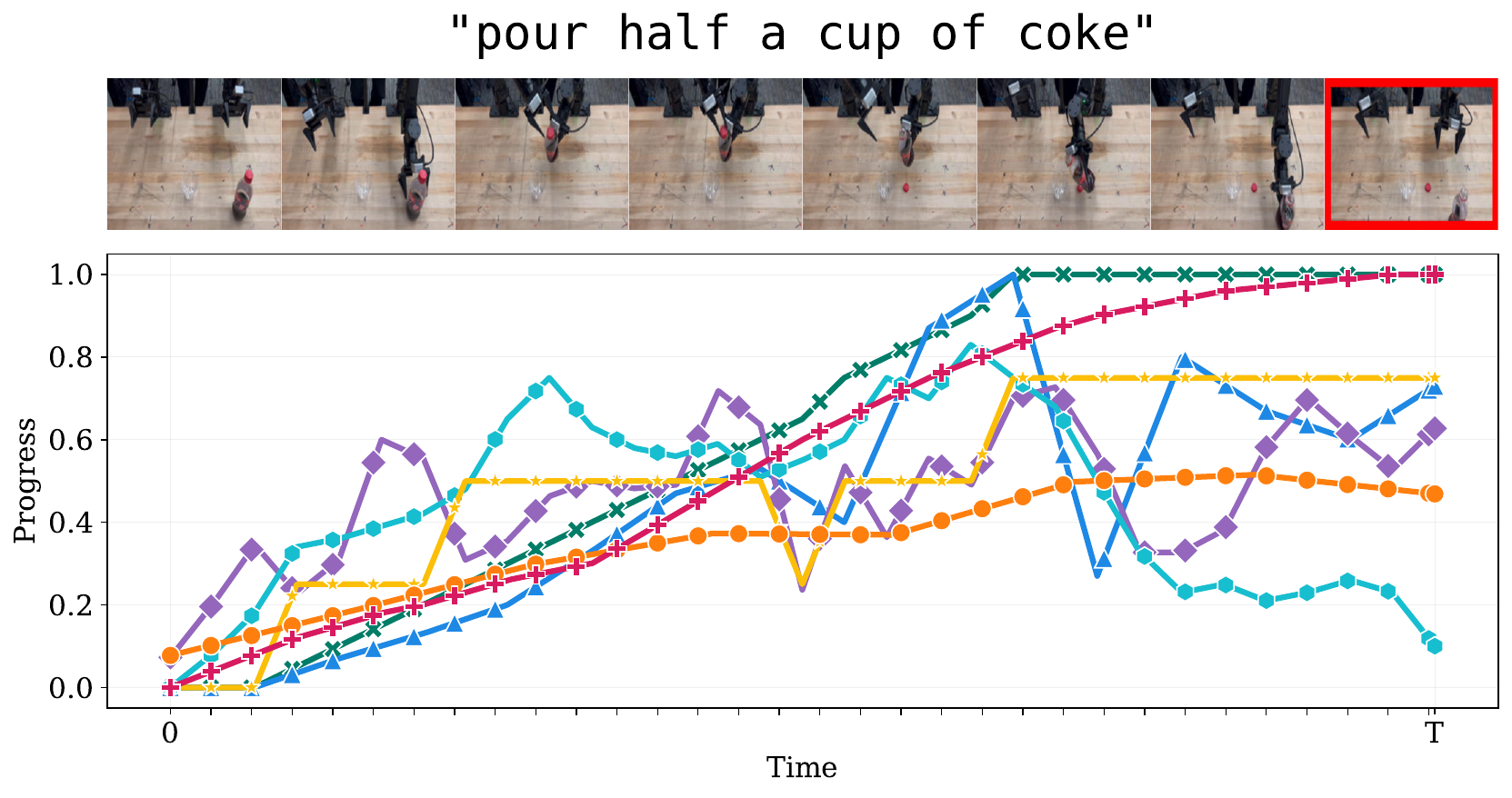}
        \caption{``pour half a cup of coke'' (failure).}
    \end{subfigure}
    \caption{\textbf{Qualitative per-trajectory predictions for ``pour half a cup of coke''.}}
    \label{fig:qualitative_coke_task}
\end{figure*}

\begin{figure*}[!h]
    \centering
    \includegraphics[width=0.6\columnwidth]{plots/legend.pdf}\\[4pt]
    \begin{subfigure}[t]{\qualimgw}
        \qualimg{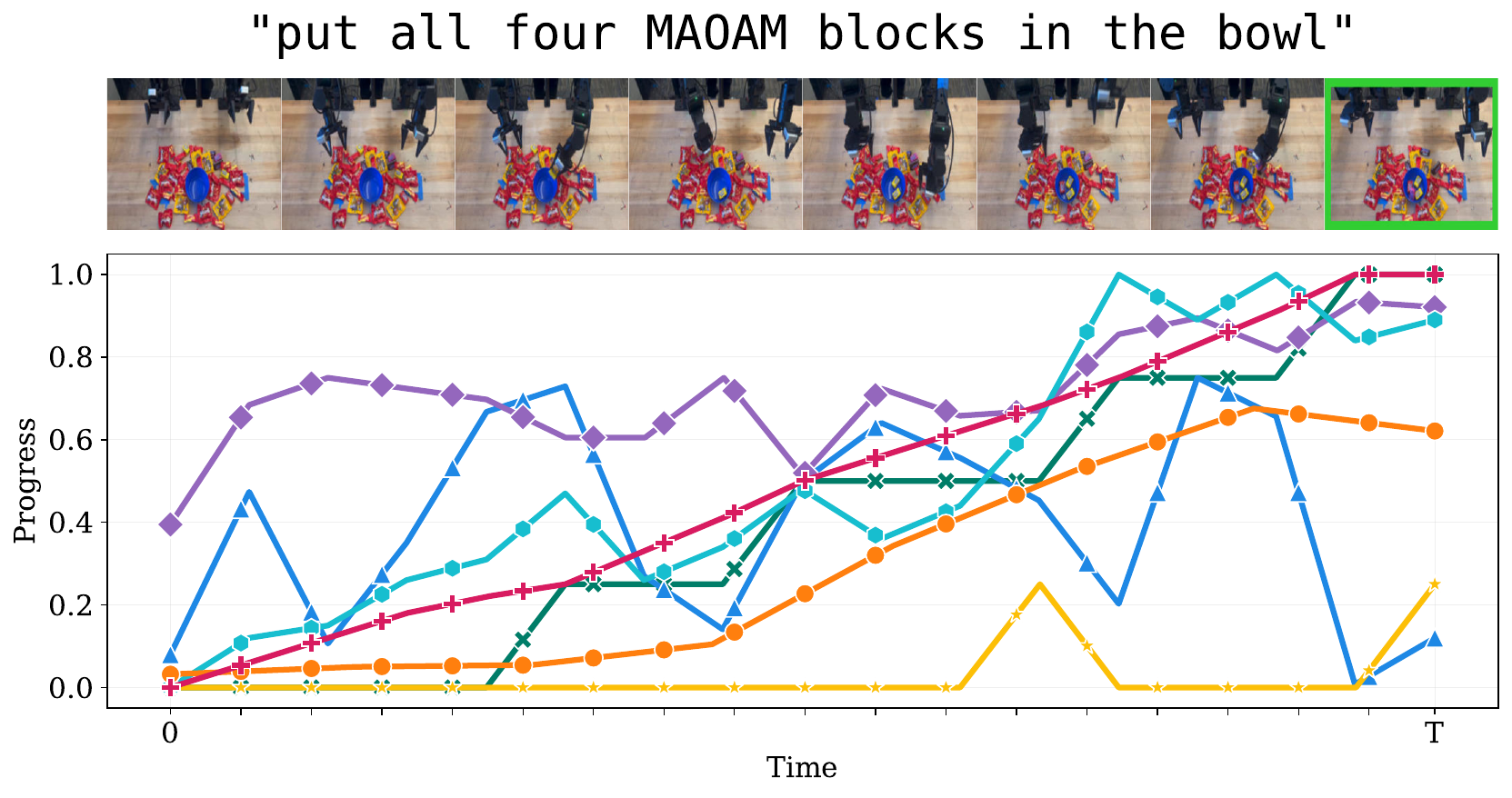}
        \caption{``put all four MAOAM blocks in the bowl'' (success).}
    \end{subfigure}
    \hfill
    \begin{subfigure}[t]{\qualimgw}
        \qualimg{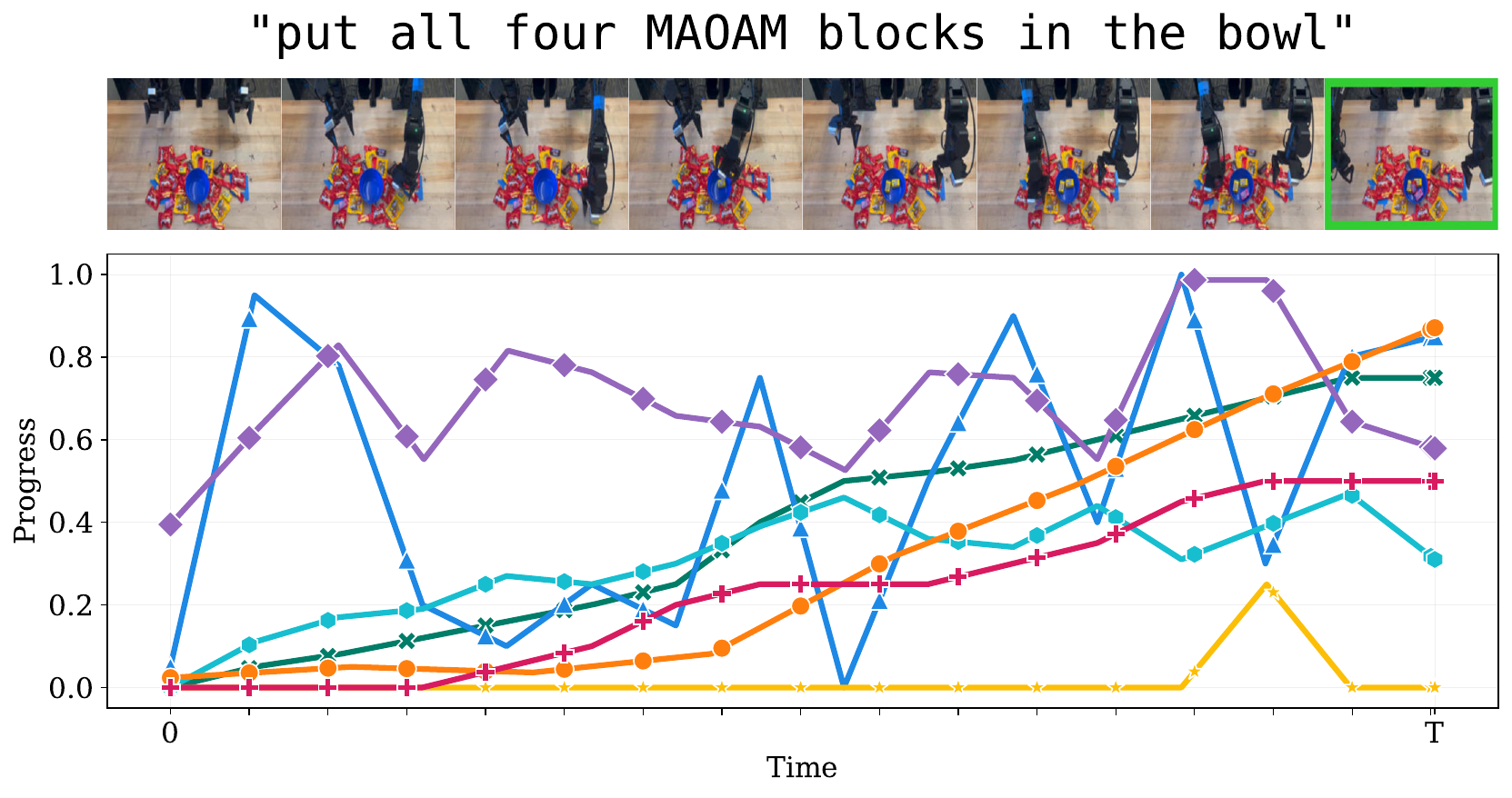}
        \caption{``put all four MAOAM blocks in the bowl'' (success).}
    \end{subfigure}

    \vspace{0.5em}
    \begin{subfigure}[t]{\qualimgw}
        \qualimg{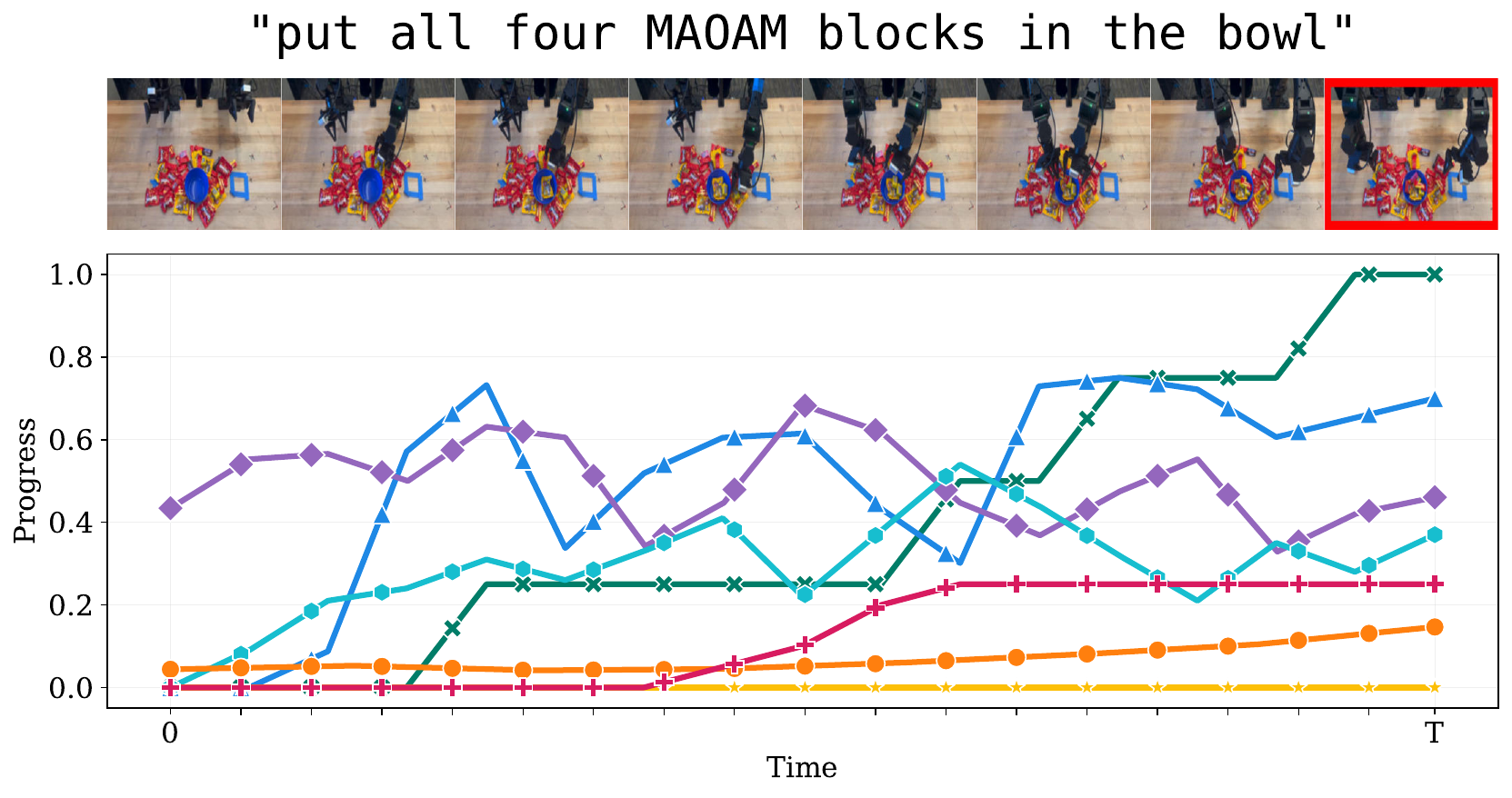}
        \caption{``put all four MAOAM blocks in the bowl'' (failure).}
    \end{subfigure}
    \hfill
    \begin{subfigure}[t]{\qualimgw}
        \qualimg{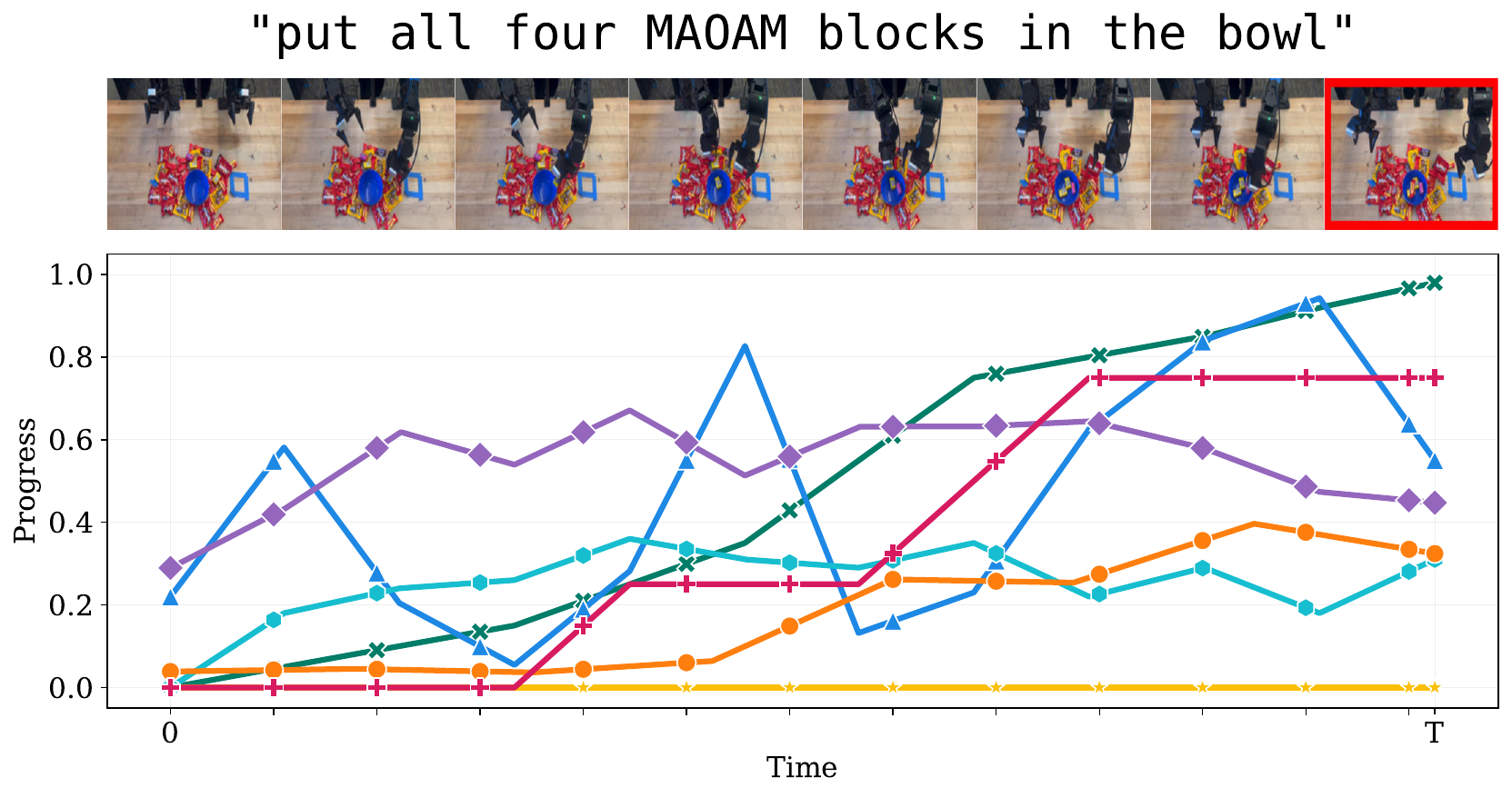}
        \caption{``put all four MAOAM blocks in the bowl'' (failure).}
    \end{subfigure}
    \caption{\textbf{Qualitative per-trajectory predictions for ``put all four MAOAM blocks in the bowl''.}}
    \label{fig:qualitative_maoam4_task}
\end{figure*}

% Force all 14 pending figure floats to flush here, before the next
% \section. Without this, LaTeX can push overflowing floats past the
% \section{...Prompts} heading (floats don't respect section boundaries),
% so prompt text ends up sandwiched between / above qualitative figures.
\clearpage

\section{\method{} Prompts}
\label{sec:appdx:prompts}
\subsection{Naive VLM Baseline}
\begin{lstlisting}[style=prompt]
Task:
"{{task}}"

1. The task text fixes which objects count (category, color, shape, and wording). The video frames are the only evidence of what is in the environment. Do not "ground" the task by renaming visible objects to match the task.
2. For each object the task requires, it may be tied only to objects visible in the video that refer to that same entity without contradiction. Do not substitute a different object because it could serve the same role. If the visual appearance of an object conflicts with the task's description of that object, or no clear referent is visible, treat that requirement as absent.
3. You may track the same physical object across frames only when it is plainly the same entity already visible in the video--not to map task nouns onto other objects.
4. For each timestep, reason about the % progress towards that fully completed state (0-100%).

Respond with a list of progress values (ordered by frame) in the following JSON format:
[
    {{
        "progress": [int, int, ...]
    }}
]
\end{lstlisting}

% description from video (grounded)
\subsection{\method}
\subsubsection{Grounding}
\mbox{}\par\nobreak
\begin{lstlisting}[style=prompt]
<role>
You are an expert in object recognition.
</role>

<task>
1. all_objects: list objects that are in the video.
2. are some objects relevant to the following task? Task: "{task}" If yes, list them in task_objects. If no, return an empty list for task_objects.
</task>

<output_format>
Respond with a list of objects in the following JSON format:
[
    {
        "all_objects": [str, str, ...],
        "task_objects": [str, str, ...],
    }
]
</output_format>
\end{lstlisting}
\subsubsection{Grounded Video Description}
\mbox{}\par\nobreak
\begin{lstlisting}[style=prompt]

<role>
You are an expert in scene understanding. Provide highly detailed descriptions including the robot and state of the following objects {objects} at each timestep. Describe robot motion and distance between robot and objects.
Without looking at the rest of the video, describe the scene in the first and last frame to ground the descriptions.
</role>

<constraints>
- The video may be captured from either a third-person or wrist-mounted (robot POV) camera.
- Do not make any judgement about what the robot is trying to accomplish. The robot is imperfect and might do things that don't make any sense to you.
- Describe the scene as objectively as possible.
</constraints>

<output_format>
Respond with exactly {N} descriptions, where {N} is the number of video frames. There is only one string per frame. in the following JSON format:
[
    {
        "description_first_frame": str,
        "description_last_frame": str,
        "descriptions": [str, str, ...],
    }
]
</output_format>
\end{lstlisting}

\subsubsection{Success Criteria and Assessment}
\mbox{}\par\nobreak
% progress from description
\begin{lstlisting}[style=prompt]
Task:
"{{task}}"

Scene descriptions:
{{description}}

1. The task text fixes which objects count (category, color, shape, and wording). Scene descriptions are the only evidence of what is in the environment. Do not "ground" the task by renaming scene objects to match the task.
2. For each object the task requires, it may be tied only to scene mentions that refer to that same entity without contradiction. Do not substitute a different object because it could serve the same role. If attributes in the scene conflict with the task's description of that object, or no clear referent exists, treat that requirement as absent.
3. You may align short paraphrases across timesteps only when they plainly denote the same physical object already present in the scene--not to map task nouns onto other objects.
4. If ALL required objects are absent by these rules, the task cannot be completed in this scene: use all 0% progress and explain in "completion state". Otherwise define precisely what the environment would look like if the task were FULLY and COMPLETELY finished (not started, not halfway--entirely done). Reward partial progress if the task is only partially completed and some required objects are present.
5. For each timestep, reason about the % progress towards that fully completed state.
6. Progress is defined as % progress towards the fully completed state. Progress does not have to be monotonic, i.e., the robot can reverse progress.

Respond with a list of progress values (ordered by frame) in the following JSON format:
[
    {{
        "completion state": str,
        "progress": [int, int, ...]
    }}
]

Before you respond, reflect whether the task is actually fully completed, otherwise adjust the progress values!
\end{lstlisting}

% progress from description - roboreward
\subsubsection{Success Criteria and Assessment - Modified for RoboRewardBenchmark}
\mbox{}\par\nobreak
\begin{lstlisting}[style=prompt]

Task:
"{{task}}"

Scene descriptions:
{{description}}

Given the task instruction and a description of the robot's rollout video (or the final frame observation), assign a discrete progress score reward (1, 2, 3, 4, 5) for the robot's performance. Judge only the final state without considering time limits. 

Rubric for End-of-Episode Progress:
* 1 - No Success: The final state shows no goal-relevant change for the command.
* 2 - Minimal Progress: The final state shows a small but insufficient change toward the goal.
* 3 - Partial Completion: The final state shows good progress toward the goal but violates more than one requirement or a major requirement.
* 4 - Near Completion: The final state is correct in region and intent but misses a single minor requirement.
* 5 - Perfect Completion: The final state visibly satisfies all requirements.

Hints & Clarifications for Scoring (Particularly 3 vs. 4):
To correctly distinguish between a Partial Completion (3) and a Near Completion (4), you must differentiate between MAJOR and MINOR task requirements based on the following rules:
* Treat a requirement as MAJOR when the PRIMARY object identity or the PRIMARY spatial relation to the reference object is not satisfied. If a major requirement is missed, the score must be 3 or lower.
* Treat a requirement as MINOR when the PRIMARY object identity and PRIMARY spatial relation to the reference object are satisfied, but an AUXILIARY constraint is missed (e.g., the orientation of the object is slightly off, or the object is placed a bit too far forward/backward but still in the correct general region). Missing a single minor requirement results in a score of 4.

Provide a brief step-by-step reasoning analyzing the observation against the rubric and hints. Conclude with your final discrete score in the following JSON format: 
[
    {{
        "score": int
    }}
]
\end{lstlisting}

\subsubsection{Success Criteria and Assessment - Modified for Failure Analysis and Feedback Prediction}
\mbox{}\par\nobreak

\begin{lstlisting}[style=prompt]

Task:
"{{task}}"

Scene descriptions:
{{description}}

1. The task text fixes which objects count (category, color, shape, and wording). Scene descriptions are the only evidence of what is in the environment. Do not "ground" the task by renaming scene objects to match the task.
2. For each object the task requires, it may be tied only to scene mentions that refer to that same entity without contradiction. Do not substitute a different object because it could serve the same role. If attributes in the scene conflict with the task's description of that object, or no clear referent exists, treat that requirement as absent.
3. You may align short paraphrases across timesteps only when they plainly denote the same physical object already present in the scene-not to map task nouns onto other objects.
4. If ALL required objects are absent by these rules, the task cannot be completed in this scene: use all 0% progress and explain in "completion state". Otherwise define precisely what the environment would look like if the task were FULLY and COMPLETELY finished (not started, not halfway-entirely done). Reward partial progress if the task is only partially completed and some required objects are present.
5. For each timestep, reason about the % progress towards that fully completed state.
6. Progress is defined as % progress towards the fully completed state. Progress does not have to be monotonic, i.e., the robot can reverse progress.
7. If the task was not fully completed, describe the failure.
8. If the task was not fully completed, give feedback on what the robot should have done differently.

Respond with a list of progress values (ordered by frame) in the following JSON format:
[
    {{
        "failure": str,
        "feedback": str,
        "completion state": str,
        "progress": [int, int, ...]
    }}
]

Before you respond, reflect whether the task is actually fully completed, otherwise adjust the progress values!
\end{lstlisting}

\newpage

\section{Failure Analysis and Feedback Prediction Examples}
\label{sec:appdx:feedback}

% \newpage
\begin{figure}[h]
    \centering
    \includegraphics[width=0.7\linewidth]{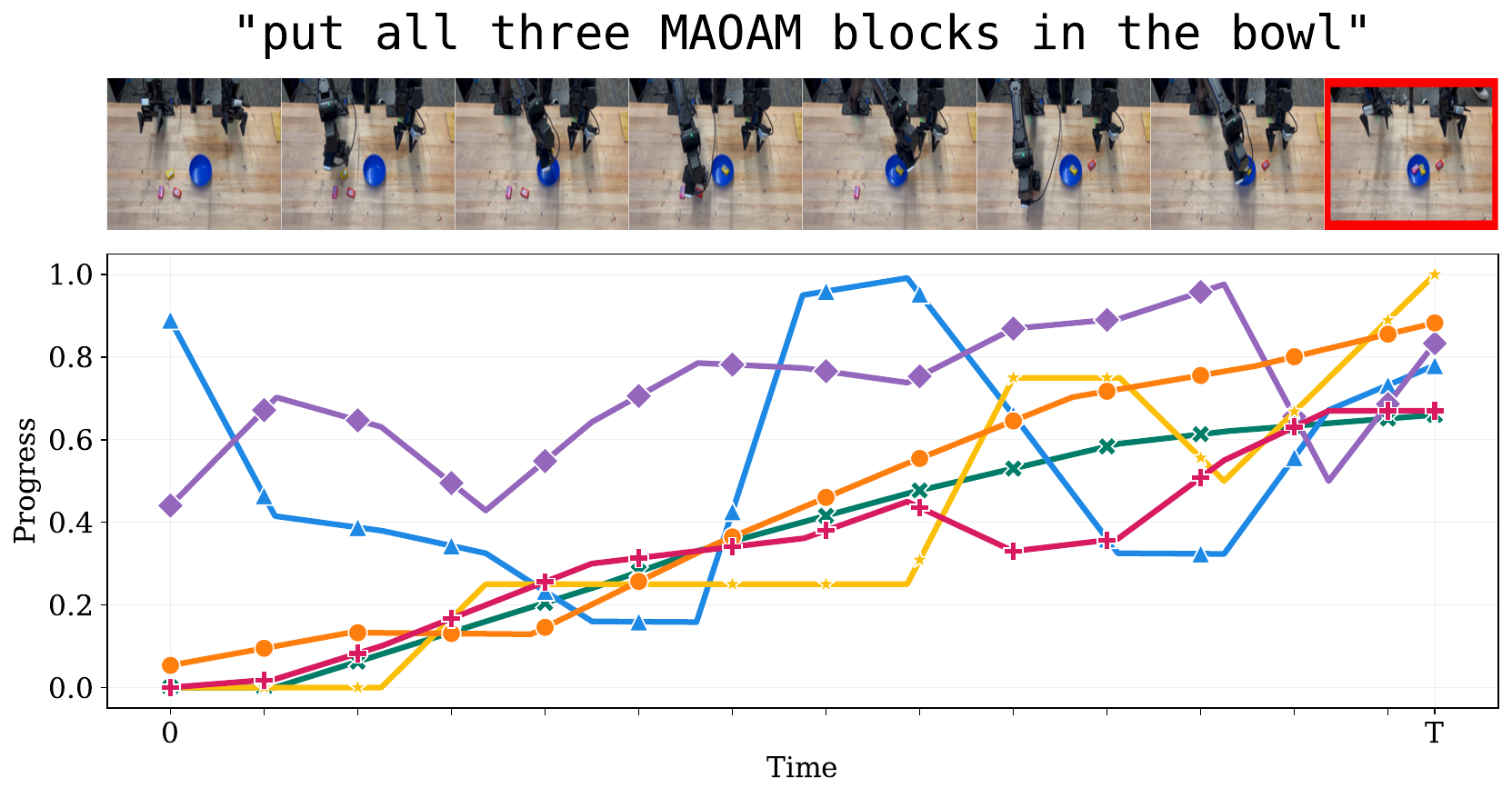}
    \includegraphics[width=0.7\linewidth]{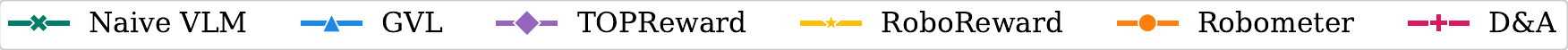}
\end{figure}

Grounded Video Description:
\begin{lstlisting}[style=prompt]
[
    {'description': 'Robotic arms idle over a wooden table featuring a central blue bowl and three candy blocks.'},
    {'description': 'The left robotic arm descends toward the yellow MAOAM block; the right arm remains stationary.'},
    {'description': 'The left arm reaches and grasps the yellow block on the wooden table.'},
    {'description': 'The left robotic arm lifts the yellow block, increasing distance from the table surface.'},
    {'description': 'The left arm transports the yellow block toward the blue bowl.'},
    {'description': 'The left arm releases the yellow block into the blue bowl.'},
    {'description': 'The left robotic arm moves from the bowl toward the red MAOAM block.'},
    {'description': 'The left arm descends and secures the red MAOAM block.'},
    {'description': 'The left arm carries the red block toward the right side of the wooden table.'},
    {'description': 'The red block is dropped onto the table surface near the right robotic arm.'},
    {'description': 'The left arm maneuvers across the table toward the pink MAOAM block.'},
    {'description': 'The left robotic arm grasps and lifts the pink block.'},
    {'description': 'The left arm drops the pink block into the blue bowl and retracts.'}
]
\end{lstlisting}

Failure Analysis, Feedback Prediction, Success Criteria, and Score:
\begin{lstlisting}[style=prompt]
[
  {
    "failure": "The robot only placed the yellow and pink MAOAM blocks into the bowl; the red block was moved to the right side of the table and dropped there instead of being placed in the bowl.",
    "feedback": "To complete the task, the robot should have moved the red block to the blue bowl after picking it up, rather than dropping it on the right side of the table.",
    "completion state": "All three MAOAM blocks (yellow, red, and pink) are contained within the blue bowl on the wooden table.",
    "progress": [0, 0, 0, 0, 0, 33, 33, 33, 33, 33, 33, 33, 67]
  }
]
\end{lstlisting}

\newpage
\begin{figure}[h]
    \centering
    \includegraphics[width=0.7\linewidth]{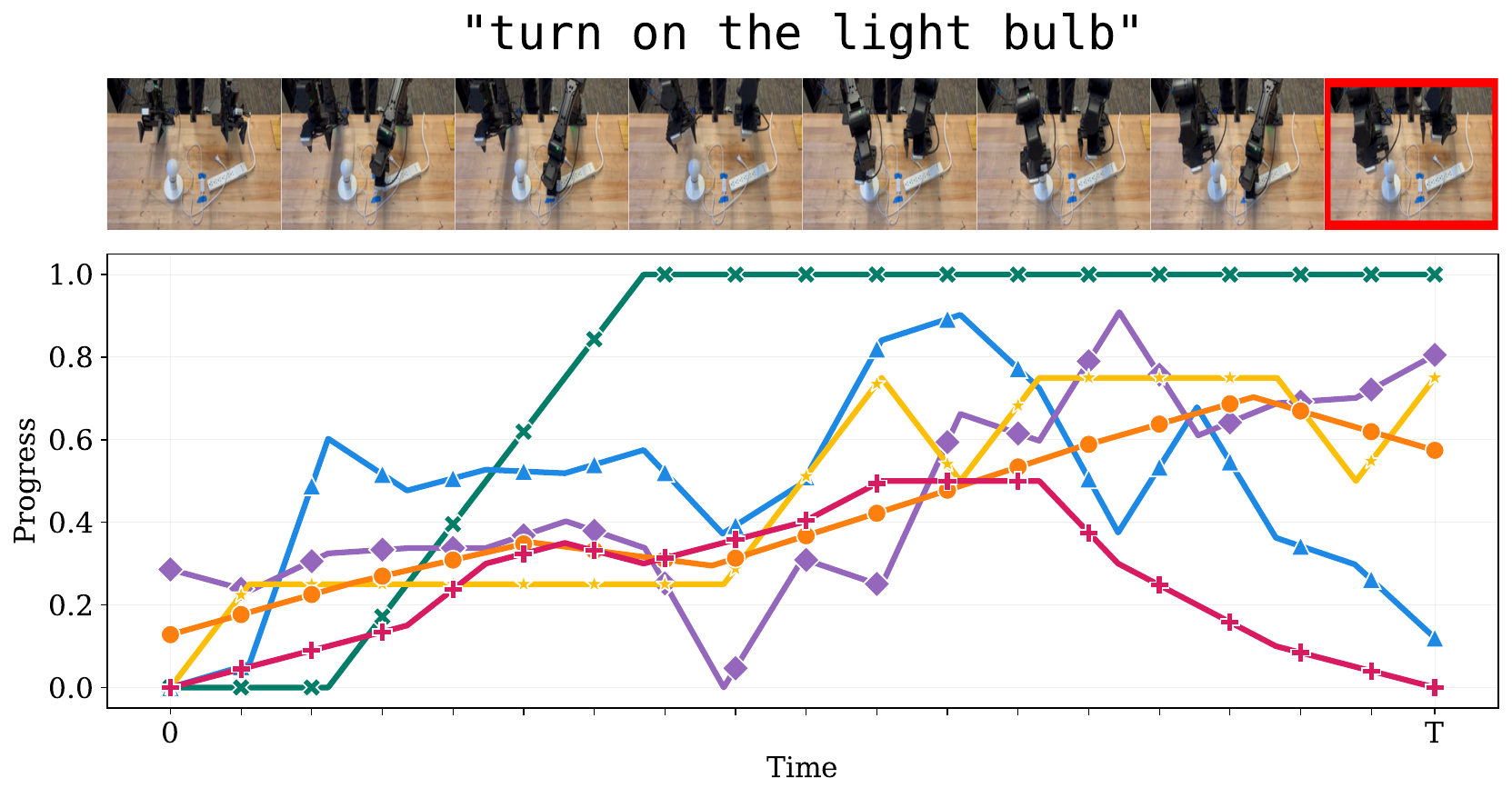}
    \includegraphics[width=0.7\linewidth]{figures/qual_legend.pdf}
\end{figure}

Grounded Video Description:
\begin{lstlisting}[style=prompt]
[
    {'description': 'In the initial frame, two robot arms are retracted above a light-colored wooden table. On the table, a white light bulb in a base is on the left, a power strip is on the right, and a white cord with a switch and a plug connects the two. The switch is taped to the table with two strips of blue tape.'},
    {'description': 'The right robot arm begins to lower its gripper towards the white switch located in the middle of the table, while the left robot arm remains stationary in its elevated position.'},
    {'description': "The right robot arm's gripper reaches the switch, closing the distance to just a few centimeters as it prepares to interact with the device."},
    {'description': "The right robot arm's gripper makes contact with the white switch, attempting to manipulate it. The light bulb remains unlit. The left arm stays in its high position."},
    {'description': 'The right robot arm continues to hold its gripper against the white switch, maintaining direct contact with the component secured by blue tape.'},
    {'description': "The right robot arm's gripper remains in contact with the switch, while the left arm is still idle at the top of the frame."},
    {'description': 'The right robot arm is still positioned with its gripper touching the switch, showing minimal movement from the previous frame.'},
    {'description': 'The right robot arm has retracted to its original height, while the left robot arm begins its descent toward the white light bulb on the left side of the table.'},
    {'description': "The left robot arm's gripper is now directly over the white light bulb, roughly 10 centimeters away, while the right arm is fully retracted."},
    {'description': 'The left robot arm lowers its gripper until it makes contact with the top surface of the white light bulb.'},
    {'description': 'The left robot arm maintains its position with the gripper touching the light bulb. The right robot arm remains inactive and elevated.'},
    {'description': 'The left robot arm continues to keep its gripper in contact with the light bulb. The distance between the gripper and the bulb remains at zero.'},
    {'description': 'The left robot arm begins to lift its gripper away from the light bulb, creating a vertical gap of several centimeters.'},
    {'description': 'As the left robot arm retracts further, the right robot arm begins a second descent, moving its gripper back towards the center of the table near the power strip and plug.'},
    {'description': "The right robot arm's gripper is now close to the white plug and power strip on the right side of the wooden table, approximately 5 centimeters away."},
    {'description': 'The right robot arm starts to move upwards again, increasing the distance between its gripper and the power strip on the table surface.'},
    {'description': 'In the final frame, both robot arms have returned to their high, retracted positions. The light bulb, power strip, switch, and cord remain on the table in their original configuration.'}
]
\end{lstlisting}

Failure Analysis, Feedback Prediction, Success Criteria, and Score:
\begin{lstlisting}[style=prompt]
[
    {
        "failure": "The robot failed to turn on the light bulb. It successfully reached the switch and the plug area but did not actuate the switch or complete the electrical connection required to light the bulb, eventually returning to its starting position with the bulb still off.",
        "feedback": "The robot should have applied the necessary force or correct motion to flip the switch to the 'on' position. It also should have ensured the plug was fully inserted into the power strip. Touching the bulb itself was not necessary for this task.",
        "completion state": "The light bulb is illuminated, which requires the power cord to be plugged into the power strip and the switch to be flipped to the 'on' position.",
        "progress": [0, 10, 20, 30, 30, 30, 30, 15, 10, 10, 10, 10, 5, 15, 25, 10, 0]
    }
]
\end{lstlisting}

\newpage
\begin{figure}[h]
    \centering
    \includegraphics[width=0.7\linewidth]{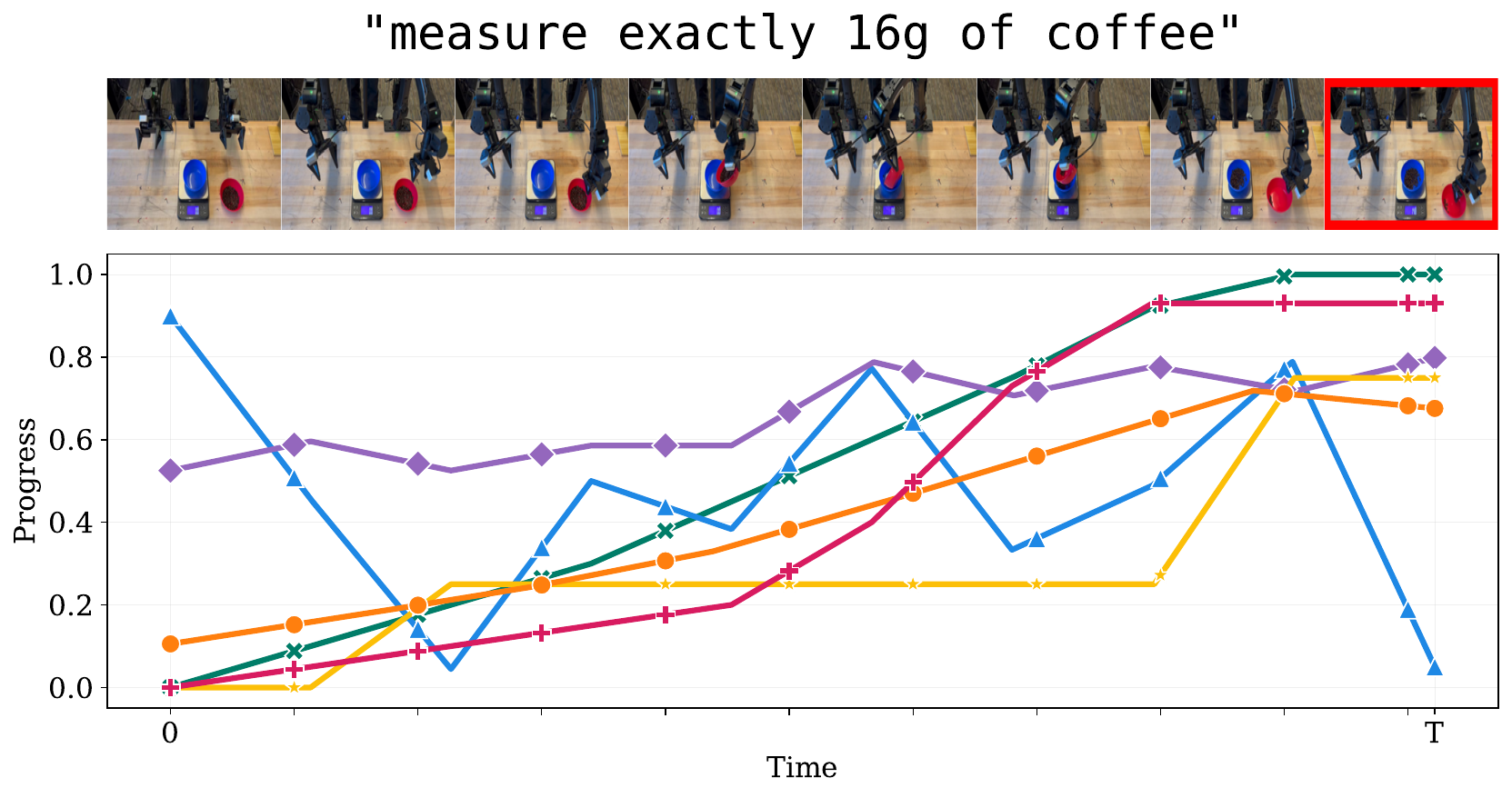}
    \includegraphics[width=0.7\linewidth]{figures/qual_legend.pdf}
\end{figure}

Grounded Video Description:
\begin{lstlisting}[style=prompt]
[
    {'description': 'Robotic arms hover ~20cm above the table. Blue bowl is on the scale; red bowl with coffee beans is stationary.'},
    {'description': 'The right robotic arm descends toward the red bowl, closing the distance to ~10cm.'},
    {'description': "Right arm gripper reaches the red bowl; left arm remains static. Scale still reads '0.0'."},
    {'description': "Right arm grasps the red bowl's rim. Distance between gripper and bowl is zero."},
    {'description': 'Right arm lifts the red bowl ~15cm and moves it horizontally toward the blue bowl on the scale.'},
    {'description': 'Right arm tilts the red bowl over the blue bowl. Coffee beans begin falling into the blue bowl.'},
    {'description': 'Right arm continues tilting; beans transfer to the blue bowl. The digital scale display fluctuates.'},
    {'description': 'Right arm levels the red bowl and begins moving it back to the right side of the table.'},
    {'description': "Right arm lowers the red bowl onto the wooden table. The scale display stabilizes at '14.8'."},
    {'description': 'Right arm releases the red bowl and retracts upward. The coffee beans remain in the blue bowl on the scale.'}
]
\end{lstlisting}

Failure Analysis, Feedback Prediction, Success Criteria, and Score:
\begin{lstlisting}[style=prompt]
[
  {
    "failure": "The robot stopped pouring coffee beans prematurely, resulting in a final measurement of 14.8g instead of the required 16.0g.",
    "feedback": "The robot should have continued to tilt the red bowl and monitor the digital scale until the weight reached exactly 16.0g before leveling the bowl and moving it away.",
    "completion state": "The blue bowl sits on the digital scale, which displays a measurement of exactly 16.0g of coffee beans.",
    "progress": [
      0,
      5,
      10,
      15,
      25,
      40,
      70,
      92,
      92,
      92
    ]
  }
]
\end{lstlisting}

\end{document}